%% file: main.tex
\pdfoutput=1
\def\paperformat{preprint}
\documentclass{article}

\usepackage{etoolbox}
\usepackage{comment}
\providecommand*{\paperformat}{iclr}
\ifboolexpr{
  test {\ifdefstring{\paperformat}{iclr}} or
  test {\ifdefstring{\paperformat}{preprint}} or
  test {\ifdefstring{\paperformat}{simbiochem}}
}{}{\errmessage{Unknown paper format: \paperformat}}
\newif\ifpaperpublic
\ifdefstring{\paperformat}{iclr}{}{\paperpublictrue}
\newif\ifpaperworkshop
\ifdefstring{\paperformat}{simbiochem}{\paperworkshoptrue}{}

\ifpaperworkshop
  \usepackage[dblblindworkshop, final]{neurips_2026}
  \workshoptitle{2nd SIMBIOCHEM Workshop at NeurIPS 2026}
  \usepackage{times}
  \setcitestyle{authoryear,round,citesep={;},aysep={,},yysep={;}}
\else
  \usepackage{iclr2027_conference,times}
  \ifpaperpublic
    \iclrfinalcopy
    \expandafter\apptocmd\csname @maketitle\endcsname{\lhead{}}{}{}
    
    \newenvironment{ack}{\subsubsection*{Acknowledgments}}{}
  \else
    \excludecomment{ack}
  \fi
\fi

\usepackage[T1]{fontenc}
\usepackage{booktabs}
\usepackage{amsfonts}
\usepackage{amsmath}
\usepackage{bm}
\newcommand{\E}{\mathbb{E}}
\newcommand{\KL}{D_{\mathrm{KL}}}
\newcommand{\mP}{{\bm{P}}}
\newcommand{\1}{\bm{1}}
\usepackage{graphicx}
\usepackage{wrapfig}
\usepackage{capt-of}
\usepackage{float}
\usepackage{enumitem}
\usepackage{xspace}
\newcommand{\rESS}{\ensuremath{\mathrm{rESS}}\xspace}
\newcommand{\rESSast}{\ensuremath{\mathrm{rESS}^{\ast}}\xspace}
\newcommand{\EW}{\ensuremath{\mathcal{E}\text{-}\mathcal{W}_2}\xspace}
\newcommand{\EWast}{\ensuremath{\mathcal{E}\text{-}\mathcal{W}_2^{\ast}}\xspace}
\newcommand{\TW}{\ensuremath{\mathbb{T}\text{-}\mathcal{W}_2}\xspace}
\newcommand{\TgSWast}{\ensuremath{\mathbb{T}\mathrm{g}\text{-}\mathcal{SW}_2^{\ast}}\xspace}
\newcommand{\TICAW}{\ensuremath{\mathrm{TICA}\text{-}\mathcal{W}_2}\xspace}
\newcommand{\CstarEW}{\ensuremath{C^{\ast}_{\mathcal{E}\text{-}\mathcal{W}_2}}\xspace}
\usepackage{microtype}
\usepackage{xcolor}
\definecolor{rebuttalgreen}{RGB}{0,100,0}
\newcommand{\rebuttal}[1]{\textcolor{rebuttalgreen}{#1}}
\ifpaperpublic
  \renewcommand{\rebuttal}[1]{#1}
  \colorlet{rebuttalgreen}{black}
\fi
\newcommand{\publictext}[2]{\ifpaperpublic#1\else#2\fi}
\usepackage{tikz}
\usetikzlibrary{arrows.meta,backgrounds,calc,decorations.pathreplacing,fit,positioning}
\usepackage[pagebackref]{hyperref}
\usepackage{arydshln}

\newlength{\catalogmaxwd}
\newlength{\mdqmtablewidth}
\newsavebox{\catalogprobebox}
\AtBeginDocument{%
  \begingroup
    \renewcommand{\resizebox}[3]{#3}%
    \setlength{\catalogmaxwd}{0pt}%
    \foreach \catalogtable in {%
      sota_plot,sota_modern_plot,dipper_mdqm9_table,%
      dipper_solvated_solvator_comparison,dipper_vacuum_peptides_table,%
      joint_scaling_plot,joint_scaling_r2}{%
      \sbox{\catalogprobebox}{\input{generated/\catalogtable.tex}}%
      \ifdim\wd\catalogprobebox>\catalogmaxwd
        \global\catalogmaxwd=\wd\catalogprobebox
      \fi
    }%
  \endgroup
}

\AddToHook{env/table/begin}{\let\resizebox\papertableresize}

\let\cite\citep
\input{generated/queries.tex}
\newcommand{\backlink}[1]{{\normalfont\small($\leftarrow$~Section~\ref{#1})}}
\renewcommand*{\backrefalt}[4]{%
  \ifnum #1=0
  \else
    \ifnum #3>1 ($\leftarrow$~pp.~#4) \else ($\leftarrow$~p.~#4) \fi
  \fi
}

\definecolor{aithyrapetrol}{HTML}{0A6478}
\hypersetup{
  pdftitle={},
  pdfauthor={},
  pdfcreator={},
  pdfproducer={},
  colorlinks=true,
  citecolor=aithyrapetrol,
  linkcolor=black,
  urlcolor=black
}
\ifpaperpublic
  \hypersetup{
    pdftitle={Prefix-Conditioned Boltzmann Generators},
    pdfauthor={Niklas Rindtorff and Alexander Tong}
  }
\fi

\newcommand{\namelong}{\textsc{Prefix-Conditioned Boltzmann Generators}\xspace}
\newcommand{\nameshort}{\textsc{PBGs}\xspace}
\title{Prefix-Conditioned Boltzmann Generators}

\ifpaperpublic
  \author{%
    Niklas Rindtorff \& Alexander Tong \\
    AITHYRA \\
    \texttt{\{nrindtorff, atong\}@aithyra.at}
  }
\else
  \author{%
    Name \\
    Affiliation \\
    Address \\
    \texttt{email}
  }
\fi

\begin{document}

\maketitle

\begin{abstract}
Reliable estimation of equilibrium molecular properties requires sampling the Boltzmann distribution, a central bottleneck in molecular simulation.
Boltzmann Generators (BGs) use generative models with tractable likelihoods to enable importance reweighting to the Boltzmann distribution.
Recent work has scaled BG training to larger and more diverse datasets, but existing approaches often specialize in particular molecular classes and single molecules.
We introduce \namelong (\nameshort), a family of autoregressive Boltzmann generators conditioned on text and coordinate prefixes, without a specialized atom-wise molecular encoder.
Across a series of models trained on \textit{Many Peptides}, we identify scaling relationships linking sampling quality to model size, training tokens, training compute, atom count, and coordinate discretization resolution.
Guided by these relationships, we scale \nameshort up to 640M parameters, achieving state-of-the-art effective sample sizes as well as energy and torsion-angle accuracy on unseen peptide systems containing up to 157 atoms.
Beyond peptides, we demonstrate the versatility of \nameshort through small-molecule sampling, solvation free energy estimation, and conditional microsolvation of systems containing over 280 atoms.
Together, these results support \nameshort as a promising approach to \rebuttal{further generalize transferable} molecular sampling.
\end{abstract}

\input{generated/alanine_dipeptide_solute_tokens_tikz}

\section{Introduction}

In statistical mechanics, the properties of a material at equilibrium, for example a therapeutic small molecule, are defined by a probability density over its atomic coordinates induced by random collisions between the molecule and its constant-temperature environment, the Boltzmann distribution.
Sampling these distributions across relevant states and environments supports the estimation of properties such as binding affinity and solubility.
However, obtaining independent equilibrium samples remains a long-standing computational challenge.
The dominant approach, molecular dynamics (MD), integrates atomic equations of motion and, with suitable temperature control, can sample the target equilibrium distribution.
Yet metastability can make transitions between conformational states rare, producing strongly correlated trajectories and limiting effective sampling~\citep{perez2025breaking,OLSSON2026103213}.

Boltzmann generators (BGs)~\citep{noe2019boltzmann} offer an alternative by generating independent molecular configurations from a learned proposal distribution with tractable likelihoods, enabling importance reweighting to the target Boltzmann distribution.
Recent work has extended this approach to transferable models that sample multiple molecular systems~\citep{klein2024transferable}.
Unlike protein structure prediction~\citep{Jumper2021}, equilibrium sampling requires capturing the relative statistical weights of different configurations, and scaling transferable BGs to larger and more diverse systems remains challenging.
Early BGs relied primarily on normalizing flows~\citep{tabak2013family,dinh2016density,rezende2015variational,chen_neural_2018}, which model atomic coordinates in continuous space~\citep{klein2023equivariant,kohler2020equivariant,tan2026scalableequilibriumsamplingsequential,tan2026amortizedsamplingtransferablenormalizing}.
More recently, however, \citet{rehmann2025autoregressive} demonstrated state-of-the-art peptide sampling using autoregressive models of discretized atomic coordinates.

These developments motivate three questions:
\textbf{(1)} How does sampling quality of autoregressive BGs scale with model size, training data, and training compute?
\textbf{(2)} What conditioning representation supports training and sampling across diverse molecular systems and environments?
\textbf{(3)} How do sampling quality and computational efficiency change as the number of atoms increases?

We introduce \namelong (\nameshort), a family of autoregressive BGs that represent molecular environments, chemical graphs, and atomic coordinates as discrete tokens.
This representation enables conditioning through text and coordinate prefixes within a shared interface (Figure~\ref{fig:dbg-ar-generation}), without the separate atom-wise conditioning encoder used by \citet{rehmann2025autoregressive}.
The autoregressive factorization preserves tractable proposal likelihoods, supporting importance reweighting.

Across models trained on \textit{Many Peptides}, we characterize empirical scaling relationships linking sampling quality to model size, training compute, atom count, and coordinate discretization resolution.
Guided by these relationships, we scale \nameshort to 640M parameters, achieving state-of-the-art effective sample sizes as well as energy and torsion-angle accuracy on unseen peptide systems containing up to 157 atoms.
Beyond improving sampling quality, we accelerate autoregressive generation using speculative decoding~\citep{leviathan2023fastinferencetransformersspeculative}, achieving speedups of up to $2.14\times$ over standard autoregressive decoding, with particular benefits for larger systems, \rebuttal{and compare to MD simulation estimates over time}.
\rebuttal{We demonstrate the flexibility of prefix conditioning by training PBG-Joint across molecular classes and environmental conditions.
PBG-Joint remains competitive with peptide-only models on \textit{Many Peptides} (Table~\ref{tab:sota}) while also enabling sampling of different solvent environments and small molecules, conditioned on amino acid sequences and SMILES.}
Finally, we apply the learned proposals to solvation free-energy estimation, extending their use from conformational sampling to thermodynamic property estimation.

Our main contributions are:
\begin{itemize}[topsep=0pt, partopsep=0pt, itemsep=0pt, parsep=0pt, leftmargin=*]
    \item We introduce \namelong (\nameshort), autoregressive molecular samplers that use a shared discrete-token representation for text and coordinate prefix conditioning while retaining tractable proposal likelihoods.
    \item We characterize empirical scaling relationships for sampling quality and use them to guide scaling to 640M parameters, achieving state-of-the-art effective sample sizes as well as energy and torsion-angle accuracy on \textit{Many Peptides}. We further accelerate generation using speculative decoding.
    \item We demonstrate joint training across molecular classes and environments and apply \nameshort to solvation free-energy estimation and conditional microsolvation of systems containing more than 280 atoms.
\end{itemize}

\section{Background}
\label{sec:background}

\paragraph{Equilibrium Sampling of Molecular Conformations.}
\phantomsection\label{par:boltzmann-distribution}
Consider an atomistic system of interest with the descriptor $c$, $n$ atoms, and a conformation $\mathbf{x}=(\mathbf{x}_1,\ldots,\mathbf{x}_n)\in\Omega\subseteq\mathbb{R}^{n\times3}$, where $\Omega\subseteq\mathbb{R}^{n\times3}$ is the configuration space. At equilibrium and fixed temperature, the probability density of a conformation $\mathbf{x}$ is given by the Boltzmann distribution

\begin{equation}
  p(\mathbf{x})
  = \frac{1}{Z_{\mathbf{x}}}
    \exp\!\left[-\beta U(\mathbf{x})\right],
  \qquad
  Z_{\mathbf{x}}
  = \int_{\Omega}
    \exp\!\left[-\beta U(\mathbf{x})\right]
    \,\mathrm{d}\mathbf{x},
  \label{eq:configurational-boltzmann-distribution}
\end{equation}

where $U(\mathbf{x})$ is the potential energy, $\beta=(k_{\mathrm B}T)^{-1}$ is the inverse thermal energy, and $Z_{\mathbf{x}}$ is the configurational partition function, an integral over the configuration space. Given a test function $\phi(\mathbf{x})$, i.e.\ a physicochemical property of interest, we can estimate its value by forming the expectation.
The required samples may be generated by simulating molecular dynamics, or by drawing independent samples from a learned proposal.

\paragraph{Boltzmann Generators.}
Boltzmann Generators are generative models which learn a parameterized proposal $q_{\theta}$ that approximates the Boltzmann distribution $p$ and admit a tractable exact likelihood~\cite{noe2019boltzmann}. Training typically minimizes $\KL(p\|q_{\theta})$ on previously sampled equilibrium conformers, but can also be performed using potential energy supervision alone via minimization of one or more $\KL(q_{\theta}\|p)$ objectives~\cite{vonklitzing2025learningboltzmanngeneratorsconstrained,neal2001annealed,noe2019boltzmann}. A transferable Boltzmann Generator shares one set of parameters $\theta$ across many systems and realizes each proposal as a conditional $q_{\theta}(\mathbf{x}\mid c)$~\cite{klein2024transferable}. Once trained, independent samples $\mathbf{x}_i\sim q_{\theta}$ can then be drawn in parallel, and self-normalized importance sampling (SNIS) enables reweighting of samples to target the intended Boltzmann distribution.

\begin{equation}
  \tilde{w}(\mathbf{x})
  =\frac{e^{-\beta U(\mathbf{x})}}{q_{\theta}(\mathbf{x})},
  \qquad
  \log\tilde{w}(\mathbf{x})
  =-\beta U(\mathbf{x})-\log q_{\theta}(\mathbf{x}),
  \qquad
  w_i
  =\frac{\tilde{w}(\mathbf{x}_i)}{\sum_{j=1}^{N}\tilde{w}(\mathbf{x}_j)}.
  \label{eq:bg-snis}
\end{equation}
For a test function $\phi$, the SNIS estimator is then $\E_{p}[\phi]\approx\sum_{i=1}^{N} w_i\,\phi(\mathbf{x}_i)$.
The variance of this estimator is controlled by the overlap of $q_{\theta}$ and $p$.

A common approach to train Boltzmann Generators are Normalizing Flows~\cite{Tabak2010DENSITYEB,dinh2016density,tan2026scalableequilibriumsamplingsequential}.
Flow-based proposals evaluate exact likelihoods through the change-of-variables formula, requiring invertible transformations with tractable Jacobian determinants.
However, these requirements constrain architectural choices and can make distributions with well-separated modes difficult to represent~\cite{pmlr-v119-cornish20a}.

\paragraph{Autoregressive Boltzmann Generators.}
Recently \citet{rehmann2025autoregressive} introduced Autoregressive Boltzmann Generators (ARBG), an alternative approach to Boltzmann Generators based on the autoregressive prediction of discretized atom coordinates.
We consider ARBGs to be part of a family of models we will refer to as \textbf{Discretized Boltzmann Generators}, composed of five largely independent components:
(1) A surjective, and non-injective \textit{Quantizer} $\kappa_{\Delta}$, which maps a continuous conformation $\mathbf{x}$ to a voxel code $\mathbf{k}\in\mathcal{A}_b^{n\times 3}$; (2) a bijective \textit{Tokenizer}, which serializes that code into a length-$m$ sequence of categorical symbols $\mathbf{z}=\tau(\mathbf{k})$; (3) a deterministic \textit{Decoding Order} $\sigma\in S_m$, which determines the order in which those $m$ symbols are to be generated; (4) a \textit{Discrete Generative Model} $q_{\theta,\Delta}$ that learns the proposal distribution over token sequences; and (5) a \textit{Dequantizer} kernel $\mathcal{R}_{\Delta}$, which assigns a voxel code $\mathbf{k}$ to a conditional distribution over continuous coordinates $\mathbf{x}$. We discuss each component in further detail in Appendix~\ref{sec:dbg}.

\looseness=-1
Briefly, ARBGs rely on a quantizer to partition a cube of side $L$, in which molecules are sampled, into $b$ equal-width cells of size $\Delta=L/b$ along each axis. The scalar map $\kappa_{\Delta}$ sends each axis coordinate to the index of the unique cell that contains it, resulting in a voxel code for the entire conformation $\mathbf{k}=(k_{id})\in\mathcal{A}_b^{n\times 3}$.
Physically, each voxel code defines a macrostate of the discretized distribution $p_{\Delta}$ which represents the actual learning target of the generative model.
Intuitively, as $\Delta\to 0$, the voxel probability per unit volume approaches the continuous Boltzmann density: $p_{\Delta}(\kappa_{\Delta}(\mathbf{x}))/\Delta^{3n}\to p(\mathbf{x})$.

\looseness=-1
The learned proposal is a categorical distribution $q_{\theta,\Delta}(\mathbf{z}\mid c,\sigma)$ on the token sequence $\mathbf{z}=\tau(\mathbf{k})$, conditioned on the system descriptor $c$ and a fixed decoding order $\sigma$. In the case of ARBGs this decoding order is derived directly from a predefined atom order $\pi$ by emitting each atom's 3
discretized spatial coordinates consecutively, yielding
$m=3n$ coordinate tokens.
Flattening along $\sigma$ gives
\begin{equation}
  q_{\theta,\Delta}(\mathbf{z}\mid c,\sigma)
  =\prod_{j=1}^{m}
  q_{\theta,\Delta}\bigl(z_{\sigma(j)}\mid z_{\sigma(<j)},c\bigr).
  \label{eq:dbg-autoregressive}
\end{equation}

Once a discretized proposal is sampled and mapped into a voxel code $\mathbf{k}$, a reconstruction kernel $\mathcal{R}_{\Delta}$ (e.g.\ a \textit{Uniform Reconstruction Kernel}) returns a continuous conformation $\mathbf{x}$ within the encoded volume of space to evaluate test functions of interest $\phi(\mathbf{x})$ and to estimate importance weights using the potential energy $U(\mathbf{x})$.

Efficient training methods and widespread adoption make autoregressive models a promising basis for scaling Boltzmann generators towards larger and more diverse molecular systems.
This requires large, chemically diverse training corpora, but current ARBG conditioning schemes to construct the conditioning information $c$ limit this flexibility.

In ROBIN, the transferable ARBG in \citet{rehmann2025autoregressive}, an encoder $\eta_\theta$ maps hand-designed features $F(G,\pi)$ of a chemical graph $G=(V,E)$ with $n=|V|$ atoms to atom embeddings,
$C_G=\eta_\theta(F(G,\pi))\in\mathbb{R}^{n\times d}$, whereby the ordering of atoms $\pi$ itself is given by a graph-based heuristic $\chi_{\mathrm{intra}}(G)$ \cite{ambertools}.
These embeddings are added to the corresponding coordinate-token representations and used as cross-attention keys and values before the causal decoder.

This current design imposes three restrictions.
First, one embedding per atom makes conditioning length grow with system size, even when a compact descriptor is available, such as a protein sequence or SMILES~\cite{smiles}.
Second, the complete atom set and its ordering must be specified before sampling, preventing, for example, generation of a variable number of explicit solvent molecules.
Third, extension beyond pre-specified chemistries requires changes to the hand-designed feature schema and its encoder.

\section{Prefix-Conditioned Boltzmann Generators}

\paragraph{Conditioning and prompt construction.}
We encode the molecular system and its environment as a tokenized text prefix $c$.
Extending the graph conditioning used in ARBGs, this prefix specifies the molecular graph, solvent environment, potential energy model, and temperature.
Environment and molecular specifications are separated by special delimiter tokens, with a final separator marking the start of the coordinate sequence (Table~\ref{tab:joint-prompt-examples}).

Molecular graphs are represented using the Hierarchical Editing Language for Macromolecules (HELM)~\citep{helmnotation}.
HELM supports linear, branched, and cyclic biopolymers, including chemical modifications.
Connectivity annotations specify links between monomers, while inline SMILES~\citep{smiles} with labeled attachment points describe residues outside the standard monomer vocabulary.
Small molecules are represented using SMILES within HELM chemical components.
This provides a shared textual interface for biopolymers, modified residues, and small molecules, without a separate atom-wise conditioning encoder.

\begin{table}[htbp]
  \centering
  \caption{\textbf{Example System Prompts.} Shown are representative molecules from public and newly generated datasets and their system prompts. \texttt{[BOS]} and \texttt{[SEP]} abbreviate the special delimiter tokens between environment, HELM, and coordinate blocks.}
  \label{tab:joint-prompt-examples}
  \begingroup
  \edef\prompttablescale{\fpeval{min(1,\textwidth/\catalogmaxwd)}}
  \normalfont\normalsize
  \setlength{\tabcolsep}{3pt}
  \renewcommand{\arraystretch}{1.25}
  \newcommand{\prompttext}[1]{\texttt{#1}}
  \sbox{\catalogprobebox}{%
  \begin{tabular}{@{}lll@{}}
    \toprule
    System & Source & Prompt \\
    \midrule
    CGSWHKQR & Many Peptides &
    \prompttext{[BOS] 310K,water,ff14sb-obc1-2nm [SEP] PEPTIDE1\{C.G.S.W.H.K.Q.R\}\$\$\$\$ [SEP]} \\
    DEMT & Vacuum Peptides &
    \prompttext{[BOS] 300K,vacuum,ff14sb [SEP] PEPTIDE1\{D.E.M.T\}\$\$\$\$ [SEP]} \\
    HYGW + 128 waters & Microsolvated Peptides &
    \prompttext{[BOS] 300K,water,ff14sb-tip3p-rf2nm [SEP] PEPTIDE1\{H.Y.G.W\}\$\$\$\$ [SEP]} \\
    cyclo-AYTW & Many Cycles &
    \prompttext{[BOS] 300K,water,ff14sb-obc1-1nm [SEP] PEPTIDE1\{A.Y.T.W\}\$PEPTIDE1,PEPTIDE1,1:R1-4:R2\$\$\$ [SEP]} \\
    Molecule 00621 & MDQM9-nc &
    \prompttext{[BOS] 300K,vacuum,gaff2-am1bcc [SEP] CHEM1\{[C[C@@H](CO)CC\#N]\}\$\$\$\$ [SEP]} \\
    2,2,4-Trimethylpentane & FreeSolv (vacuum) &
    \prompttext{[BOS] 300K,vacuum,gaff2-am1bcc [SEP] CHEM1\{[CC(C)CC(C)(C)C]\}\$\$\$\$ [SEP]} \\
    Bromotrifluoro-methane & FreeSolv (implicit water) &
    \prompttext{[BOS] 300K,water,gaff2-am1bcc-obc1-2nm [SEP] CHEM1\{[FC(F)(F)Br]\}\$\$\$\$ [SEP]} \\
    Phenyl formate + 128 waters & FreeSolv (microsolvated) &
    \prompttext{[BOS] 300K,water,gaff2-am1bcc-tip3p-rf2nm [SEP] CHEM1\{[O=COc1ccccc1]\}\$\$\$\$ [SEP]} \\
    \bottomrule
  \end{tabular}
  }%
  \scalebox{\fpeval{min(\prompttablescale,\linewidth/\wd\catalogprobebox)}}{\usebox{\catalogprobebox}}
  \endgroup
\end{table}

\looseness=-1
\paragraph{Variable length coordinate generation.}
After the final \texttt{[SEP]} token, the model predicts discretized
Cartesian coordinates autoregressively, with three coordinate tokens
per atom. Termination differs between nonsolvated and explicitly
solvated systems. For nonsolvated systems, training sequences end with
an \texttt{[EOS]} token. If $z_{1:m}$ is a complete sequence of
$m=3n$ coordinate tokens, its probability is
\[
q_\theta(z_{1:m},\texttt{[EOS]}\mid c)
=
\left[\prod_{t=1}^{m}q_\theta(z_t\mid c,z_{<t})\right]
q_\theta(\texttt{[EOS]}\mid c,z_{1:m}).
\]
\looseness=-1
For explicitly solvated systems, however, training sequences contain no
\texttt{[EOS]} token. At inference, the number of atoms to generate
is specified externally, and sampling stops after the corresponding
number of coordinate tokens is reached, allowing, for example, for sampling of a variable number of solvent molecules.

\looseness=-1
\paragraph{Supramolecular atom ordering.}
Autoregressive Boltzmann generators generate coordinate tokens
according to an atom order $\pi$.
We obtain the order within each molecule from an existing graph-based heuristic
$\chi_{\mathrm{intra}}$ \cite{rdkit}.
To generalize the canonicalization of atom orders to supramolecular systems, such as explicitly solvated small molecules, $G=G_0\sqcup G_1\sqcup\cdots\sqcup G_K$, comprising a reference solute $G_0$ and $K$ identical solvent molecules, we rely on the indistinguishability of solvent molecules and define
$\chi_{\mathrm{inter}}(G,\mathbf{x})=\rho$ as the permutation
that orders, for example, waters by increasing distance from their oxygen
atom to the nearest solute atom: $d_{\rho(1)}\le\cdots\le d_{\rho(K)}$ with $d_j=\min_{i\in V(G_0)}\|\mathbf{x}_{o_j}-\mathbf{x}_i\|_2$, where $o_j$ denotes the oxygen atom of water $j$.
The complete atom order places the solute first, followed by
the ordered solvent molecules: $\pi=\chi(G,\mathbf{x})
  =\chi_{\mathrm{intra}}(G_0)
  \,\Vert\,
  \chi_{\mathrm{intra}}(G_{\rho(1)})
  \,\Vert\,\cdots\,\Vert\,
  \chi_{\mathrm{intra}}(G_{\rho(K)})$.
This coordinate-dependent rule selects an ordering of indistinguishable solvent molecules in the
training conformations. We found this greatly reduced collisions during sampling as compared to random ordering during preliminary experimentation with discrete diffusion~\citep{shi2025simplifiedgeneralizedmaskeddiffusion,sahoo2024simpleeffectivemaskeddiffusion}. At inference, we generate waters autoregressively without enforcing
increasing solute--water distance. This leads to a framework where during training we canonicalize atom and molecular order, and softly encourage global rotation and translation equivariance through data augmentation.

\section{Experiments}
\subsection{Increasing Sampling Quality via Model Scaling}

\paragraph{Baselines.} We reproduce the transferable Autoregressive Boltzmann Generator ROBIN~\cite{rehmann2025autoregressive} as the closest method to our work. In addition, we also compare with published results using autoregressive normalizing flows, Prose~\citep{tan2026amortizedsamplingtransferablenormalizing} and TarFlow~\citep{zhai2024normalizing}.

\paragraph{Metrics.} \phantomsection\label{par:metrics} Following prior work, we evaluate sampling quality using the stand-alone relative Effective Sample Size (\rESS)~\cite{kish1957confidence}, as well as three Wasserstein metrics describing similarities between sampled and reference conformations: (a) The 2-Wasserstein distance of the dimensionless reduced potential energy (\EW), where $\mathcal{E}(\mathbf{x})\equiv\beta U(\mathbf{x})$, (b) the periodicity-aware 2-Wasserstein distance (\TW) of the backbone torsion angles, and (c) the 2-Wasserstein distance along the first two components (\TICAW) of a time-lagged independent component analysis (TICA).
In contrast to recent work \cite{rehmann2025autoregressive}, we re-emphasize the \rESS as a metric, given its ability to provide a necessary but not sufficient indicator of sampling quality in contexts in which no reference samples are available. To increase the computational throughput and generality towards non-peptide molecules, we also introduce a sliced 2-Wasserstein distance of \textit{all} torsion angles of a given molecule (\TgSWast) (Appendix \ref{metrics:sliced}).
For scaling model fitting we evaluate the previously described metrics without the common 0.2\% clipping of importance weights. We mark these unclipped metrics with a superscript, yielding \rESSast, \EWast and \TgSWast.

\paragraph{Sampling Quality Scales with Compute and Decreases with System Size.}
\phantomsection\label{par:scaling}

\begin{figure}[htbp]
  \centering
  \includegraphics[width=\linewidth]{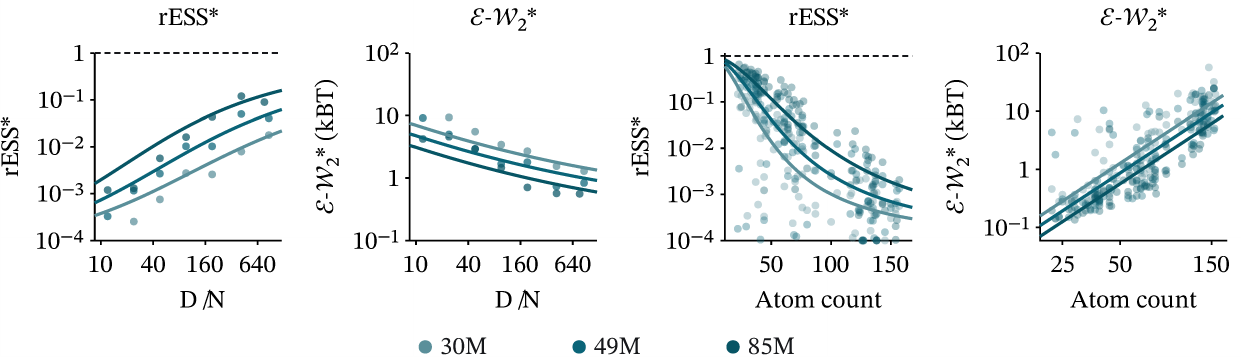}
  \caption{
  \textbf{Sampling Quality Scales with Compute and Decreases with System Size.} Left: Median \rESSast and \EWast on \textit{Many Peptides} test systems versus training budget in loss tokens per non-embedding parameter ($D/N$). Right: per-system \rESSast and \EWast versus atom count for representative models at $D/N=192$.}
  \label{fig:dipper-d-metric}
\end{figure}

In order to assess the ability for Prefix-Conditioned Boltzmann Generators to scale towards higher sampling quality and large atomistic system sizes, we investigated whether sampling performance could be anticipated with a scaling model covering model size, compute, system atom count and discretization resolution.
To this end, we trained a series of \fittingcohortsize autoregressive PBGs with 8M-85M non-embedding parameters ($N$) on the \textit{Many Peptides} and assessed training compute budgets from $D/N=12$ to $D/N=1760$ loss tokens per non-embedding parameter, where $D$ counts loss tokens, using a Warmup-Stable-Decay schedule~\cite{wsdscalinglaw} approach.
\rebuttal{In an ablation ladder from the ROBIN~\cite{rehmann2025autoregressive} training protocol to PBG at 30M, 85M and 160M (Appendix~\ref{sec:robin-pbg}, Table~\ref{tab:robin-pbg-ladder}), most of the sampling quality gain at similar loss tokens came from an improved training protocol, while prefix conditioning showed no penalty against dedicated atom feature embeddings.}

\begin{table}[htb]
  \caption{\textbf{Evaluation results on \textit{Many Peptides}.} Shows means across the test systems listed in the headers, with $10^4$ samples each. \rebuttal{The 8AA means (\P) exclude YFPHAGYT across literature derived \textit{and} observed measurements, due to stereochemical artifacts in the \textit{Many Peptides} data (Appendix~\ref{sec:yfphagyt}).} The first section of values are literature derived, following sections are measured. Measured values are means for three sampling seeds. \rebuttal{The largest deviation is shown in $\pm$ in units of the last decimal place. Each measured row is a single trained checkpoint. A set of model training replicates are shown in Appendix~\ref{sec:seed-variation}.} Best values in \textbf{bold}, with second-best \underline{underlined}. $D/N$ next to a model name gives the number of loss tokens per non-embedding parameter for that model. $T$ denotes the logit temperature.}
  \label{tab:sota}
  \centering
  \input{generated/sota_plot.tex}
\end{table}

Once trained, the \rESSast, \EWast and \TgSWast evaluation metrics on the \textit{Many Peptides} test systems and the terminal training loss were modeled using a shared factor scaling model based on~\cite{rosenfeld2021scalinglawsdeeplearning, kaplan2020scaling} via Bayesian Inference (Figure \ref{fig:dipper-d-metric})~\cite{phan2019composable, ruan2024observational}.
Briefly, a latent factor based on the common form $G = A\,(N/10^{9})^{-\alpha} + B\,(D/10^{9})^{-\beta}$
was estimated to model the training loss and three evaluation metrics jointly.
We tied exponents ($\alpha=\beta$) given the limited set of observations and previous results of approximately proportional compute-optimal scaling of model size and training data in language modeling~\cite{hoffmann2022training}.
Each metric $m$, including the loss, was fitted with an \emph{adapter} that maps $G$ and the atom count $n$ to the observed metric value through a shared functional form:
\begin{equation}
  Y_m = a_m + p_m \log\!\left(\frac{n}{n_0}\right) + k_m \log G + \delta_m[b],
  \qquad \delta_m[4096] = 0,
  \label{eq:adapter-general}
\end{equation}
where $a_m$ is the intercept, $p_m$ the atom-scaling coefficient, $k_m$ the $G$-sensitivity, and $\delta_m[b]$ is a per bin-size offset.
For the NLL, we set $p_{\mathrm{NLL}} = 0$ and thereby reduce the adapter to $G$ plus a bin offset $\delta_m[b]$ and an irreducible error $a_{\mathrm{NLL}}$, in line with previous scaling models~\cite{rosenfeld2021scalinglawsdeeplearning}.
For \rESSast, the output $-\log(\rESSast)$ is additionally bounded by the finite sample count via a sigmoid: $-\log(\rESSast) = C \cdot \mathrm{sigmoid}(Y_{\rESS})$ (Appendix \ref{par:bayesian}).

We estimated atom-count coefficients $p_E=\DipperScalingEnergyAtomExponent{}$ for $\log(\EWast)$ and $p_R=\DipperScalingrESSAtomExponent{}$ within the sigmoid model of $-\log(\rESSast)$.
At fixed training resources and discretization, energetic error grows approximately quadratically with atom count.
\rebuttal{This trend does not depend on the specification of the scaling model: across all specifications and refits, $p_E$ ranged from \DipperScalingAblationEnergyAtomMin{} to \DipperScalingAblationEnergyAtomMax{} (median \DipperScalingAblationEnergyAtomMedian), and also with per-system offsets it remained near-quadratic ($p_E=\DipperScalingAblationSystemEnergyAtom$, 95\% CI \DipperScalingAblationSystemEnergyAtomLow--\DipperScalingAblationSystemEnergyAtomHigh, Appendix~\ref{par:scaling-sensitivity}).}
Similarly, outside of saturated \rESSast regimes, we observe a near-quadratic growth of $-\log(\rESSast)$, implying a rapid loss of effective samples as systems grow.
In contrast, $\log(\TgSWast)$, $p_m = \DipperScalingSlicedAtomExponent{} \approx 1$ shows linear scaling with system size (Figure \ref{fig:dipper-torus-scaling}) \rebuttal{and is robust across different scaling model specifications and refits (median \DipperScalingAblationSlicedAtomMedian, ranged from \DipperScalingAblationSlicedAtomMin{} to \DipperScalingAblationSlicedAtomMax{}, Appendix~\ref{par:scaling-sensitivity}).}

\begin{wrapfigure}{R}{0.75\textwidth}
  \centering
  \includegraphics[width=\linewidth]{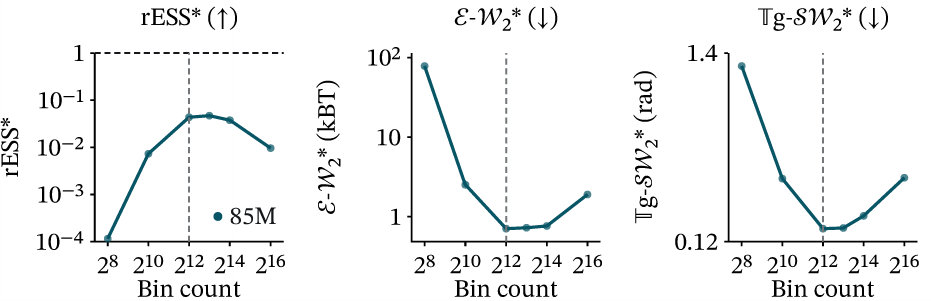}
  \caption{\textbf{Discretization Resolution Effect on Evaluation Metrics.} Sampling Quality versus discretization bin count and resolution. Shown are results for PBGs trained on a resolution grid of $2^{8}$--$2^{16}$ bins. Points are observed per-checkpoint medians of \textit{Many Peptides} test systems at $D/N=192$.
  Dashed line represents $2^{12}$ bin reference.}
  \label{fig:dipper-bin-panels}
\end{wrapfigure}
\looseness=-1

To understand the degree to which the discretization of atomic coordinates alone leads to an increase in energetic error, we analyzed the expected energy bias caused by midpoint reconstruction of quantized conformers (Appendix~\ref{sec:extensive-msf}).
In the fine-bin regime, midpoint reconstruction induces an
expected energy bias proportional to the atom count $\mathbb E[\Delta U^{\mathrm{mid}}]\propto n\Delta^2$ which is thereby lower-bounding the unweighted roundtrip energy Wasserstein error, $\EWast\ge\beta|\mathbb E[\Delta U^{\mathrm{mid}}]|$.
Experiments with quantization--reconstruction \textit{roundtrips} of equilibrium \textit{Many Peptides} conformers support this result, suggesting that the steep decay of sampling quality metrics with atom count is in part due to discretization-induced error.

\rebuttal{During the training of transferable Boltzmann generators, compute can be allocated towards increased \textit{system diversity} (i.e.\ a larger number of distinct chemical systems) or \textit{simulation depth} (i.e.\ a larger and better equilibrated set of conformational states per system).
To estimate the effects of either factor, we ran ablation experiments in which PBGs (30M, 49M and 85M) were trained at $D/N=48$ on $1/d$ of the \textit{Many Peptides} training frames, kept either as the first $200/d$\,ns of every trajectory or as $1/d$ of the training systems with their full trajectories.
Both arms used the same number of frames and the same compute at each $d$.
Shortening the trajectories to 6.25\,ns per system changed sampling quality little, whereas keeping 681 or fewer of the 21{,}700 training systems collapsed sampling quality on unseen tetra- and octapeptides (Appendix~\ref{sec:data-fraction}).
Within the studied budget and evaluation context, sampling quality is therefore limited by system diversity rather than by individual simulation depth.}

Overall, the \rebuttal{empirical} scaling model implies a compute-optimal ratio of $D^\ast/N=\DipperScalingDstarTN$ (95\% CI \DipperScalingDstarTNLow--\DipperScalingDstarTNHigh) (Table \ref{tab:joint-scaling}). \rebuttal{Across alternative specifications of the scaling model and refits, this estimate ranged widely from \DipperScalingAblationTppCoreMin{} to \DipperScalingAblationTppCoreMax{} (median \DipperScalingAblationTppCoreMedian, Appendix~\ref{par:scaling-sensitivity}), however, far below the $D/N\approx1100$ of ROBIN~\citep{rehmann2025autoregressive}.
\textit{MDQM9-nc}~\cite{mdqm9} gave related estimates of $D^\ast/N$ (Appendix~\ref{sec:scaling-mdqm9}).}
Based on this \rebuttal{allocation rule}, we proceeded to train larger Prefix-Conditioned Boltzmann Generators, spanning 160M, 320M, and 640M parameters on the \textit{Many Peptides} dataset.
In line with the model, we observed strong results on the \textit{Many Peptides} test set (Table \ref{tab:sota} and Appendix Figure~\ref{fig:dipper-compute-metrics}).
PBGs scored best across all peptide-length combinations, with the exception of Prose's results for tetrapeptide \TICAW.
\rebuttal{At a matched budget of $D/N=48$, sampling quality metrics, including \rESS, increase with model size at every peptide length (Appendix~\ref{sec:seed-variation}).}

\paragraph{Discretization Resolution Has a Model-Dependent Optimum.}
Besides varying model parameters and training data, we also investigated the effect of varying the number of discretization bins $b$, and thus the spatial resolution $\Delta$ with which conformers are sampled. In contrast with previous reports~\cite{rehmann2025autoregressive}, we observed an optimum of discretization depth with both low ($b\le 2^{10}=1024$ bins, $\Delta\ge 3.2\ \mathrm{pm}$) and high ($b\ge2^{13}=8192$ bins, $\Delta\le 0.4\ \mathrm{pm}$) spatial resolution leading to reduced sampling quality. These observations are in line with a two-regime model, where both too few discretization bins (quantization-limited regime) and too many discretization bins (vocabulary-limited regime) lead to reduced sampling quality.
\par
\WFclear

\subsection{Increasing Sampling Rate via Speculative Decoding}
\paragraph{Speculative Decoding Increases Sampling Efficiency.}
\begin{wrapfigure}{R}{0.5\textwidth}
  \vspace{-\intextsep}
  \setlength{\parskip}{0pt}
  \centering
  \includegraphics[width=\linewidth]{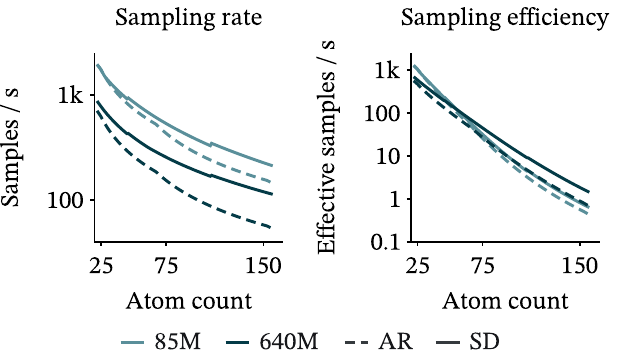}
  \setlength{\abovecaptionskip}{3pt}
  \caption{\textbf{Sampling Rate and Efficiency.} Sampling rates (left) and \rESSast{}-weighted sampling efficiency (right) for 85M and 640M PBGs. Dashed: autoregression (AR); solid: best AR or speculative decoding (SD) with a 15M drafter.}
  \label{fig:sota-sampling-comparison}
  \vspace{-\intextsep}
\end{wrapfigure}
In practice, the utility of a sampling method, its sampling efficiency, is a function of both the sampling quality and the sampling rate on available hardware.
Previously, autoregressive Boltzmann generators with discretized coordinates were shown to produce samples at a lower rate than existing flow-based alternatives~\cite{rehmann2025autoregressive}.

To improve the sampling rate of autoregressive Boltzmann generators while maintaining exact base-model likelihoods, we performed speculative decoding with an existing 15M PBG draft model~\cite{leviathan2023fastinferencetransformersspeculative}.
We sampled peptides at varying length and fixed batch size of 512 on an H100 NVL from representative, state-of-the-art PBGs.
The 85M and 640M models achieved mean observed speedups relative to standard autoregressive sampling of $\DipperSpecEightyFiveMMeanSpeedup\times$ and $\DipperSpecSixFortyMMeanSpeedup\times$, with maxima of $\DipperSpecEightyFiveMMaxSpeedup\times$ and $\DipperSpecSixFortyMMaxSpeedup\times$, respectively (Figures \ref{fig:sota-sampling-comparison} and \ref{fig:dipper-core-eff-speed-atoms}).

\begin{figure}[t]
  \centering
  \includegraphics[width=\linewidth]{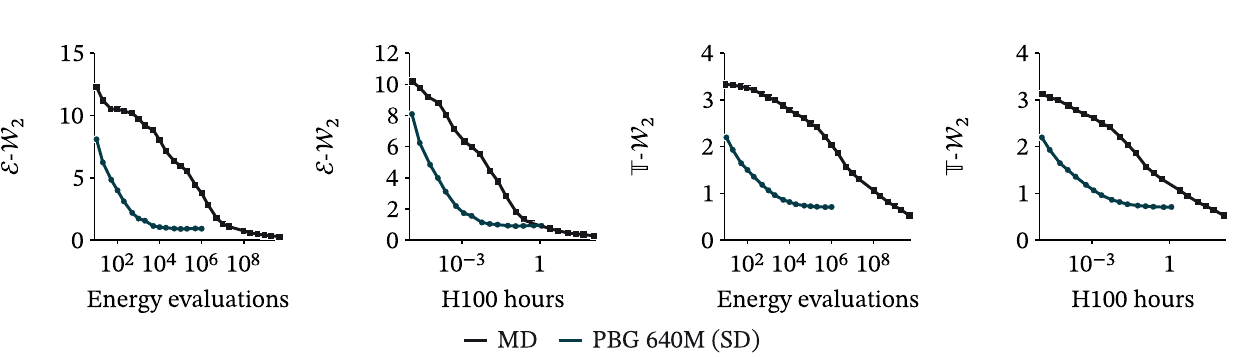}
  \caption{\rebuttal{\textbf{Sampling Efficiency against Molecular Dynamics.} \EW and \TW of MD and PBG 640M against energy evaluations and H100-hours. Mean across \textit{Many Peptides} test systems.}}
  \label{fig:cost-vs-md-main}
\end{figure}

\rebuttal{Timed on the same hardware, estimates based on PBG 640M with speculative decoding converged after about $10^5$ proposals (0.12 H100-hours per system), whereas MD required about $10^9$ steps (30 to 35 H100-hours) to reach the same \TW and \TICAW, and about one H100-hour for an equivalent \EW estimate (Figure~\ref{fig:cost-vs-md-main}, Appendix~\ref{sec:cost-vs-md}).}

\looseness=-1
Despite these improvements in sampling rate alone, we generally notice a steep decline in overall sampling efficiency of Boltzmann generators as atom count increases: As both sampling rate and sampling quality decline with atom count independently, Boltzmann generators' sampling efficiency drops by a factor of ca.\ $10^3$ as system size increases by a factor of ca.\ $2^3$. These results highlight the need to improve both sampling
quality and sampling rate as Boltzmann generators are extended to larger systems.

\subsection{\rebuttal{Joint Training across Molecular Classes and Environments}}

\paragraph{Prefix-Conditioned Boltzmann Generators sample across chemistries and solvent environments.}

In order to estimate the properties of a material, for example a therapeutic small molecule, its Boltzmann distribution needs to be sampled across different environments. For example, the estimation of solvation free energies requires conformer samples in vacuum and water.

To enable solvation free energy estimation using Boltzmann Generators, we aimed to extend their transferability towards small molecules and across solvent environments.
We trained a Prefix-Conditioned Boltzmann Generator on a collection of public and newly generated datasets including \textit{Many Peptides}, \textit{MDQM9-nc}~\cite{mdqm9} (non-cyclic small molecules in vacuum), \textit{Vacuum Peptides} (peptides from \textit{Many Peptides} in vacuum) and \textit{FreeSolv} (small molecules sourced from \citet{mobley_guthrie_2014_freesolv} in vacuum and implicit solvent). The resulting model was termed PBG-Joint (160M, $D/N=198$).

PBG-Joint achieved strong performance on \textit{Many Peptides} (Table~\ref{tab:sota}), \rebuttal{equivalent to a peptide-only model with the same 160M architecture and matched training compute (Appendix Table~\ref{tab:joint-matched-control}).}
Sampling quality across small molecules varied with chemical complexity and was generally equivalent to dedicated, fine-tuned models (Appendix Table~\ref{tab:mdqm9-native} and \phantomsection\label{ref:fig:dipper-mdqm9-metric-atoms}Figure~\ref{fig:dipper-mdqm9-metric-atoms}).

\begin{figure}[htbp]
  \centering
  \includegraphics[width=\linewidth]{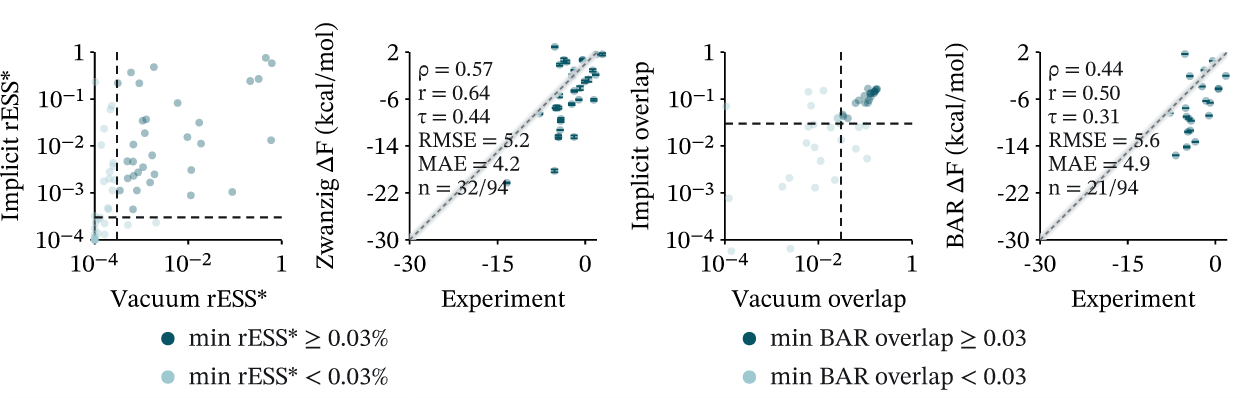}
  \caption{\textbf{Solvation Free Energy Estimation using Prefix-Conditioned Boltzmann Generators.} \rebuttal{FreeSolv test molecules versus experimental measurements. Left pair: forward Zwanzig estimates without reference MD samples, with per-state forward \rESSast. Inclusion requires \rESSast $\ge0.03\%$ in both states.
  Right pair: BAR estimates with reference MD samples of both states, with per-state BAR overlaps.
  Inclusion requires both BAR overlaps $\ge0.03$.
  Free energies are in kcal/mol.}}
  \label{fig:asfe-dipper}
  \label{fig:asfe-dipper-freesolv}
  \label{fig:asfe-forward-freesolv}
\end{figure}

Boltzmann Generators enable solvation free-energy estimation through the Zero Free Energy Proposal cycle~\cite{ding2021deepbar} (Appendix~\ref{par:zero-free-energy-proposals}).
\rebuttal{We assessed the model's ability to generate solvation free energy estimates for previously unseen test peptides and FreeSolv small molecules (Appendix~\ref{sec:asfe-estimation}).}
BAR estimates reached overlaps $\ge0.03$ in both states for \AsfeQualifiedCount{} of \AsfeTotalCount{} peptide systems~\cite{Klimovich2015GuidelinesFT} and \FreeSolvAsfeQualifiedCount{} of \FreeSolvAsfeTotalCount{} FreeSolv test systems.
In this subset, estimates correlated with experimental solvation free energy measurements ($r=\AsfeFreesolvPearson$) for small molecules within the \textit{FreeSolv} dataset.
\rebuttal{Of note, also direct Zwanzig estimation using PBG-Joint, which does not depend on reference MD samples, yielded comparable estimates for the FreeSolv dataset ($r=\FreeSolvZwanzigPearson$, Figure~\ref{fig:asfe-dipper}).}
Mean absolute errors against experiment were \FreeSolvBarMae{} (BAR) and \FreeSolvZwanzigMae\,kcal/mol (Zwanzig), against \FreeSolvMdReferenceMae\,kcal/mol for an MD-only reference, indicating that parts of the error are attributable to force field and implicit solvation bias (Appendix~\ref{par:asfe-cutoff}).

\begin{figure}[H]
  \centering
  \includegraphics[width=\linewidth]{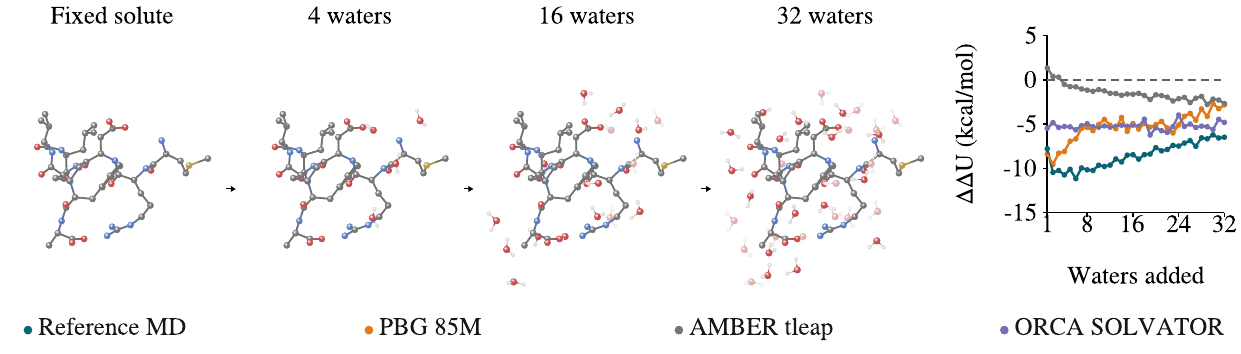}
  \addtolength{\abovecaptionskip}{-4mm}
  \caption{\textbf{Autoregressive microsolvation of an octapeptide.} Left: Microsolvation of 4, 16, and 32 waters around a fixed MRDPVLFA conformer in the original generation order by a PBG. Pale red color marks earlier waters. Right: Median energy change per added water for all 14 net-neutral octapeptide systems. Curves show adjacent differences of cohort-median $U(k)-U(0)$ in kcal/mol.}
  \label{fig:solvation-generation}
\end{figure}

Compared to implicit solvation, solvation free energy estimation approaches that sample explicit solvent-solute interactions are considered the state-of-the-art \cite{openfe}. We therefore prepared an additional \textit{Microsolvated Peptides} and \textit{FreeSolv} microsolvation dataset, containing conformers of solutes and their 128 nearest water molecules cropped from short explicit solvent MD trajectories run in $1\,\mathrm{nm}$ solvent-padded boxes with periodic boundary conditions.
To calculate potential energies for such a cropped system during inference time we resorted to a reaction field approach. Notably, this introduced a mismatch between the underlying data generating potential and the inference time potential, leading to collapsed effective sample sizes of trained models.
Given these constraints, we assessed a dedicated PBG model's ability to sample favorable intermolecular interactions with a microsolvation task, in which a variable number of 32--80 water molecules were placed around fixed solute conformers from explicit-solvent MD \rebuttal{(net-neutral peptides only, due to absence of counterions in solvent clusters)}.
For evaluation, we measured total potential energy and wall-clock time.
We compared our approach against existing water placement tools, ORCA's Solvator \cite{ORCA,orca6solvator} and Amber's tleap \cite{ambertools}, and observed lower minimum and median energies, as well as faster microsolvation for octapeptides relative to these baselines (Table \ref{tab:solvator-comparison} and Figure \ref{fig:solvation-generation}).

\begin{table}[H]
  \centering
  \caption{\textbf{Microsolvation Task.} For each test net-neutral peptide, 512 distinct $300\,\mathrm{K}$ solute conformers are held fixed and solvated by different methods. \rebuttal{Columns show the mean across systems. The largest deviation is shown in $\pm$ in units of the last decimal place.}}
  \label{tab:solvator-comparison}
  \input{generated/dipper_solvated_solvator_comparison.tex}
\end{table}

\section{Conclusion}
In this work we define Prefix-Conditioned Boltzmann Generators, a flexible extension of autoregressive Boltzmann generators.
We demonstrate their ability to learn transferable proposals across peptides, small molecules, solvent environments and towards microsolvated states.
These proposals enable initial solvation free energy estimations with just one generative model.
\rebuttal{Within the study's parameter range and data, Prefix-conditioned Boltzmann generators demonstrate scaling behavior which indicates a more parameter-intensive compute-optimal training at $D^\ast/N\approx\DipperScalingDstarTN$, compared to previous works \cite{rehmann2025autoregressive}, and an overall steep scaling of training compute requirement to maintain performance metrics, such as energetic error, as system size increases \CstarEW $\approx n^{\DipperScalingComputeParityExponent}$
at fixed $b=4096$, with an exponent above \DipperScalingAblationParityMin{} in every specification with a fitted atom-count exponent (Appendix~\ref{par:scaling-sensitivity}).}
Despite methods such as speculative decoding, both sampling quality \textit{and} sampling rate decay with atom count, leading to strong reductions in overall sampling efficiency as BGs are applied to larger systems.
Identifying sampling methods which can maintain the statistical properties of Boltzmann Generators while scaling more favorably remains an open research problem.

\paragraph{Limitations.}
The presented study is constrained by a range of experimental limitations, which include the scaling model and free energy estimation accuracy.
The presented scaling model \rebuttal{spans a limited number of model sizes, molecule categories, and} has a limited set of extrapolating training runs which together are not sufficient to draw generalized conclusions in the form of a \textit{scaling law} for autoregressive Boltzmann generators.

The sampling quality of small molecule and vacuum conformational ensembles is reduced compared to the \textit{Many Peptides} reference results, which is also reflected in solvation free energy estimates. We attribute this to a set of factors, which include \textbf{(1)} limited size of newly sampled datasets compared to established baselines (Table \ref{tab:datasets}), \textbf{(2)} reduced electrostatic screening in vacuum environments which might increase the energetic consequences of conformational errors in samples, and \textbf{(3)} challenges in prompting PBGs with small molecule sequence representations such as SMILES without pre-training on large, diverse small molecule data to learn this representation itself.

We anticipate the development of future methods which can improve sampling quality in such challenging, yet highly relevant, contexts, e.g.\ via iterative updating of decoded atom coordinates.

\begin{ack}
The authors would like to thank Charlie Tan for regular, in-depth discussion, in particular for identifying the box-volume bias of the importance weights under constrained translation augmentation and its correction (Appendix~\ref{sec:box-volume}).
The authors would also like to thank Zander Blasingame, Sam McCallum and Luka Mucko for discussions related to the project.
The authors acknowledge a compute grant by the EuroHPC joint undertaking on the LUMI-G (Reference EHPC-BEN-2026B08-112) and Arrhenius cluster (Reference EHPC-AIF-2026SC04-111), an allocation by the Austrian Scientific Computing (ASC) infrastructure via project ÖAW-MUSICA-2026, as well as compute resources provided by AITHYRA (\url{https://www.oeaw.ac.at/aithyra}).
\end{ack}

\clearpage

\subsection*{AI use statement}
In this work, we used generative AI tools to assist with writing code to run the experiments and analyses, searching the literature for related work, formulating mathematical claims, and text editing to improve clarity.
We have not used generative AI tools for generating synthetic data sets.
We take responsibility for the final content of this work, including text, claims and artifacts produced.

\subsection*{Reproducibility statement}
\publictext{We release the code required to reproduce the experimental results on GitHub under \href{https://github.com/NiklasTR/pbg}{\texttt{niklastr/pbg}}.}{We will release the code required to reproduce the experimental results on GitHub upon acceptance.}
Model architectures, training settings and evaluation procedures are described in the appendix.
\publictext{The trained model checkpoints are available on Hugging Face under \href{https://huggingface.co/niklastr/pbg}{\texttt{niklastr/pbg}}, and the sampled rollouts under \href{https://huggingface.co/datasets/niklastr/pbg_rollouts}{\texttt{niklastr/pbg\_rollouts}}.}{The trained model checkpoints and sampled rollouts will be released on Hugging Face upon acceptance.}
\publictext{Novel datasets and their Hugging Face repositories are described below.}{Novel datasets are described below and will be released on Hugging Face upon acceptance.}

{\small
\color{black}
\bibliographystyle{iclr2027_conference}
\bibliography{references}
}

\clearpage
\appendix

\section{Additional Background}
\paragraph{Free Energy Estimation.}
Consider two clearly identifiable states of our atomistic system of interest, for example the same molecule exposed to vacuum ($A$) and to water ($B$), with potentials $U_A$ and $U_B$. At equilibrium and fixed temperature, their free-energy difference is defined by the implied partition functions $Z_A$ and $Z_B$
\begin{equation}
  \Delta F_{A\to B}
  = F_B-F_A
  = -\beta^{-1}\log\!\left(Z_B/Z_A\right).
  \label{eq:free-energy-difference}
\end{equation}
When both states are accessible, this corresponds to the log-odds of the system occupying state $A$ over state $B$: $P(A)/P(B)=e^{\beta\Delta F_{A\to B}}$, a direct correspondence to the Boltzmann odds that neighboring microstates receive in~\eqref{eq:configurational-boltzmann-distribution}, with $F$ in place of $U$~\cite{zuckerman2010statistical}.

Multiple approaches to free energy estimation exist~\cite{lelievre2010freeenergycomputation}. Zwanzig's exponential averaging formula estimates the free energy difference from samples of a single state~\cite{zwanzig1954high}. It is the one-sided importance-sampling estimator of $\log(Z_B/Z_A)=-\beta\Delta F_{A\to B}$, with unnormalized densities $\tilde p_B=e^{-\beta U_B}$ and $\tilde p_A=e^{-\beta U_A}$ and raw importance weight $e^{-\beta\Delta U}$. The forward estimator draws $\mathbf{x}\sim p_A$ and the reverse estimator draws $\mathbf{x}\sim p_B$,
\begin{equation}
  \begin{aligned}
    \Delta F_{A\to B}
    &= -\beta^{-1}\log\E_{\mathbf{x}\sim p_A}\!\left[\exp \bigl(-\beta\Delta U(\mathbf{x})\bigr)\right] \\
    &= +\beta^{-1}\log\E_{\mathbf{x}\sim p_B}\!\left[\exp \bigl(+\beta\Delta U(\mathbf{x})\bigr)\right],
  \end{aligned}
  \label{eq:zwanzig}
\end{equation}
where $\Delta U(\mathbf{x})=U_B(\mathbf{x})-U_A(\mathbf{x})$. Jensen's inequality implies that a finite-sample reverse estimate is biased low and a finite-sample forward estimate is biased high,
$\E[\widehat{\Delta F}^{\mathrm{rev}}]\le\Delta F_{A\to B}\le\E[\widehat{\Delta F}^{\mathrm{fwd}}]$.

Bennett's acceptance ratio (BAR) combines samples from both states
through optimal bridge sampling~\cite{bennett1976efficient,meng1996simulating}.
The Boltzmann densities and the earlier free-energy relation give
$p_B(\mathbf{x})/p_A(\mathbf{x})
=e^{\beta[\Delta F_{A\to B}-\Delta U(\mathbf{x})]}$.
Pooling $n_A$ samples from $A$ and $n_B$ from $B$, a logistic function
converts these density ratios into probabilities of sample origin.
BAR estimates $\Delta F_{A\to B}$ by fitting a single parameter
$\delta$ through logistic maximum likelihood, where $p_{\delta}(B\mid\mathbf{x})$
models the probability of origin $B$ given $\mathbf{x}$:
\begin{equation}
  p_{\delta}(B\mid\mathbf{x})
  = \sigma\!\left(
      \beta[\delta-\Delta U(\mathbf{x})]
      + \log\frac{n_B}{n_A}
    \right),
  \qquad \sigma(t)=(1+e^{-t})^{-1}.
  \label{eq:bar}
\end{equation}

\paragraph{Zero Free Energy Proposals.}
\phantomsection\label{par:zero-free-energy-proposals}
Consider a Boltzmann Generator trained as a proposal $q_{\theta,A}$ for one physical state $A$, with potential $U_A$ and Boltzmann density $p_A$. Unconventionally, we can treat the proposal $q_{\theta,A}$ as a standalone thermodynamic state $A^{\circ}$ with potential $U_{A^{\circ}}(\mathbf{x})=-\beta^{-1}\log q_{\theta,A}(\mathbf{x})$~\cite{ding2021deepbar}. Because $q_{\theta,A}$ is normalized, this proposal state has unit partition function and therefore zero free energy $F_{A^{\circ}} = -\beta^{-1}\log Z_{A^{\circ}} = 0$.

\begin{wrapfigure}{r}{0.54\linewidth}
  \centering
  \begin{tikzpicture}[
    black,
    font=\footnotesize,
    >=Stealth,
    node distance=11mm and 26mm,
    state/.style={
      draw, rounded corners=2pt, minimum width=8mm, minimum height=6mm,
      inner sep=1.2pt, line width=0.6pt, align=center
    },
    arr/.style={-{Stealth}, line width=0.65pt},
    target/.style={arr, dashed, dash pattern=on 2.4pt off 1.6pt},
    elabel/.style={font=\footnotesize, inner sep=1.2pt, fill=white}
  ]
    \node[state] (A) {$A$};
    \node[state, right=of A] (B) {$B$};
    \node[state, below=of A] (Ao) {$A^{\circ}$};
    \node[state, below=of B] (Bo) {$B^{\circ}$};
    \draw[target] (A) -- node[elabel, above] {$\Delta F_{A\to B}$} (B);
    \draw[arr] (A) -- node[elabel, left] {$\Delta F_{A\to A^{\circ}}$} (Ao);
    \draw[arr] (B) -- node[elabel, right] {$\Delta F_{B\to B^{\circ}}$} (Bo);
    \draw[arr] (Ao) -- node[elabel, below] {$\Delta F_{A^{\circ}\to B^{\circ}}=0$} (Bo);
  \end{tikzpicture}
  \caption{\textbf{Zero-Free-Energy Proposal Cycle.} Normalized generative proposal states $A^{\circ}$ and $B^{\circ}$ have vanishing free energy, so $\Delta F_{A^{\circ}\to B^{\circ}}=0$.}
  \label{fig:zfep-cycle}
\end{wrapfigure}
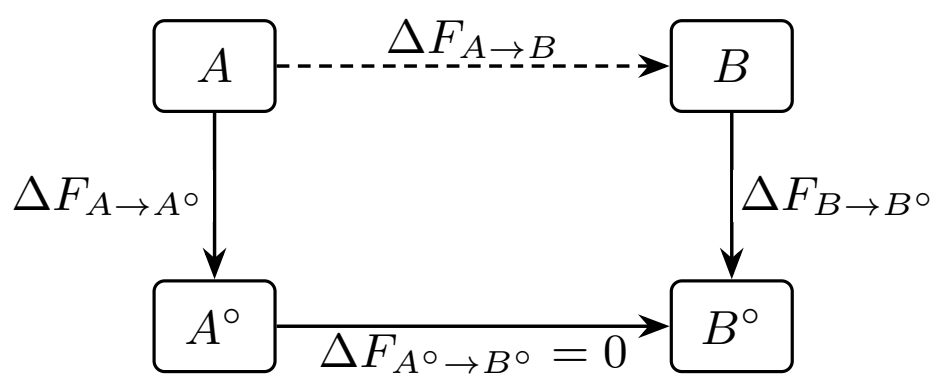
Sampling $\mathbf{x}\sim q_{\theta,A}$ is sampling the proposal state $A^{\circ}$, and the unnormalized importance weights routinely estimated during SNIS~\eqref{eq:bg-snis} can be directly used within a free energy estimator, such as the Zwanzig method, yielding the absolute free energy of $A$ due to the proposal state $A^{\circ}$'s zero free energy. A free energy difference between two physical states, such as vacuum ($A$) and water ($B$), now reduces to two such independent proposal-to-target absolute free energy estimates (Figure~\ref{fig:zfep-cycle}). Each physical state has its own learned proposal state, accessible via its corresponding $q_{\theta}$, and the proposal-to-proposal leg vanishes, $\Delta F_{A^{\circ}\to B^{\circ}}=0$.
The resulting cycle identity is $\Delta F_{A\to B}=\Delta F_{A\to A^{\circ}}-\Delta F_{B\to B^{\circ}}$, where each leg can be estimated by the forward or reverse Zwanzig formula, or by BAR, depending on the availability of samples from the physical state. In summary, the Zero Free Energy Proposal approach enables straightforward free energy estimation using Boltzmann Generators.

\newpage
\section{Discretized Boltzmann Generators}
\label{sec:dbg}

\paragraph{Quantization.}
\label{sec:uniform-quantization}

The coordinates of an atomistic system similar to the one described in Section~\ref{par:boltzmann-distribution} can be coarse-grained through a quantization of each spatial dimension. We write each atomic coordinate as $\mathbf{x}_i=(x_{i1},x_{i2},x_{i3})$ with $i\in\{1,\ldots,n\}$ and $d\in\{1,2,3\}$.
An interval $I=[a,a+L)$ along each axis, with origin $a$ and width $L$, contains the coordinates to be discretized. In practice this width may match the cubic periodic box of a molecular simulation.
The interval is partitioned into $b\ge 1$ equal cells of width $\Delta=L/b$,
\begin{equation}
  \Delta=\frac{L}{b},
  \qquad
  I_l=[a+l\Delta,a+(l+1)\Delta),
  \qquad
  l\in\mathcal{A}_b=\{0,\ldots,b-1\}.
  \label{eq:uniform-quantization-cells}
\end{equation}
The scalar quantizer is $\kappa_{\Delta}(x)=l$ iff $x\in I_l$. Applied independently to each Cartesian component, it yields the voxel code $\mathbf{k}=(k_{id})\in\mathcal{A}_b^{n\times 3}$ with $k_{id}=\kappa_{\Delta}(x_{id})$.

\label{sec:quantized-canonical}
Each code $\mathbf{k}$ is a voxel macrostate: $C(\mathbf{k})=\prod_{i=1}^{n}\prod_{d=1}^{3}I_{k_{id}}$. Its configurational partition function, free energy, and equilibrium probability are
\begin{equation}
  Z_{\Delta}(\mathbf{k})
  =\int_{C(\mathbf{k})}
  \exp[-\beta U(\mathbf{x})]
  \,\mathrm{d}\mathbf{x},
  \qquad
  F_{\Delta}(\mathbf{k})
  =-\beta^{-1}\log Z_{\Delta}(\mathbf{k}),
  \qquad
  p_{\Delta}(\mathbf{k})
  =\frac{Z_{\Delta}(\mathbf{k})}{Z_{\mathbf{x}}}.
  \label{eq:quantized-canonical-pmf}
\end{equation}
The cells partition the continuous configuration space $\Omega$, so $Z_{\mathbf{x}}=\sum_{\mathbf{k}}Z_{\Delta}(\mathbf{k})$. In practice, a sample from $p_{\Delta}$ is obtained by drawing $\mathbf{x}\sim p$ and setting $k_{id}=\kappa_{\Delta}(x_{id})$.

\paragraph{Tokenization.}
\label{sec:tokenisation-principle}
A tokenizer $\tau$ is a bijection from a voxel code $\mathbf{k}$ to a length-$m$ sequence $\mathbf{z}=\tau(\mathbf{k})$. As such, it does not change the cells, but only serializes them into a sequence.

\textit{Dimensional tokenization} emits the three axis indices $(k_{i1},k_{i2},k_{i3})$ as separate symbols, so $m=3n$. Each symbol is a bin index on one Cartesian axis, and the alphabet size equals the number of cells $b$ on that axis of the cube. This approach is used within this study and \citet{rehmann2025autoregressive}.

\textit{Octree tokenization} serializes the same cube by nested octants. At each step the current cell is split into eight equal children, and the symbol records which child contains the atom. A sequence of such octants is a path from the cube to a leaf voxel. With $T$ symbols per atom, $m=Tn$. The leaf grid matches dimensional tokenization when the path depth matches the degree of axis binning.

\paragraph{Decoding Order.}
\label{sec:graph-automorphisms-notation-anchor}

Given a tokenization, we are required to select at least one decoding order $\sigma$. We choose a pre-defined atom-inserting decoding order following an atom order $\pi$ of the chemical graph of the system $G$.
Because the space of possible decoding orders is large, we constrain it empirically. Based on previous work \cite{gebauer2019symmetry,cheng2025scalable} and physical intuition, we aim to decode atoms in an order that is spatially local.
Intuitively, atoms that are close to each other within the decoding order should be more likely to be close in coordinate space: During training, we place the solute first, followed by solvent molecules ordered by $\rho=\chi_{\mathrm{inter}}(G,\mathbf{x})$.
\textit{Within} a given molecule ($\chi_{\mathrm{intra}}$) we rely on the ordering induced by a molecular graph canonicalization heuristic \cite{rdkit, cheng2025scalable} to arrive at an ordering.

Once a decoding order $\sigma$ has been defined, each atom within the graph order has to be assigned to a 3D coordinate for training. We define $\operatorname{Aut}(G)\le S_n$ as the atom permutations that preserve bonds and labels. If $G$ is a disjoint union of molecular components, this group splits into intra-component automorphisms and permutations of isomorphic components. Each $\gamma\in\operatorname{Aut}(G)$ acts on a frame by row permutation $\mathbf{x}\mapsto\mP_\gamma\mathbf{x}$ and leaves the unlabeled conformer unchanged and, for a symmetry-invariant potential, its Boltzmann density unchanged.

\textit{Within} a given molecule, equivalent atom labels can be permuted among coordinate positions (e.g.\ terminal methyl-group rotamers). We restrict ourselves to independent terminal orbits of $\operatorname{Aut}(G)$. During training we either sample these orbits, or break them with an RDKit canonicalization heuristic \cite{rdkit}. In practice we have found the active re-sampling of atom assignment order in rotamer orbits to be inconsequential.

\phantomsection\label{sec:intermolecular-indistinguishability}
\textit{Between} identical molecules, we select their training order with $\rho=\chi_{\mathrm{inter}}(G,\mathbf{x})$. For water, this permutation satisfies $d_{\rho(1)}\le\cdots\le d_{\rho(K)}$, where $d_j=\min_{i\in V(G_0)}\|\mathbf{x}_{o_j}-\mathbf{x}_i\|_2$ and $o_j$ denotes the oxygen atom of water $j$. Thus, waters are ordered by increasing distance from their oxygen to the nearest solute atom. The complete atom order is $\pi=\chi(G,\mathbf{x})=\chi_{\mathrm{intra}}(G_0)\,\Vert\,\chi_{\mathrm{intra}}(G_{\rho(1)})\,\Vert\,\cdots\,\Vert\,\chi_{\mathrm{intra}}(G_{\rho(K)})$.

\input{generated/alanine_dipeptide_two_waters_tikz}

\paragraph{Dequantization.}
\label{sec:midpoint-cubature}
A reconstruction kernel $\mathcal{R}_{\Delta}$ maps a voxel $\mathbf{k}$ to a continuous conformation in that cell, so that test functions $\phi(\mathbf{x})$ and the potential $U(\mathbf{x})$ can be evaluated. We consider uniform sampling within a cell and its deterministic midpoint $r_l=a+(l+\tfrac{1}{2})\Delta$, the conditional expectation of the former, applied independently to each Cartesian component,
\begin{equation}
  \mathcal{R}_{\Delta}^{\mathrm{mid}}(\mathrm{d}y\mid l)
  \equiv\delta_{r_l}(\mathrm{d}y),
  \qquad
  \mathcal{R}_{\Delta}^{\mathrm{unif}}(\mathrm{d}y\mid l)
  \equiv\frac{\1_{y\in I_l}}{\Delta}\,\mathrm{d}y.
  \label{eq:scalar-reconstruction-kernels}
\end{equation}

The two kernels provide estimates of the same cell integral $Z_{\Delta}(\mathbf{k})$. Under uniform reconstruction, $e^{-\beta U(\mathbf y)}$ is an unbiased one-sample Monte Carlo estimate of $Z_{\Delta}(\mathbf{k})/\Delta^{3n}$. In contrast, the midpoint kernel acts as a zeroth-order cubature at the cell center,
\begin{equation}
  F_{\Delta}(\mathbf{k})
  \approx
  U(\mathbf{r}_{\mathbf{k}}) - \beta^{-1}\,3n\log\Delta,
  \label{eq:midpoint-cubature}
\end{equation}
which is approximate but has zero reconstruction variance conditional on $\mathbf{k}$. The log-volume term is constant across $\mathbf{k}$ and cancels in self-normalized ratios for a fixed system.

\newpage
\section{Related Work}

\paragraph{Discretized Representations of Biomolecules.}
Discretizing biomolecular structure is an established approach to improve the sampling of and search over molecular conformations, albeit primarily for proteins and at considerable coarse graining.
Historic examples include simple 2D lattice protein models confining residue positions to discrete sites~\cite{lau1989lattice, sali1994how}, 3D fragment libraries~\cite{kolodny2002small} and basic, learned structural motifs~\cite{debrevern2000bayesian}.
Recently, learned vector-quantized codebooks of protein conformation~\cite{oord2017neural, vankempen2024fast} have been identified, which have also been paired with discrete generative protein models~\cite{hayes2025simulating}, albeit without any goal of matching the Boltzmann distribution or accessible likelihoods. While the presented work relies on a relatively fine-grained quantization to reduce the error during importance weighting, more coarse-grained and learned quantization methods for PBGs are conceivable, although challenges related to non-uniformly sized quantization cells and high variance during the macrostate free energy estimation need to be anticipated.

\paragraph{Autoregressive Generation of Biomolecule Conformers.}
Autoregressive models have primarily sampled 3D structure of biomolecules via iterative atom insertion. Early work, such as \textit{G-SchNet}, inserted atoms sequentially into continuous space using learned pairwise, internal, distance distributions~\cite{gebauer2019symmetry}. More recent work applied similar approaches to the sampling of global coordinates~\cite{cheng2025scalable}. In contrast, octree-based methods~\cite{lu2025uni3dar} autoregressively denoise atom locations with a final step mapping atom coordinates into continuous space. In contrast to these works, Autoregressive Boltzmann Generators~\cite{rehmann2025autoregressive} and the family of Prefix-Conditioned Boltzmann Generators we describe operate over discrete coordinate representations and rely on parameter-free quantization and dequantization methods.

\paragraph{Generative Modeling for Free Energy Estimation.}
Boltzmann generators have previously been used for single-system free energy estimation using the zero free energy proposal approach, including for Guest-Host binding free energy~\cite{ding2021deepbar}. Recently, this approach was extended to transferable Boltzmann generators, and applied to implicit solvation free energy estimations of small molecules~\cite{yu2026card}. Our work is closely related to~\cite{yu2026card}, and primarily differs by choosing a purely discrete coordinate modeling, a simple prefix conditioning based prompt, atom insertion over atom denoising, random $SE(3)$ augmentations to achieve learned equivariance instead of conformer pose canonicalization, and by moving towards explicit solvation.
Alternative generative modeling approaches to facilitate free energy estimation include Boltzmann emulators which sample high quality starting conformers for free energy simulations, but lack accessible and exact likelihoods~\cite{bioemu}. Recently, such emulators, which can be referred to as alchemical Boltzmann emulators, have been extended to sample alchemical states, too, thereby further accelerating alchemical free energy simulations~\cite{bengio2026aquagen}.

\clearpage
\section{Additional Results}
\label{sec:additional-results}

\subsection{Scaling Model on Many Peptides}

\begin{wrapfigure}{R}{0.36\textwidth}
  \centering
  \includegraphics[width=0.6834\linewidth]{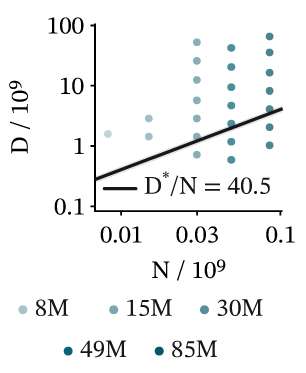}
  \caption{\textbf{Scaling Model Data.} Points show evaluated $D$ and $N$ combinations for 8M--85M models with 4096 coordinate bins. The compute-optimal scaling model estimate $D^\ast(N)$ is shown as a line.}
  \label{fig:dipper-compute-optimal}
\end{wrapfigure}

\paragraph{Scaling Data.}
In order to study the relationship of PBG model size and training compute with sampling quality and system size, a series of \fittingcohortsize models were trained. These models spanned \fittingmodelsizecount distinct model sizes between 8M and 85M parameters, ca.\ \fittingtokenbudgetcount training compute budget ranges spanning $D/N=12$ to $D/N=1760$, as well as \fittingbinsizecount distinct bin counts / discretization resolutions from \(b=256\), \(\Delta=12.8\,\mathrm{pm}\) to \(b=65{,}537\), \(\Delta\approx0.05\,\mathrm{pm}\).
All models were trained with \textit{Many Peptides} data using a Warmup-Stable-Decay schedule for amortized estimation of training compute effects. Once trained, 91 molecular systems were evaluated per checkpoint. Next to the terminal training loss, the three previously introduced, unclipped and sliced metrics were estimated (Table~\ref{tab:sota-modern}) yielding a final dataset of 130 checkpoints with  $91 \times3 + 1$ values each.

\paragraph{Scaling Model.}
Our goal is to fit a simplified scaling model using Bayesian inference.
Starting from the general scaling law $G=A N^{-\alpha}+B D^{-\beta}$ \cite{rosenfeld2021scalinglawsdeeplearning}, we tie the two exponents $\alpha=\beta$ for modeling stability and based on the assumption of approximate equivalence~\cite{kaplan2020scaling}, yielding $G=A N^{-\alpha}+B D^{-\alpha}$.
Next to the exponent equivalence assumption, we rely on the assumption that training compute $C$ is proportional to model size and loss tokens seen $C \approx kND$, which is often simplified to $C=6ND$.
Given these assumptions, our primary interest is the compute-optimal ratio of $r=D^\ast/N^\ast$. Preferably, we would like to directly place a prior over this ratio.

We proceed to rewrite the general scaling law to directly estimate the compute-optimal ratio. At fixed compute, we must choose how to allocate resources between model size $N$ and training tokens $D$ to minimize $G$. Using $C\approx kND$, we rewrite:
\[
D=\frac{C}{kN},
\qquad
G(N\mid C)
=\frac{A}{N^\alpha}
+\frac{B}{(C/(kN))^\alpha}
=\frac{A}{N^\alpha}
+B\left(\frac{k}{C}\right)^\alpha N^\alpha.
\]
Increasing the model parameter size $N$ reduces the first term but increases the second. To find the minimum at fixed compute, we differentiate $G$ with respect to $N$:
\[
\frac{\partial G}{\partial N}
=-\alpha A N^{-\alpha-1}
+\alpha B\left(\frac{k}{C}\right)^\alpha N^{\alpha-1}
=0
\qquad \text{at } N=N^\ast.
\]
Rearranging and substituting $D^\ast=C/(kN^\ast)$ gives
\[
\frac{A}{(N^\ast)^\alpha}
=\frac{B}{(D^\ast)^\alpha}.
\]
Inserting $r=D^\ast/N^\ast$, we obtain
\begin{equation}
r=\left(\frac{B}{A}\right)^{1/\alpha}.
  \label{eq:scaling-optimal-ratio}
\end{equation}

Knowing the compute-optimal ratio $r$ does not determine both coefficients $A$ and $B$. Therefore we choose to define $G_0=2\sqrt{AB}$. Using $B=Ar^\alpha$, we obtain
\[
G_0=2Ar^{\alpha/2},
\qquad
A=\frac{G_0}{2}r^{-\alpha/2},
\qquad
B=\frac{G_0}{2}r^{\alpha/2}.
\]
Substituting these expressions into the scaling law gives a rewritten equivalent form, dependent on $G_0$, $r$, and $\alpha$:
\begin{equation}
\boxed{G=\frac{G_0}{2}
\left[
(N\sqrt{r})^{-\alpha}
+(D/\sqrt{r})^{-\alpha}
\right]}.
  \label{eq:scaling-resource-law}
\end{equation}
Here, $G_0$ controls the overall scale of the resource-dependent $G$,
$r$ specifies the compute-optimal allocation, and $\alpha$ controls
how quickly $G$ decreases as resources increase.

\begin{table}[H]
  \caption{\textbf{Sampling with Unclipped Importance Weights.} Median metrics across \textit{Many Peptides} test peptides, with $10^4$ samples per system. Measured values are means for three sampling seeds. The largest deviation is shown in $\pm$ in units of the last decimal place. Best values are \textbf{bold}; second-best values are \underline{underlined}.}
  \label{tab:sota-modern}
  \centering
  \input{generated/sota_modern_plot.tex}
\end{table}

\paragraph{Model Specification.}

Consider a checkpoint with index $i$ and a system with index $j$. Let $N_i$ be the non-embedding parameter count, $D_i$ the delivered loss-token count, $b_i$ the bin count, and $n_j$ the atom count.
\[
N_{9i}=N_i/10^9,\quad D_{9i}=D_i/10^9,\quad x_j=\log(n_j/65)
\]
The terminal training loss $L_i$ and the transformed metrics $Y_{mij}$ are defined as
\[
Y_{Eij}=\log\!\left({\EWast}_{ij}\right),\quad Y_{Sij}=\log\!\left({\TgSWast}_{ij}\right),\quad Y_{Rij}=-\log\!\left({\rESSast}_{ij}\right)
\]
Here $m\in\{E,S,R\}$ indexes \EWast, \TgSWast, and \rESSast, respectively. The transformed metrics are drawn from a Student's $t$-distribution, with a sigmoid applied to \rESSast given its upper ($=1$) and lower ($=1/n_{\mathrm{sample}}$) bounds.

\begin{align*}
L_i&\sim t_{\nu_L}(\mu_{Li},\sigma_L),\\
Y_{mij}&\sim t_{\nu_m}(\mu_{mij},\sigma_m),\\
\mu_{Eij}&=\eta_{Eij},\quad \mu_{Sij}=\eta_{Sij},\\
\mu_{Rij}&=\log(10^4)\operatorname{sigmoid}(\eta_{Rij}).
\end{align*}

The latent variables $\eta$ are dependent on $G_i$ via a set of adapters for each transformed metric. $E$ denotes the irreducible training loss and $a_m$ an analogous metric specific intercept. The covariate $p_m$ captures the scaling of the transformed metric with atom count, while $k_m$ does the equivalent for $G$. A $\delta_m$ intercept represents discretization resolution dependent effects.

\begin{equation}
\begin{aligned}
  G_i&=\frac{G_0}{2}\left[(N_{9i}\sqrt{r})^{-\alpha}+(D_{9i}/\sqrt{r})^{-\alpha}\right],\\
\mu_{Li}&=E+G_i+\delta_L[b_i],\\
\eta_{mij}&=a_m+p_m x_j+k_m\log G_i+\delta_m[b_i].
\end{aligned}
  \label{eq:scaling-model-adapters}
\end{equation}

\begingroup\small
\noindent\begin{tabular}{p{0.25\linewidth}p{0.69\linewidth}}
\toprule
Parameter & Prior \\
\midrule
$E$ & $\mathcal N(3.55,0.15^2)$ \\
$\log G_0$ & $\mathcal N\!\left(\log\!\left(2\sqrt{0.003\cdot0.05}\right),\frac{1.5^2}{2}\right)$ \\
$r$ & $\mathcal N^{+}(20,10)$, truncated to $r>0$ \\
$\log\alpha$ & $\mathcal N(\log 0.8,0.75^2)$ \\
$\delta_L[b]$, $b\ne4096$ & $\mathcal N(0,3^2)$ \\
$\delta_m[b]$, $b\ne4096$ & $\mathcal N(0,1^2)$, $m\in\{E,S,R\}$ \\
$a_E,\ a_S,\ a_R$ & $\mathcal N(5,2^2),\quad\mathcal N(1,2^2),\quad\mathcal N(4.5,2^2)$ \\
$p_E,\ p_S,\ p_R$ & $\mathcal N(2,0.5^2),\quad\mathcal N(0.7,0.5^2),\quad\mathcal N(1.2,0.8^2)$ \\
$k_E,\ k_S,\ k_R$ & $\mathcal N(1.2,0.8^2),\quad\mathcal N(0.7,0.8^2),\quad\mathcal N(1.2,0.8^2)$ \\
$\log\sigma_L$ & $\mathcal N(\log 0.01,1^2)$ \\
$\log\sigma_E,\ \log\sigma_S$ & $\mathcal N(\log 0.5,0.75^2)$ for each parameter \\
$\log\sigma_R$ & $\mathcal N(\log 1.5,0.75^2)$ \\
$\log(\nu_m-2)$ & $\mathcal N(\log 10,0.75^2)$, $m\in\{L,E,S,R\}$ \\
\bottomrule
\end{tabular}
\endgroup

A generalized model posterior was sampled via MCMC. In order to weight the model towards accuracy of high-compute checkpoints, likelihoods were tilted by relative compute intensity.
\[
w_i=\frac{(6N_iD_i)^{1/2}}{\frac{1}{130}\sum_{k=1}^{130}(6N_kD_k)^{1/2}}.
\]
\[
\pi_w(\theta\mid\mathcal D)\propto p(\theta)
\prod_{i=1}^{130}\left[
t_{\nu_L}(L_i;\mu_{Li},\sigma_L)
\prod_{j=1}^{91}\prod_{m\in\{E,S,R\}}
t_{\nu_m}(Y_{mij};\mu_{mij},\sigma_m)
\right]^{w_i}.
\]

\paragraph{Model Estimates.}
The scaling model matched the terminal training NLL with high accuracy, and explained ca.\ 60--70\% of variance in the transformed evaluation metrics (Table~\ref{tab:joint-scaling-r2}).
Among these evaluation metrics, \rESSast had the strongest coupling to the model loss, followed by \EWast and \TgSWast (Table~\ref{tab:joint-scaling}).
Both \EWast and \rESSast presented a near-quadratic scaling with increases in atomic system size, while \TgSWast showed linear scaling.

\paragraph{Comparison with Extensive Scaling.}
A natural reference for the atom-count coefficients $p_m$ is extensive scaling. If a molecule consists of weakly coupled parts and the model is equally accurate for every atom, the errors add up over the atoms: the KL divergence between the Boltzmann density and the proposal, the variance of the log importance weights and the mean energy bias all grow in proportion to $n$. For log-normal importance weights, $-\log\rESS=\operatorname{Var}[\log w]$, and the energy Wasserstein distance is at least the absolute mean energy bias, $\EWast\ge\beta\lvert\E[\Delta U]\rvert$. Extensive scaling therefore corresponds to $p_m=1$ in Equation~\ref{eq:scaling-model-adapters}. It is a special case of our adapters, so the fitted $p_m$ test it directly.
Two error sources follow this case: the torsion error ($p_S=\DipperScalingSlicedAtomExponent$, Figure~\ref{fig:dipper-torus-scaling}) and the energy error that quantization alone causes, which grows linearly with $n$ in theory and in roundtrip experiments (Appendix~\ref{sec:extensive-msf}). The energy error and \rESSast{} of generated samples grow faster ($p_E=\DipperScalingEnergyAtomExponent$, $p_R=\DipperScalingrESSAtomExponent$), so the accuracy per atom of the generator itself decreases as systems grow.
In line with this result, a refit which fixes $p_E=p_R=1$ (row F3, Appendix~\ref{par:scaling-sensitivity}) has the largest held-out errors and underpredicts \EWast{} for long peptides.

\begin{table}[htbp]
  \centering
  \edef\pairedfigurescale{\fpeval{\textwidth/(210mm)}}%
  \edef\pairedtablescale{\fpeval{min(1,\textwidth/\catalogmaxwd)}}%
  \renewcommand{\resizebox}[3]{%
    \makebox[\linewidth][c]{%
      \scalebox{\pairedtablescale}{\normalfont\normalsize #3}}%
  }
  \begin{minipage}[t]{0.44\linewidth}
  \vspace{0pt}
  \caption{\textbf{Scaling Model Parameters.} Joint fit to training NLL and sampling quality metrics. Values are posterior medians $\pm$ half the width of the 95\% credible interval.}
  \label{tab:joint-scaling}
  \centering
  \input{generated/joint_scaling_plot.tex}
  \end{minipage}\hfill
  \begin{minipage}[t]{0.53\linewidth}
  \vspace{0pt}
  \caption{\textbf{Scaling Model Fit.} In-sample $R^2$ for NLL and the transformed sampling metrics.}
  \label{tab:joint-scaling-r2}
  \centering
  \input{generated/joint_scaling_r2.tex}
  \par\smallskip
  \includegraphics[scale=\pairedfigurescale]{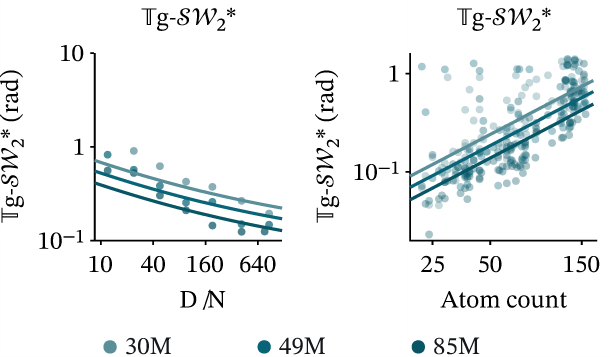}
  \captionof{figure}{\textbf{Torsion Sampling Error Scales with Compute and System Size.} Left: \TgSWast on \textit{Many Peptides} versus loss tokens per non-embedding parameter ($D/N$). Right: Per-system \TgSWast versus atom count for 15M--85M models at $D/N=192$. Both panels use 4096 coordinate bins. Lines show scaling-model estimates at $n=65$ atoms (left) and $D/N=192$ (right).}
  \label{fig:dipper-torus-scaling}
  \end{minipage}
\end{table}

\begin{figure}[H]
  \centering
  \includegraphics[width=\linewidth]{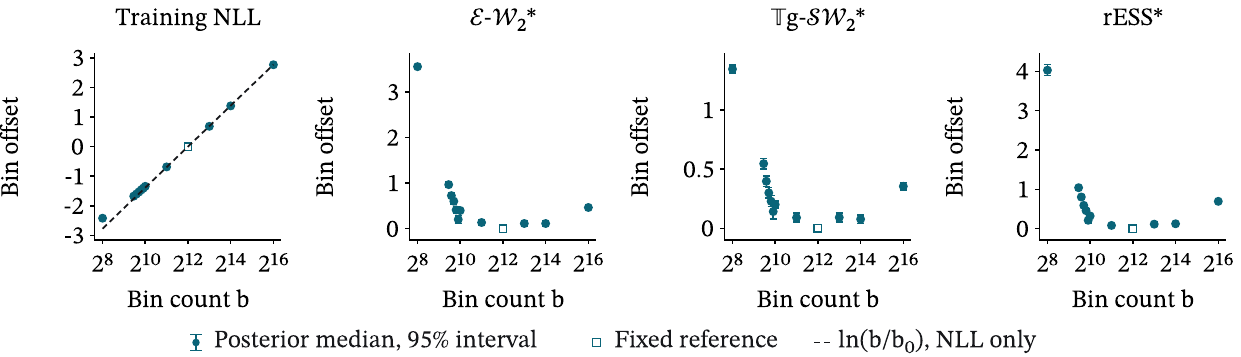}
  \caption{\textbf{Fitted Discretization Resolution Offsets.} Shown are estimated discretization effects on training NLL, \EWast, \TgSWast, and \rESSast. Points show posterior medians with 95\% credible intervals. Open squares mark offsets fixed at zero for $b_0=4096$. The dashed curve in the NLL panel represents $\ln(b/b_0)$.}
  \label{fig:scaling-bin-offsets}
\end{figure}

Among the discretization resolutions explored, we observed a steep penalty in sampling quality at coarse resolutions of $b\le 2^{10}=1024$ bins per axis ($\Delta\ge 3.2\,\mathrm{pm}$) (Figure \ref{fig:scaling-bin-offsets}). In contrast to \citet{rehmann2025autoregressive}, we also observed detrimental effects of high discretization resolutions, an effect we ascribe to a logit gradient bottleneck as we increase the vocabulary size and keep the latent token dimension fixed~\cite{godey2026lostbackpropagationlmhead}.

\begin{figure}[H]
  \centering
  \includegraphics[width=\linewidth]{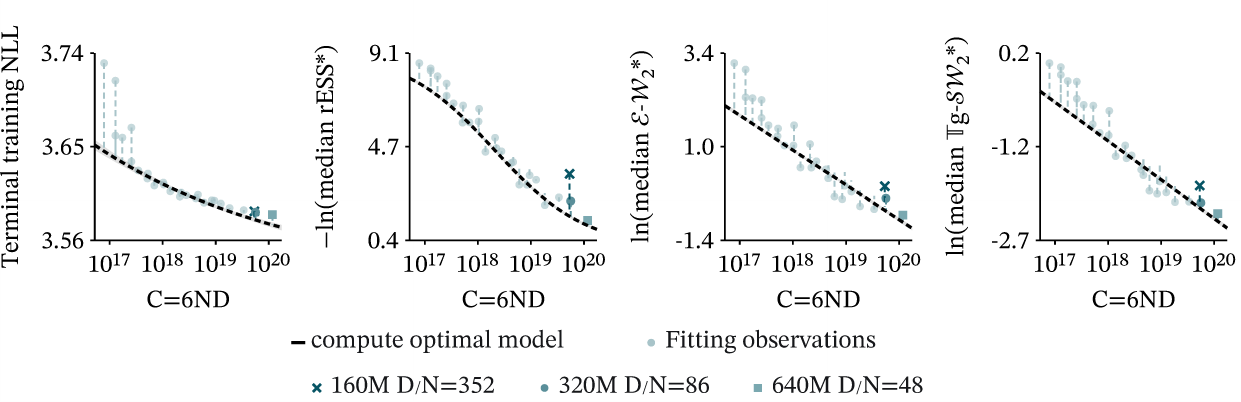}
  \caption{\textbf{Compute Scaling and Held-Out Checkpoint Predictions.} From left to right: terminal training NLL, $-\ln(\rESSast)$, $\ln(\operatorname{median}\EWast)$, and $\ln(\TgSWast)$ versus the training-compute proxy $C=6ND$. All evaluation summaries use the same \textit{Many Peptides} systems\,\P. Filled markers show observed checkpoints and their connecting lines end at the scaling model's predicted sampling quality. The black curves show compute-optimal sampling quality estimates with 95\% generalized-posterior intervals. Three of the four held-out checkpoints (160M, 320M and 640M) are shown, the fourth (85M, $D/N=1536$) is omitted.}
  \label{fig:dipper-compute-metrics}
\end{figure}

When extrapolating to larger models of 160M, 320M, and 640M, we observed reasonable adherence to the empirical scaling model (Figure \ref{fig:dipper-compute-metrics}).

\newpage
\paragraph{\rebuttal{Sensitivity of the Scaling Model.}}
\label{par:scaling-sensitivity}
\rebuttal{%
Within a sensitivity analysis we refitted the empirical scaling model under alternative specifications (Table~\ref{tab:scaling-ablation}), and refitted each of these specifications five times without one model size's data (Table~\ref{tab:scaling-ablation-loso}).
The growth of sampling error with atom count was robust (median $p_E$ \DipperScalingAblationEnergyAtomMedian, 5-fold range across specifications \DipperScalingAblationEnergyAtomMin--\DipperScalingAblationEnergyAtomMax), as was the extrapolation of the training loss to larger models (median held-out NLL RMSE \DipperScalingAblationHeldoutNllRmseMedian{} nats, 5-fold range across specifications  \DipperScalingAblationHeldoutNllRmseMin--\DipperScalingAblationHeldoutNllRmseMax).
The compute-optimal ratio $D^\ast/N$ and the underlying scaling exponents varied more strongly ($D^\ast/N$ at $N=85.5$M: median \DipperScalingAblationTppCoreMedian, 5-fold range \DipperScalingAblationTppCoreMin--\DipperScalingAblationTppCoreMax{} without specification A1, and an extreme value at \DipperScalingAblationTppUnconstrained{} in A1).
Without tied exponents (specification D1), the default model estimate of $D^\ast/N$ was \DipperScalingAblationUntiedTpp{} (5-fold range \DipperScalingAblationUntiedTppMin--\DipperScalingAblationUntiedTppMax).
A naive model without evaluation metric adapters (specification A0) had an estimate of \DipperScalingAblationNaiveTpp{} (5-fold range \DipperScalingAblationNaiveTppMin--\DipperScalingAblationNaiveTppMax).
We consider $D^\ast/N\approx40$ an empirical, evaluation metric sensitive, allocation rule for \textit{Many Peptides}.
The exponent of the training compute required at fixed \EWast{} also varied (median \DipperScalingAblationParityMedian, 5-fold range across specifications \DipperScalingAblationParityMin--\DipperScalingAblationParityMax), because it combines $p_E$ with the less certain $G$-sensitivity and scaling exponents. In every specification with a fitted atom-count exponent, the required training compute grows faster than $n^{\DipperScalingAblationParityMin}$.
Treating \rESSast{} values with an effective sample size of at most \DipperScalingAblationCensorOneEss{} as censored (row C1) changed no estimate beyond its credible interval.
}

\rebuttal{%
Given that ROBIN~\citep{rehmann2025autoregressive}, a previous transferable state-of-the-art model, was trained with an approximate $D/N\approx1100$ (author correspondence), about 30 times our default estimate and far above the range across specifications and held-out model sizes at ROBIN's size of \DipperScalingAblationRobinParamsMillion{}M parameters ($D^\ast/N$ of \DipperScalingAblationTppRobinMin{} to \DipperScalingAblationTppRobinMax, median \DipperScalingAblationTppRobinMedian), previous transferable Boltzmann generators were likely smaller than compute-optimal for their training budget.
}

\begin{table}[htbp]
  \caption{\rebuttal{\textbf{Scaling Model Ablation.} Rows A0--A6 add one modeling choice at a time. A4 is the default model. Rows D1--D3 remove one choice from A4. A3 and D3 are the same fit.
  Rows F1--F3 change the atom-count term of A4 (F1: exponential in $n$; F2: power law plus curvature; F3: extensive, $p_E=p_R=1$ fixed), and row C1 treats \rESSast{} values with an effective sample size of at most \DipperScalingAblationCensorOneEss{} as censored.
  A dash marks a quantity that the row does not define. $D^\ast/N$ is given at $N=85.5$M, and the \CstarEW exponent is the compute exponent at fixed \EWast. Values are posterior medians with 95\% credible intervals. For $D^\ast/N$ above 1000 the interval is omitted.
  The last four columns give the RMSE on the four held-out checkpoints: 85M at $D/N=1536$ and the 160M, 320M and 640M checkpoints of Table~\ref{tab:sota}.
  Best values are \textbf{bold}; second-best values are \underline{underlined}.}}
  \label{tab:scaling-ablation}
  \centering
  \input{generated/dipper_scaling_ablation_table.tex}
\end{table}

\begin{table}[htbp]
  \caption{\rebuttal{\textbf{Scaling Model Ablation with Holdout Model Size.} Each row of Table~\ref{tab:scaling-ablation} was refitted five times, each time without all checkpoints of one model size (8M to 85M). A3 and D3 are the same fit. $D^\ast/N$ is given at $N=85.5$M. Cells give the mean of the five posterior medians and, in brackets, their minimum and maximum. For $D^\ast/N$ above 1000 the interval is omitted. Average RMSE values across 5-fold holdout checkpoints are shown.}}
  \label{tab:scaling-ablation-loso}
  \centering
  \input{generated/dipper_scaling_ablation_loso_table.tex}
\end{table}

\clearpage
\subsection{\rebuttal{Scaling Model on MDQM9-nc}}
\label{sec:scaling-mdqm9}
\paragraph{\rebuttal{Scaling Data and Model.}}
\rebuttal{%
We trained PBG models on MDQM9-nc (Appendix~\ref{sec:datasets}) with the configuration of the \textit{Many Peptides} models of the same size, without repeating training frames, and evaluated \DipperMdqmScalingCheckpointCount{} checkpoints on the MDQM9-nc test molecules.
Given the relatively small amount of data compared to \textit{Many Peptides}, the range of both testable model sizes (\DipperMdqmScalingSizeMin{} to \DipperMdqmScalingSizeMax) and training compute ($D/N=\DipperMdqmScalingTrainedTppMin$ to $\DipperMdqmScalingTrainedTppMax$) was constrained.
As for \textit{Many Peptides} (Appendix~\ref{par:scaling-sensitivity}), we refitted the scaling model under alternative specifications to the checkpoints of \DipperMdqmScalingFitSizeMin{} to \DipperMdqmScalingFitSizeMax{} (Table~\ref{tab:mdqm9-scaling-ablation}), and refitted each specification once without one model size (Table~\ref{tab:mdqm9-scaling-ablation-loso}).
The \DipperMdqmScalingHeldoutCount{} checkpoints of the largest model (\DipperMdqmScalingSizeMax) were held out.
}

\paragraph{\rebuttal{Model Estimates.}}
\rebuttal{%
The extrapolation of the training loss to the held-out checkpoints was robust (median NLL RMSE \DipperMdqmScalingAblationHeldoutNllRmseMedian{} nats, range across specifications and holdouts \DipperMdqmScalingAblationHeldoutNllRmseMin--\DipperMdqmScalingAblationHeldoutNllRmseMax).
\EWast grew with atom count only when test systems entered as hierarchical effects (specification A5, $p_E=\DipperMdqmScalingAblationSystemEnergyAtom$, range across holdouts \DipperMdqmScalingAblationSystemEnergyAtomMin--\DipperMdqmScalingAblationSystemEnergyAtomMax).
The compute-optimal ratio varied strongly ($D^\ast/N$ at the reference size of Table~\ref{tab:mdqm9-scaling-ablation}: median \DipperMdqmScalingAblationTppMedian, range \DipperMdqmScalingAblationTppMin--\DipperMdqmScalingAblationTppMax{} across specifications and holdouts).
Without evaluation metric adapters (A0), the estimate was \DipperMdqmScalingAblationNaiveTpp{} (range across holdouts \DipperMdqmScalingAblationNaiveTppMin--\DipperMdqmScalingAblationNaiveTppMax).
With adapters and untied exponents (A1), it was \DipperMdqmScalingAblationUntiedAdapterTpp{} (\DipperMdqmScalingAblationUntiedAdapterTppMin--\DipperMdqmScalingAblationUntiedAdapterTppMax), and for the default model without tied exponents (D1) it was \DipperMdqmScalingAblationUntiedTpp{} (\DipperMdqmScalingAblationUntiedTppMin--\DipperMdqmScalingAblationUntiedTppMax).
Together, these results support the notion above (Appendix~\ref{par:scaling-sensitivity}) that previous autoregressive transferable Boltzmann generators \citep{rehmann2025autoregressive} were smaller relative to the compute-optimal parameter count given their training compute budget: without the unconstrained specification A1, all MDQM9-nc estimates of $D^\ast/N$ lie between \DipperMdqmScalingAblationTppCoreMin{} and \DipperMdqmScalingAblationTppCoreMax{} (median \DipperMdqmScalingAblationTppCoreMedian), below ROBIN's $D/N\approx1100$.
}

\begin{table}[htbp]
  \caption{\rebuttal{\textbf{MDQM9-nc Scaling Model Ablation.} Layout as in Table~\ref{tab:scaling-ablation}, refitted to the MDQM9-nc checkpoints of \DipperMdqmScalingFitSizeMin{} to \DipperMdqmScalingFitSizeMax.
  $D^\ast/N$ is given at $N=85.5$M.
  The held-out columns give the RMSE on the \DipperMdqmScalingHeldoutCount{} checkpoints of the largest model. Values are posterior medians with 95\% credible intervals.
  Best values are \textbf{bold}; second-best values are \underline{underlined}.}}
  \label{tab:mdqm9-scaling-ablation}
  \centering
  \input{generated/dipper_mdqm9_scaling_ablation_table.tex}
\end{table}

\begin{table}[htbp]
  \caption{\rebuttal{\textbf{MDQM9-nc Scaling Model Ablation with Holdout Model Size.} Each row of Table~\ref{tab:mdqm9-scaling-ablation} was refitted once without each model size, as in Table~\ref{tab:scaling-ablation-loso}.}}
  \label{tab:mdqm9-scaling-ablation-loso}
  \centering
  \IfFileExists{generated/dipper_mdqm9_scaling_ablation_loso_table.tex}
    {\input{generated/dipper_mdqm9_scaling_ablation_loso_table.tex}}
    {\rebuttal{[Exported after the 36-endpoint refit.]}}
\end{table}

\clearpage
\subsection{\rebuttal{Effects of System Count and Simulation Depth}}
\label{sec:data-fraction}
\rebuttal{%
Transferable Boltzmann generators train across multiple systems (i.e.\ distinct peptide sequences) each containing conformational states saved as frames from a molecular dynamics simulation.
As such, two design factors within a dataset emerge: the system diversity and simulation depth.
To estimate the effects of reduced diversity and simulation depth on PBG training, we trained PBG models at $D/N=48$ on $1/d$ of the \textit{Many Peptides} training frames.
To ablate simulation depth we kept the first $200/d$\,ns of every trajectory while retaining all 21{,}700 training systems.
To ablate diversity, we kept only $1/d$ of the training systems, stratified by peptide length, while retaining their full 200\,ns trajectories.
Both arms had the same number of frames and the same compute at each $d$.
We trained PBG 30M, 49M and 85M in both arms and PBG 160M in the diversity arm only, each with selected values of $d\in\{2,4,8,16,32,64\}$.
Models were trained with the same recipe, training seed and NVIDIA GH200 hardware.
Of note, as with all other training runs in this study, we never trained for more than one epoch.
}

\begin{figure}[htb]
  \centering
  \begin{minipage}{0.75\linewidth}
  \includegraphics[width=\linewidth]{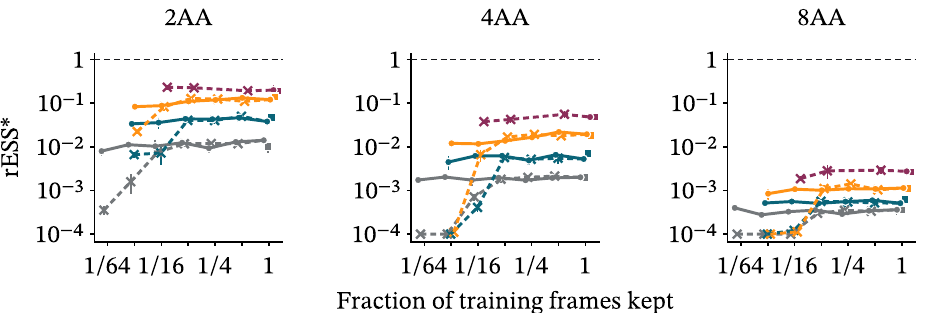}
  \includegraphics[width=\linewidth]{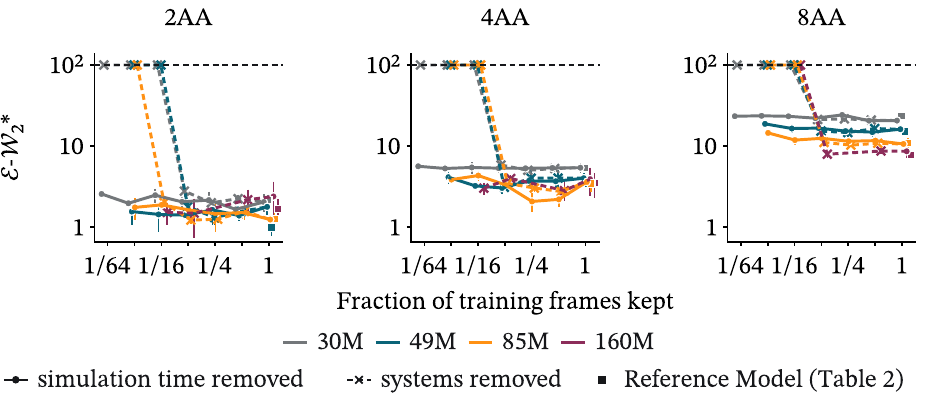}
  \end{minipage}
  \caption{\rebuttal{\textbf{Sampling Quality against Simulation Depth and System Diversity.} \rESSast (top) and \EWast (bottom) of PBG 30M, 49M, 85M and 160M at $D/N=48$ against the fraction of training frames kept relative to the full \textit{Many Peptides} data.
  Solid lines shorten every simulation trajectory, dashed lines remove whole training systems. PBG 160M has only the system arm.
  Points are means over the test systems without YFPHAGYT (Appendix~\ref{sec:yfphagyt}), as the mean of the three sampling-seed means with their SD as bars. The 85M point at $1/32$ of the systems on the 8AA systems uses the production sample set, because one of its seed sample sets has no conformer with the intended stereochemistry for MRDPVLFA. Squares are the seed-42 full-data replicates. Dashed lines mark the maximum \rESSast of 1 and a ceiling of $10^{2}$ for \EWast. Larger \EWast values are drawn at the ceiling.}}
  \label{fig:data-fraction}
\end{figure}

\rebuttal{%
Reduced simulation depth changed sampling quality little, down to 6.25\,ns (Figure~\ref{fig:data-fraction}).
In contrast, at every model size, reduced system diversity caused an observable effect below $1/8$ of the original systems (2{,}713).
At $1/16$ (1{,}358 systems), this effect became weaker as the model grew: \EWast rose above $10^{2}$ in at least one sampling seed on all 29 test octapeptides for PBG 30M, on 25 for PBG 49M, on 21 for PBG 85M, and on only 2 for PBG 160M (EKYYWMQT and NKEKFFQH).
With $1/32$ (681 systems) or fewer, \rESSast on unseen tetra- and octapeptides fell to its minimum of $10^{-4}$ and \EWast rose by multiple orders of magnitude.
At this budget, sampling quality is limited by the number of distinct training systems, not by the simulated time per system.
}

\clearpage
\subsection{\rebuttal{Training Run Variation}}
\label{sec:seed-variation}
\begin{table}[htb]
  \caption{\rebuttal{\textbf{Training Run Variation at $D/N=48$.} Runs of one size have the same optimizer steps and global batch size, but can differ in hardware and GPU count.
  Values with $\pm$ are means for three sampling seeds, with the largest deviation in units of the last decimal place.
  Best values in \textbf{bold}, with second-best \underline{underlined}.}}
  \label{tab:training-seed}
  \centering
  \input{generated/sota_training_seed_plot.tex}
\end{table}

\rebuttal{%
To estimate variation across training runs we trained PBG models to $D/N=48$ with multiple training seeds (Table~\ref{tab:training-seed}).
Across training runs observed \rESS varies about \SeedVarRatioRess{}$\times$ as much as the sampling seed, \EW \SeedVarRatioEnergy{}$\times$, \TW \SeedVarRatioTorus{}$\times$, and \TICAW \SeedVarRatioTica{}$\times$ as much, respectively.
}

\clearpage
\subsection{Stereochemistry of Generated Samples}
\label{sec:stereo}

Maximum-likelihood training rewards the reference handedness of a conformer only weakly: a model that generates both mirror images equally often, $q_\theta(\mathbf{x})=\tfrac{1}{2}p(\mathbf{x})+\tfrac{1}{2}p(R\mathbf{x})$, has an expected negative log-likelihood on the training data only $\log 2$ nats per conformer above that of $p$, because $p(R\mathbf{x})\approx 0$ for training conformers.
This penalty falls on the one decoding step that fixes the handedness and is small against the negative log-likelihood of a whole conformer, so a model can reach a low loss without learning which hand to choose.
We compare the stereocenters of each sample with the reference sequence (Appendix~\ref{sec:yfphagyt}): a sample is \emph{valid as generated}, \emph{fully mirrored} (every stereocenter inverted) or \emph{otherwise invalid} (some, but not all, inverted).
Fully mirrored samples are reflected, which changes coordinates only: we retain the log-likelihood of the generated conformer (Appendix~\ref{sec:experimental-methods}) and its box-volume terms (Appendix~\ref{sec:box-volume}).
Otherwise invalid samples are excluded, as was \StereoNonfiniteFrames{} of the \StereoPbgSamplesMillion{} million PBG samples of Table~\ref{tab:sota} with a non-finite energy or log-likelihood.

\paragraph{Half of the PBG Samples Are Enantiomers.}

For every PBG model and peptide length, \StereoReflectedOfKeptMinPercent\% of the kept samples are reflected \emph{fully mirrored} images (Table~\ref{tab:stereo-validity}).
The excluded share of \emph{otherwise invalid} is \StereoExcludedMeanPercent\% of samples on average, grows with peptide length and with the number of stereocenters (Table~\ref{tab:stereo-centers}), and is largest for PBG 160M and PBG-Joint 160M at 8AA.
In contrast, ROBIN generates almost no mirror images.

\paragraph{Stereochemistry Changes with Random Translations and Training Compute.}
Along the ablation ladder from ROBIN to PBG training (Appendix~\ref{sec:robin-pbg}, Table~\ref{tab:stereo-ladder}), all checkpoints except \StereoLadderCentredMixedCount{} reflect either at least 40\% of their kept samples or at most \StereoLadderNoneMaxPercent\%.
We observe a transition towards half-mirrored sampling with the introduction of random translation (L7 and L8).
In the scaling cohort, the excluded share decreases with training compute (Pearson correlation \StereoScalingExcludedComputeCorrelation{} on logarithmic scales).

\begin{table}[H]
  \edef\pairedtablescale{\fpeval{min(1,\textwidth/\catalogmaxwd)}}%
  \renewcommand{\resizebox}[3]{%
    \makebox[\linewidth][c]{%
      \scalebox{\pairedtablescale}{\normalfont\normalsize #3}}%
  }
  \begin{minipage}[t]{0.595\linewidth}
    \vspace{0pt}
    \centering
    \caption{\textbf{Stereochemistry on \textit{Many Peptides}.} Samples valid as generated (Gen.), reflected (Refl.) or excluded (Excl.). Systems as in Table~\ref{tab:sota}. Worst: largest excluded share of one system.}
    \label{tab:stereo-validity}
    \input{generated/sota_stereo_validity.tex}
  \end{minipage}\hfill
  \begin{minipage}[t]{0.375\linewidth}
    \vspace{0pt}
    \centering
    \caption{\textbf{Stereochemistry in the Scaling Cohort.} Ranges over the checkpoints of each model size.}
    \label{tab:stereo-scaling}
    \input{generated/dipper_scaling_stereo.tex}
  \end{minipage}
  \par\medskip
  \begin{minipage}[t]{0.48\linewidth}
    \vspace{0pt}
    \centering
    \caption{\textbf{Stereochemistry by Number of Stereocenters.} PBG samples of Table~\ref{tab:stereo-validity}, pooled over models and seeds. Worst: largest excluded share of one system.}
    \label{tab:stereo-centers}
    \input{generated/sota_stereo_by_stereocenters.tex}
  \end{minipage}\hfill
  \begin{minipage}[t]{0.48\linewidth}
    \vspace{0pt}
    \centering
    \caption{\textbf{Stereochemistry along the ROBIN-to-PBG Ladders.} Rungs of Table~\ref{tab:robin-pbg-ladder}; means over systems and seeds.}
    \label{tab:stereo-ladder}
    \input{generated/robin_pbg_ladder_stereo.tex}
  \end{minipage}
\end{table}

\clearpage
\subsection{\rebuttal{D-Proline Artifact in \textit{Many Peptides}}}
\label{sec:yfphagyt}

\rebuttal{%
In the published \textit{Many Peptides} data, some peptides contain D-prolines instead of L-prolines in every frame of their trajectory.
These peptides contain a phenylalanine directly before the proline (precision \DproRulePrecision\%, recall \DproRuleRecall\%; Table~\ref{tab:dpro-prev}).
}

\begin{figure}[htb]
  \centering
  \includegraphics[width=\linewidth]{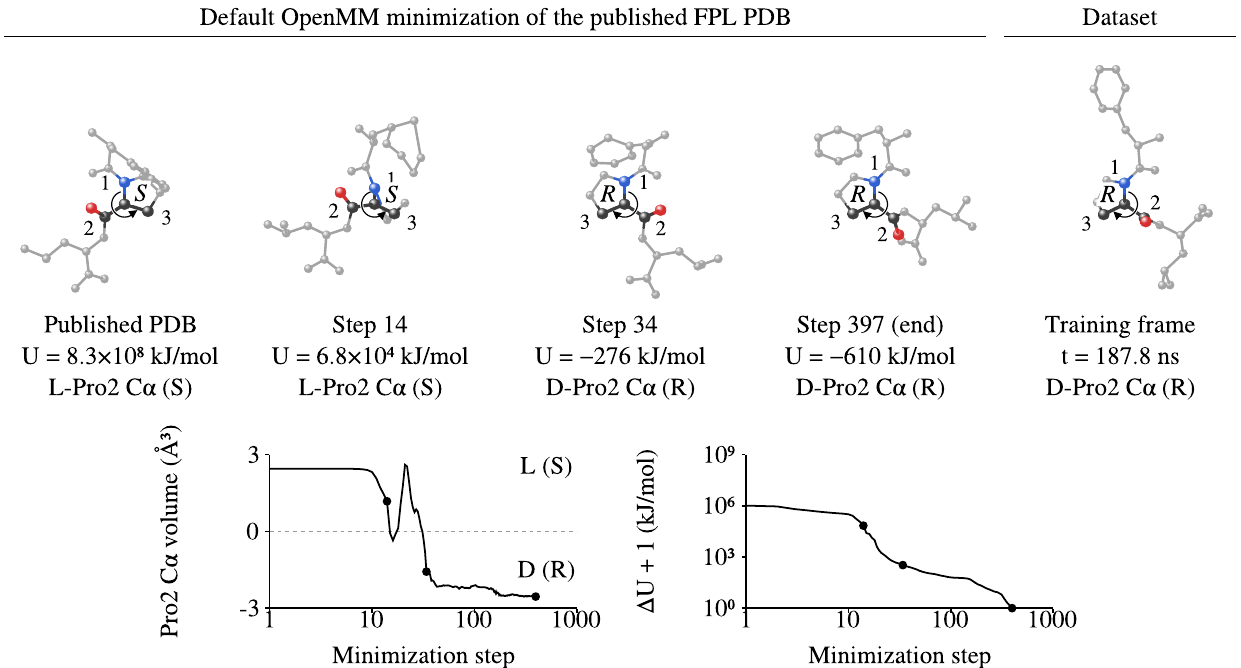}
  \caption{\rebuttal{\textbf{Energy minimization inverts the proline of a published \textit{Many Peptides} structure.} Top: the published starting structure of the training peptide \DproExampleSequence{}, three steps of the default OpenMM minimization, and a training frame of the dataset. Each panel shows the Pro2 C$\alpha$ with its H pointing away from the viewer and its CIP priorities 1--3 (N, C$'$, C$\beta$).
  The C$\alpha$, its N, C$'$ and C$\beta$, and the oxygen on C$'$ have element colors (C dark grey, N blue, O red).
  In the published structure, a Phe1 side-chain atom lies within 1\,\AA{} of the Pro2 ring, and the potential energy is $\DproExampleStartEnergy$\,kJ/mol.
  Bottom: signed volume of the Pro2 C$\alpha$ (positive for L) and potential energy above its minimum.}}
  \label{fig:dpro-minimization}
\end{figure}

\rebuttal{%
While the published starting structure PDB files in \textit{Many Peptides} are all in L form, these are built residue by residue from fixed templates, and for every proline after Phe, Tyr, Trp, His or Asn, a side-chain atom of the preceding residue lies within 1\,\AA{} of the proline ring.
The default OpenMM energy minimization before each simulation removes this overlap, and, for most Phe-Pro pairs, pushes the proline C$\alpha$ through the plane of its three heavy substituents, causing a stereochemistry flip (Figure~\ref{fig:dpro-minimization}).
Of the \DproTrainPeptides{} training peptides, \DproTrainProPeptides{} contain a proline, and \DproTrainAffected{} of these (\DproTrainAffectedPercent\%) have the artifact (Table~\ref{tab:dpro-prev}). The validation peptide IVTVAYFP and the test octapeptide YFPHAGYT are also affected.
When reproducing the default minimization on the published starting structures, all \DproMinimizedAgree{} prolines we tested converged into the stereochemical form found in the dataset.
}

\begin{table}[htb]
  \caption{\rebuttal{\textbf{D-proline artifact by the residue before the proline.} Shown are statistics for the Training Set. Clash: a side-chain heavy atom of the preceding residue lies within 1\,\AA{} of the proline ring in the published starting structure. Reproduced: default OpenMM energy minimization of the published starting structures, as in the dataset pipeline.}}
  \label{tab:dpro-prev}
  \centering
  \input{generated/many_peptides_dpro_prev_residue.tex}
\end{table}

\rebuttal{%
Given these stereochemistry issues, YFPHAGYT was excluded from the reported test systems considered.
Accordingly, Table~\ref{tab:sota} reports octapeptide means over 29 systems.
Importantly, we also report literature derived values in Table~\ref{tab:sota} based on these 29 systems, after correspondence with the authors~\citep{tan2026amortizedsamplingtransferablenormalizing,rehmann2025autoregressive}.
}

\begin{table}[htb]
  \caption{\rebuttal{\textbf{Evaluation results on \textit{Many Peptides} with structure based stereochemistry filter.} Same models, metrics and $10^4$ samples per system as Table~\ref{tab:sota}, but the 8AA columns include the YFPHAGYT artifact. Literature 8AA rows are the published 30-system means~\citep{tan2026amortizedsamplingtransferablenormalizing,rehmann2025autoregressive}. Best values in \textbf{bold}, with second-best \underline{underlined}.}}
  \label{tab:sota-30}
  \centering
  \input{generated/sota_30_plot.tex}
\end{table}

\rebuttal{%
Further inspection showed that in the three sampling runs of Table~\ref{tab:sota}, \DproYfphagytPbgMin--\DproYfphagytPbgMax\% of the PBG samples and \DproYfphagytRobinMin--\DproYfphagytRobinMax\% of the ROBIN samples have D-Pro3, after reflection of mirror images, and at most \DproYfphagytLMax\% have L-Pro3, showing that this was an existing issue across previous transferable Boltzmann generators trained on \textit{Many Peptides}.
This raises the question: Why was such an artifact not identified earlier?
We propose that the implementation choices in stereochemistry filters made by us and previous authors working on \textit{Many Peptides} are responsible.
Our evaluation compares sampled structures against a \textbf{reference sequence} and filters conformers that do not match it.
In contrast, the public evaluation code of \citet{tan2026amortizedsamplingtransferablenormalizing} instead compares the stereocenters of each sample with one \textbf{reference structure}, sourced from an MD frame.
It treats the reference as correct and therefore cannot detect an artifact that the reference and the training data share.
For comparison, we performed an evaluation with a reference structure based filter and report the results in Table~\ref{tab:sota-30}.
It repeats Table~\ref{tab:sota} on all 30 octapeptides, including the YFPHAGYT artifact.
For the literature rows, it uses the published 30-system means.
}

\clearpage
\subsection{\rebuttal{ARBG and PBG}}
\label{sec:robin-pbg}

\rebuttal{ROBIN~\citep{rehmann2025autoregressive} (ARBG) and PBGs are transferable autoregressive Boltzmann generators that emit one coordinate token per atom and axis.
Apart from the conditioning and network architecture, they differ in seven additional aspects of the training protocol, listed in Table~\ref{tab:robin-pbg-differences}.}

\begin{table}[h]
  \centering
  \color{rebuttalgreen}
  \caption{Differences between the ROBIN and PBG training protocols.}
  \label{tab:robin-pbg-differences}
  \small
  \newcommand{\tabitem}{\par\noindent\rightskip=0pt plus 1fil\hangindent=0.8em\hangafter=1\relax}
  \begin{tabular}{@{}lp{0.37\linewidth}p{0.37\linewidth}@{}}
    \toprule
    & ROBIN & PBG \\
    \midrule
    Optimizer~(L1) &
      \tabitem Muon~\citep{jordan2024muon} with fan-in update scaling on every weight matrix, including the embedding and output head
      \tabitem Learning rate $5\times10^{-4}$; AdamW~\citep{loshchilov2019decoupled} at $5\times10^{-5}$
      \tabitem No weight decay
      \tabitem Global gradient-norm clipping at 1.0
      \tabitem Adam $\beta_2=0.999$ &
      \tabitem Scalable Muon~\citep{liu2025muonscalablellmtraining} with update RMS 0.2 on attention and MLP matrices
      \tabitem Learning rate $10^{-3}$ at width 512
      \tabitem Tied embedding on Adam
      \tabitem Weight decay 0.1, scheduled with the squared learning rate
      \tabitem AdaGC clipping~\citep{wang2025adagc}
      \tabitem Adam $\beta_2=0.95$ \\
    \cmidrule{1-3}
    Schedule~(L2) &
      \tabitem One-cycle cosine~\citep{smith2019superconvergence}
      \tabitem 1\% warmup &
      \tabitem Warmup--stable--decay~\citep{wsdscalinglaw}
      \tabitem Warmup over $5N$ tokens
      \tabitem $1-\sqrt{t}$ cooldown over the last sixth \\
    \cmidrule{1-3}
    Precision~(L3) &
      \tabitem float32 &
      \tabitem bfloat16~\citep{kalamkar2019study} forward and backward
      \tabitem float32 master weights \\
    \cmidrule{1-3}
    Batches~(L4) &
      \tabitem One molecule per row, padded to 178 atoms
      \tabitem 448 molecules per step &
      \tabitem Molecules packed into rows of 1028 tokens~\citep{krell2021efficient}
      \tabitem Attention only within each molecule \\
    \cmidrule{1-3}
    Loss~(L5) &
      \tabitem Negative log-likelihood of the coordinate tokens
      \tabitem Mean per molecule, then over molecules &
      \tabitem Negative log-likelihood of the coordinate tokens and the end-of-coordinates token
      \tabitem Mean over all these tokens in the batch \\
    \cmidrule{1-3}
    Quantization~(L6) &
      \tabitem 8192 bins over $\pm1.575\,\mathrm{nm}$
      \tabitem Bin width $0.38\,\mathrm{pm}$ &
      \tabitem 4096 bins over $\pm1.638\,\mathrm{nm}$
      \tabitem Bin width $0.8\,\mathrm{pm}$ \\
    \cmidrule{1-3}
    Translation~(L7) &
      \tabitem Center of mass at the origin
      \tabitem No translation &
      \tabitem Random translation in the box \\
    \cmidrule{1-3}
    Conditioning~(L8) &
      \tabitem Per-atom features: atom type, residue type, residue position, peptide length and atom index
      \tabitem An MLP embeds the features; the result is added to the input and read by one global attention layer &
      \tabitem HELM prompt~\citep{helmnotation} and environment text in the context \\
    \midrule
    Shared & \multicolumn{2}{p{0.74\linewidth}}{%
      \tabitem One token per atom and axis, in atom order; nearest-bin quantization
      \tabitem Random rotation
      \tabitem Muon with five Newton--Schulz steps and Nesterov momentum 0.95
      \tabitem No EMA or label smoothing} \\
    \bottomrule
  \end{tabular}
\end{table}

\rebuttal{To estimate the effect of each protocol difference, we train a ladder of nine models at each of 30M, 85M and 160M parameters, each on $48N$ loss tokens.
Rung L0 applies the ROBIN protocol to the PBG network, with the atom features as an adapter.
Each further rung switches the next row of Table~\ref{tab:robin-pbg-differences} to the PBG setting. L8 is a PBG.
All rungs share the PBG network, so the ladder does not test the architecture.}

\begin{table}[h]
  \centering
  \color{rebuttalgreen}
  \caption{\textbf{Sampling quality along the ladders from ROBIN to PBG training protocol.} All rungs are trained on $48N$ loss tokens. The 30M L0 model samples no conformer with the intended stereochemistry for several octapeptides, so its 8AA cells are empty. The largest deviation of a seed mean is shown in $\pm$ in units of the last decimal place.}
  \label{tab:robin-pbg-ladder}
  \input{generated/robin_pbg_ladder.tex}
\end{table}

\begin{figure}[h]
  \centering
  \color{rebuttalgreen}
  \begin{minipage}{0.75\linewidth}
  \includegraphics[width=\linewidth]{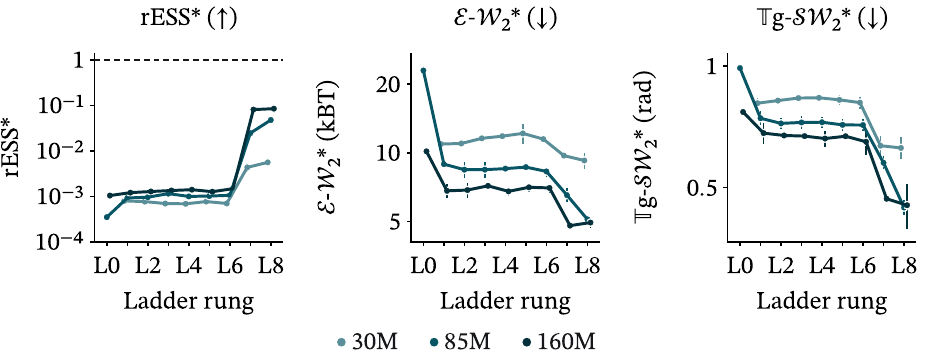}
  \end{minipage}
  \caption{\textbf{Sampling Quality along the Ladders from ROBIN to PBG Training Protocol.} \rESSast, \EWast and \TgSWast of the 30M, 85M and 160M rungs of Table~\ref{tab:robin-pbg-ladder}. Points are means over the 89 test systems of Table~\ref{tab:robin-pbg-ladder} and the three sampling seeds. Bars are the SD of the three seed means.}
  \label{fig:robin-pbg-ladder-panels}
\end{figure}

\rebuttal{At every model size, three of the eight ladder steps account for nearly all of the protocol-driven improvement (Table~\ref{tab:robin-pbg-ladder}, Figure~\ref{fig:robin-pbg-ladder-panels}): the Muon optimizer settings (L1), random translation (L7) and the HELM prompt (L8).
Together they give at least 89\% of the L0-to-L8 change of each metric in log scale.
The share is lower only for \EWast{} and \TgSWast{} on the 160M 2AA systems (75\% and 76\%), which have the smallest total change.
The other changes (L2 to L6) have small effects.}

\newpage
\subsection{\rebuttal{Cross-Class Transfer at Matched Compute}}
\label{sec:joint-matched-control}
\rebuttal{To estimate the extent to which training a PBG model across multiple datasets (PBG-Joint) led to differences in sampling quality on the largest dataset, \textit{Many Peptides}, we trained 160M PBG models with the same architecture, optimizer, seed, batch rows and hardware, and near equal loss tokens.
Training across multiple datasets neither significantly improved nor reduced sampling quality on \textit{Many Peptides}.}

\begin{table}[htbp]
  \centering
  \color{rebuttalgreen}
  \caption{\textbf{Sampling quality for PBG-Joint against a comparable PBG model trained on \textit{Many Peptides} only.}}
  \label{tab:joint-matched-control}
  \input{generated/joint_matched_peptide_control.tex}
\end{table}

\clearpage
\subsection{Quantization and Reconstruction Effects}
\label{sec:extensive-msf}

Sampling error in Prefix-Conditioned Boltzmann Generators can be ascribed to either \textit{Quantization Error}, an insufficiency of the discretization resolution, or \textit{Generator Error}, an insufficiency in learning the quantized target distribution $p_{\Delta}$.
In the following we describe a set of experiments and theoretical considerations to understand \textit{Quantization Error} in isolation.

Recall the midpoint and uniform reconstruction kernels, $\mathcal R_\Delta^{\mathrm{mid}}$ and $\mathcal R_\Delta^{\mathrm{unif}}$, respectively (Appendix~\ref{sec:midpoint-cubature}). Given a quantized macrostate $\mathbf k$, the midpoint kernel returns the cell center $\mathbf r_{\mathbf k}\in\mathbb R^{n\times3}$, while the uniform kernel draws a conformation $\mathbf y\sim\mathcal R_\Delta^{\mathrm{unif}}(\cdot\mid\mathbf k)$ uniformly within the cell $C(\mathbf k)$, with $\mathbf y\in\mathbb R^{n\times3}$.

\paragraph{Energetic Discretization Error under Midpoint Reconstruction.}
To isolate Quantization Error, we draw $\mathbf x\sim p$ and set $\mathbf k=\kappa_\Delta(\mathbf x)$, applying the quantizer componentwise. We then directly reconstruct the conformer in a \textit{roundtrip}, starting with the midpoint reconstruction kernel. This gives us a midpoint reconstruction induced energy error and its expectation
\begin{align*}
  \Delta U^{\mathrm{mid}}
  &\equiv U(\mathbf r_{\mathbf k})-U(\mathbf x),
  \\
  \mathbb E[\Delta U^{\mathrm{mid}}]
  &=\sum_{\mathbf k}p_\Delta(\mathbf k)
  \left[U(\mathbf r_{\mathbf k})-\mathbb E_p[U(\mathbf x)\mid\mathbf k]\right].
\end{align*}

Let's denote $\boldsymbol\delta=\mathbf x-\mathbf r_{\mathbf k}\in\mathbb R^{n\times3}$ the displacement of the original conformation from its cell center. Each Cartesian component of $\boldsymbol\delta$ lies in $[-\Delta/2,\Delta/2)$. Considering the limit $\Delta\to0$ at fixed $n$ and assuming a smooth potential and finite required moments, a Taylor expansion around the cell center gives
\begin{equation*}
  \begin{split}
    U(\mathbf x)
    ={}&U(\mathbf r_{\mathbf k})
    +\nabla U(\mathbf r_{\mathbf k})^\top\boldsymbol\delta
    +\frac12\boldsymbol\delta^\top\nabla^2U(\mathbf r_{\mathbf k})\boldsymbol\delta
    +o(\|\boldsymbol\delta\|^2).
  \end{split}
\end{equation*}

The conditional Boltzmann density of the displacement favors lower-energy regions. Expanding this density within the cell gives
\begin{align*}
  \mathbb E_p[\boldsymbol\delta\mid\mathbf k]
  &=-\frac{\beta\Delta^2}{12}\nabla U(\mathbf r_{\mathbf k})+o(\Delta^2),
  \\
  \mathbb E_p[\boldsymbol\delta\boldsymbol\delta^\top\mid\mathbf k]
  &=\frac{\Delta^2}{12}\mathbf I_{3n}+o(\Delta^2),
\end{align*}
where $\mathbf I_{3n}$ is the identity matrix and coordinate arrays are treated as vectors in $\mathbb R^{3n}$. Taking the conditional expectation of $\Delta U^{\mathrm{mid}}=U(\mathbf r_{\mathbf k})-U(\mathbf x)$ in the Taylor expansion gives
\begin{equation*}
  \mathbb E[\Delta U^{\mathrm{mid}}\mid\mathbf k]
  =-\nabla U(\mathbf r_{\mathbf k})^\top
  \mathbb E_p[\boldsymbol\delta\mid\mathbf k]
  -\frac12\operatorname{tr}\!\left[
    \nabla^2U(\mathbf r_{\mathbf k})
    \mathbb E_p[\boldsymbol\delta\boldsymbol\delta^\top\mid\mathbf k]
  \right]
  +o(\Delta^2).
\end{equation*}
Substituting the two displacement moments from above then yields
\begin{equation*}
  \mathbb E[\Delta U^{\mathrm{mid}}\mid\mathbf k]
  =\frac{\beta\Delta^2}{12}\|\nabla U(\mathbf r_{\mathbf k})\|^2
  -\frac{\Delta^2}{24}\operatorname{tr}\nabla^2U(\mathbf r_{\mathbf k})
  +o(\Delta^2).
\end{equation*}

Averaging over $\mathbf k\sim p_\Delta$ and replacing the cell-center averages by their continuum limits in the leading coefficient leaves us with
\begin{equation*}
  \mathbb E[\Delta U^{\mathrm{mid}}]
  =\frac{\beta\Delta^2}{12}\mathbb E_p[\|\nabla U\|^2]
  -\frac{\Delta^2}{24}\mathbb E_p[\operatorname{tr}\nabla^2U]
  +o(\Delta^2).
\end{equation*}
Provided the boundary term vanishes, integration by parts gives
\begin{equation*}
  \begin{aligned}
    0
    &=\int_\Omega\nabla\!\cdot\!\left[p(\mathbf x)\nabla U(\mathbf x)\right]\,\mathrm d\mathbf x\\
    &=\mathbb E_p[\operatorname{tr}\nabla^2U]
    -\beta\mathbb E_p[\|\nabla U\|^2].
  \end{aligned}
\end{equation*}
The mean curvature term thus cancels half of the first contribution, leaving us with a leading-order description of the expected energetic discretization error under midpoint reconstruction as a function of the mean squared force $\mathbb E_p[\|\nabla U\|^2]$ and the squared bin size $\Delta^2$.
\begin{equation}
  \boxed{\mathbb E[\Delta U^{\mathrm{mid}}]
  =\frac{\beta\Delta^2}{24}\mathbb E_p[\|\nabla U\|^2]+o(\Delta^2).}
  \label{eq:midpoint-mean-energy-shift}
\end{equation}

\paragraph{Energetic Discretization Error under Uniform Reconstruction.}
We can now proceed to describe the expected energetic discretization error under uniform reconstruction. The uniform reconstruction can be expressed as an additional random displacement $\boldsymbol\eta$ from the cell center:
\begin{equation*}
  \mathbf y=\mathbf r_{\mathbf k}+\boldsymbol\eta,
  \qquad
  \eta_{id}\overset{\mathrm{iid}}{\sim}\operatorname{Unif}[-\Delta/2,\Delta/2),
\end{equation*}
drawn independently of $\mathbf x$ conditional on $\mathbf k$. Its energy change satisfies
\begin{equation*}
  \Delta U^{\mathrm{unif}}
  \equiv U(\mathbf y)-U(\mathbf x)
  =\Delta U^{\mathrm{mid}}
  +U(\mathbf r_{\mathbf k}+\boldsymbol\eta)-U(\mathbf r_{\mathbf k}).
\end{equation*}
The uniform displacement has zero mean, so its linear contribution vanishes. Integrating the quadratic term over the cell gives
\begin{equation*}
  \begin{split}
    \mathbb E[\Delta U^{\mathrm{unif}}\mid\mathbf k]
    ={}&\mathbb E[\Delta U^{\mathrm{mid}}\mid\mathbf k]
    +\frac{\Delta^2}{24}\operatorname{tr}\nabla^2U(\mathbf r_{\mathbf k})
    +o(\Delta^2).
  \end{split}
\end{equation*}
After averaging over cells and applying the same canonical identity, we obtain the analogous leading-order energetic discretization error under uniform reconstruction. Of note, to leading order, this expected error is double the expected energetic discretization error under midpoint reconstruction $\mathbb E[\Delta U^{\mathrm{unif}}]=2\,\mathbb E[\Delta U^{\mathrm{mid}}]+o(\Delta^2)$
\begin{equation}
  \boxed{\mathbb E[\Delta U^{\mathrm{unif}}]
  =\frac{\beta\Delta^2}{12}\mathbb E_p[\|\nabla U\|^2]+o(\Delta^2).}
  \label{eq:uniform-mean-energy-shift}
\end{equation}

\paragraph{Empirical Extensivity of Mean Squared Force introduces Atom Count Dependence for Energetic Discretization Error.}
\begin{wrapfigure}{R}{0.55\textwidth}
  \centering
  \includegraphics[width=\linewidth]{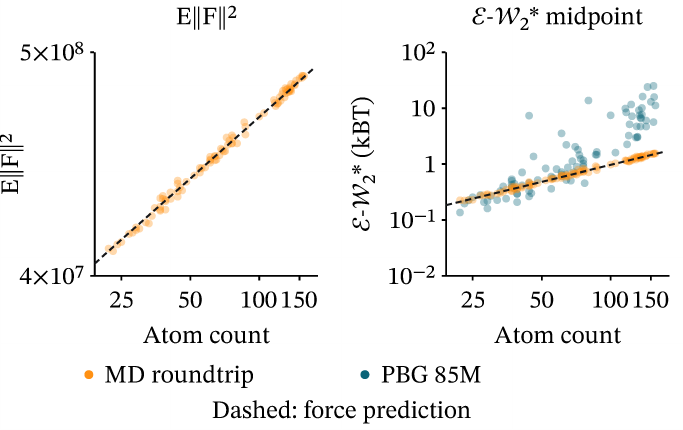}
  \caption{\textbf{Force Scaling and Roundtrip Quantization Error.} Total mean-squared force (left) and midpoint \EWast (right) versus atom count at $b=4096$ ($\Delta=0.8\,\mathrm{pm}$). MD roundtrips are shown in orange, while petrol denotes PBG 85M ($D/N=192$) samples. Dashed curves show force-based predictions (Equation~\ref{eq:midpoint-mean-energy-shift}).}
  \label{fig:force-msf-scatter}
\end{wrapfigure}

Recall that the total mean squared force is the sum over atom-wise mean squared forces. If the mean squared force per atom is approximately constant across systems (Extensivity), the total grows proportionally to $n$. At fixed temperature, the leading mean energy error then scales as
\begin{equation}
  \boxed{\mathbb E[\Delta U^{\mathrm{mid}}]\propto n\Delta^2,}
  \label{eq:reconstruction-energy-scaling}
\end{equation}
within the regime where the small-cell expansion remains valid. At sufficiently fine resolution, halving the bin width $\Delta$ reduces the expected energy error by approximately a factor of four, while doubling the atom count approximately doubles it under the stated extensive force-scaling assumption.

The energetic Wasserstein distance \EWast between the uniformly weighted reconstructed and reference energy distributions is at least their absolute difference in mean energy:
\begin{equation}
  \EWast\ge\beta\left|\mathbb E[\Delta U]\right|.
  \label{eq:reconstruction-energy-w2-bound}
\end{equation}

To test the mean squared force extensivity assumption and dependent predictions, we inspect reference MD conformers from the \textit{Many Peptides} dataset. In fact, the expected mean square force is well approximated as an extensive property within this dataset (Figure \ref{fig:force-msf-scatter}).
We then proceeded to quantize and reconstruct them in a roundtrip, using uniform importance weights to calculate an \EWast.
Observed roundtrip energetic error was matched by Equation \ref{eq:reconstruction-energy-w2-bound} and scaled with atom count as a power of 1 (Figure \ref{fig:force-msf-scatter}).
In contrast, representative samples drawn from a PBG showed superlinear scaling with atom count, in line with the approximately power of 2 scaling in the empirical scaling model. Overall, these results suggest that parts of the sampling quality scaling observed in PBGs can be attributed to \textit{Quantization Error} alone, which scales linearly with atom count.

\WFclear
\newpage
\begin{wrapfigure}{R}{0.55\textwidth}
  \centering
  \includegraphics[width=\linewidth]{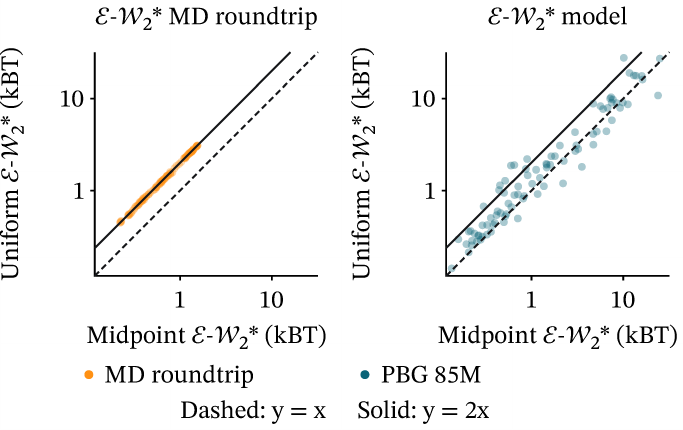}
  \caption{\textbf{Midpoint versus Uniform Reconstruction.} Energy \EWast comparisons for MD roundtrips (left) and representative PBG 85M samples (right) at $b=4096$ ($\Delta=0.8\,\mathrm{pm}$), with $D/N=192$ for the generator. Dashed lines mark equality.}
  \label{fig:reconstruction-comparison}
\end{wrapfigure}

In line with Equations \ref{eq:midpoint-mean-energy-shift} and \ref{eq:uniform-mean-energy-shift}, roundtrip samples produced by roundtrips with uniform reconstruction had twice the \EWast as equivalent samples produced via midpoint reconstruction (Figure \ref{fig:reconstruction-comparison}).
Relatedly, representative PBG samples showed an advantage for midpoint over uniform reconstruction, too.

\begin{wrapfigure}{R}{0.75\textwidth}
  \centering
  \includegraphics[width=\linewidth]{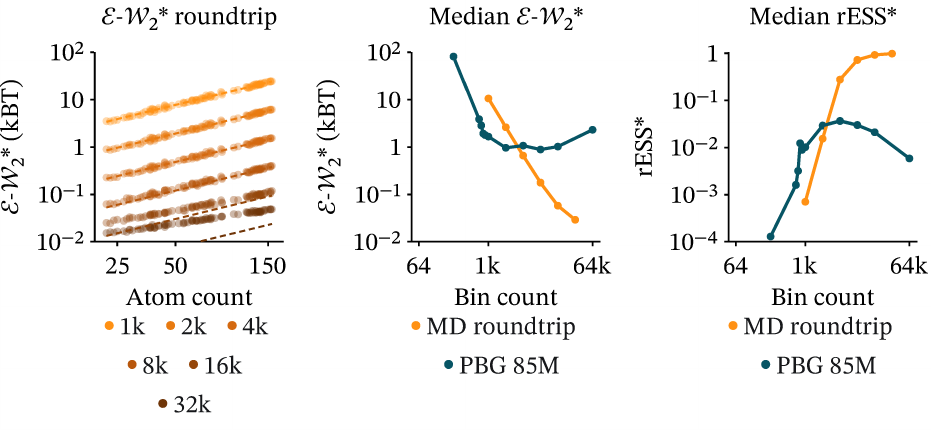}
  \caption{\textbf{Quantization Error across Resolutions.} Left: per-system midpoint roundtrip \EWast versus atom count, with force-based predictions dashed. Middle and right: median \EWast and \rESSast across systems for midpoint roundtrips (orange) and PBG 85M (petrol), as a function of bin count.}
  \label{fig:roundtrip-paper-row}
\end{wrapfigure}

\paragraph{Discretization Resolution.}
To evaluate the accuracy of Equations \ref{eq:midpoint-mean-energy-shift} and \ref{eq:uniform-mean-energy-shift} along the bin size $\Delta$ axis, too, we performed roundtrip experiments at varying resolutions (Figure~\ref{fig:roundtrip-paper-row}).
Finer binning resolution reduces the discretization error. As previously described, while roundtrip experiments show decreasing \EWast, eventual generative model specific aspects limit further increases in quantization resolution without changes in the overall box size $L$ or the model capacity.

\par
\ifnum\value{WF@wrappedlines}>1
  \vspace{\dimexpr\value{WF@wrappedlines}\baselineskip-\baselineskip\relax}
\fi
\WFclear

\begin{figure}[H]
  \centering
  \includegraphics[width=\linewidth]{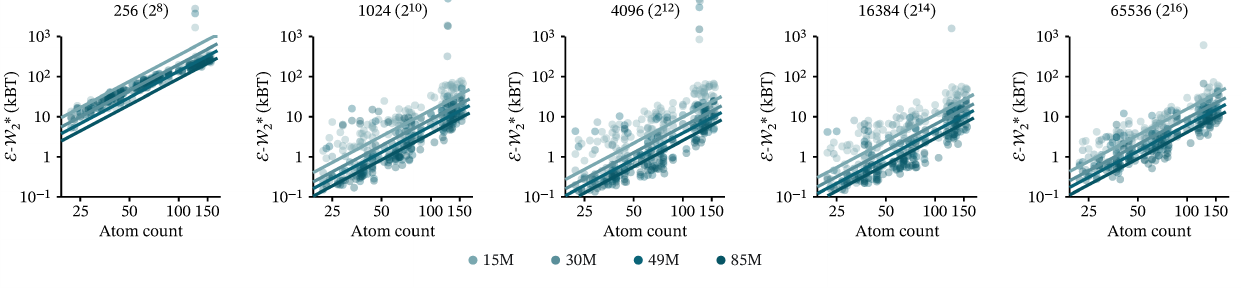}
  \caption{\textbf{Energetic Error versus System Size and Quantization Resolution.} Per-system \EWast for PBGs 15M--85M on \textit{Many Peptides}. Solid curves show the fitted scaling law; the display is capped at $10^3$.}
  \label{fig:dipper-energy-atoms}
\end{figure}

\clearpage
\subsection{Box-Volume Correction of Importance Weights}
\label{sec:box-volume}

Random translation (Table~\ref{tab:robin-pbg-differences}) rotates each centred conformer $s$ by a uniformly random rotation $R$ if its implied axis extents $\Delta_a(s,R)$ fit the coordinate box $[-h,h]^3$ ($2h=3.28$\,nm).
Once rotated, it then translates it uniformly over all positions inside the box, which fill a volume $V(s,R)=\prod_{a=1}^{3}\bigl(2h-\Delta_a(s,R)\bigr)$.
With $F(s)$ the fraction of rotations that fit the box, a model that matches the augmented data has the density $q(x)\propto p(s)/\bigl(F(s)\,V(s,R)\bigr)$.
Because the energy is invariant to rotation and translation, the importance weights for the target $\pi(x)\propto e^{-\beta U(x)}$ are then proportional to $F(s)\,V(s,R)$ even for a perfect model: they favour compact conformers, which have more room in the box, and self-normalization does not remove this.
The target $\pi(x)\propto e^{-\beta U(x)}/\bigl(F(s)\,V(s,R)\bigr)$ has the conformational marginal $p(s)$, so when considering the Box-Volume effects, the corrected log weights are
\begin{equation}
  \log w(x)=-\beta U(x)-\log q(x)-\log V(s,R)-\log F(s).
  \label{eq:box-volume-weights}
\end{equation}
Here $F=1$ when no interatomic distance exceeds $2h$; otherwise we estimate $F$ from 1024 rotations.

\paragraph{Box-Volume Corrected Bias of a Perfect Model.}

To measure this bias, we treated the $10^4$ reference MD frames of every test system as samples of a perfect model.
We rotated them as in training and weighted them by $F(s)\,V(s,R)$ (Table~\ref{tab:box-volume}, first row).
All frames had $F=1$.

The bias has a small effect on \rESSast{}.
The standard deviation of the log weights grew with peptide length to only \BoxVolumeLogWeightSdEightAA{} nats at 8AA.
A perfect model would still reach a \rESSast{} of at least \BoxVolumePerfectRessMin{}.
The bias explains only a small part of the decrease of \rESSast{} with atom count (Figure~\ref{fig:dipper-d-metric}).

The bias also shifts the energy.
Compact conformers have lower energies, so the uncorrected weights shifted the mean reduced energy of the 8AA reference by $\BoxVolumeMeanShiftEightAA\,k_BT$.
This shift sets a floor for the \EWast{} of a perfect model.
Torsions were hardly affected.

\paragraph{Box-Volume Corrected Evaluation.}

We recomputed the metrics of the stored rollouts with the corrected weights for the \BoxVolumeModels{} models trained with random translation (the PBG rows of Table~\ref{tab:sota} and rungs L7 and L8 at 30M, 85M and 160M, Appendix~\ref{sec:robin-pbg}).
The correction raised \rESSast{} slightly for all models and changed each metric by at most \BoxVolumeMaxChangeRessPercent\% (\rESSast), \BoxVolumeMaxChangeEnergyPercent\% (\EWast) and \BoxVolumeMaxChangeTorusPercent\% (\TgSWast).

\begin{table}[H]
  \centering
  \caption{\textbf{Box-Volume Correction of Importance Weights.} First row: reference MD frames rotated as in training and weighted by $F(s)\,V(s,R)$ (\rESSast{} is the perfect-model rESS; \EWast{} and \TgSWast{} compare the reference with its reweighted copy). Other rows: models trained with random translation, with the standard weights and with the corrected weights of Equation~\eqref{eq:box-volume-weights}, as means over test systems and the three sampling seeds; $\pm$ is the largest deviation of a seed mean in units of the last decimal.}
  \label{tab:box-volume}
  \input{generated/box_volume_correction.tex}
\end{table}

\clearpage
\subsection{Sampling Cost against Molecular Dynamics}
\label{sec:cost-vs-md}
\rebuttal{We compare PBG 640M with the MD that generated \textit{Many Peptides} on the 89 test systems of Table~\ref{tab:sota}, timing both on the same H100 GPUs.
For PBG, speculative decoding was performed with the 15M drafter.}

\begin{figure}[H]
  \centering
  \begin{minipage}{0.5\linewidth}
  \includegraphics[width=\linewidth]{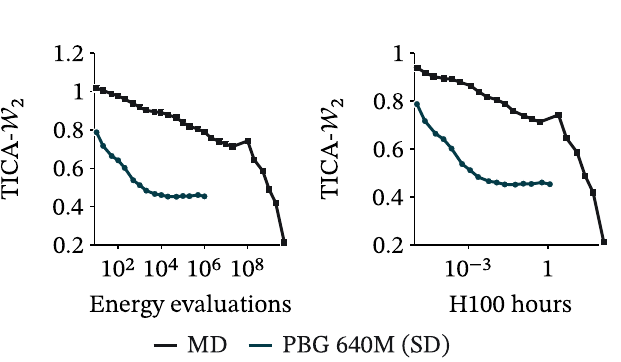}
  \end{minipage}
  \caption{\rebuttal{\textbf{TICA Error against Sampling Cost.} \TICAW for all test systems. Companion to Figure~\ref{fig:cost-vs-md-main}.}}
  \label{fig:cost-vs-md-tica}
\end{figure}

\begin{figure}[H]
  \centering
  \begin{minipage}{0.75\linewidth}
  \includegraphics[width=\linewidth]{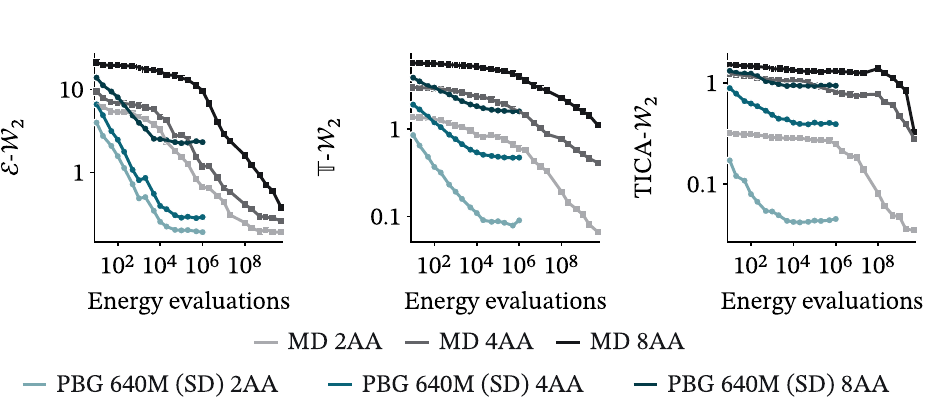}
  \end{minipage}
  \caption{\rebuttal{\textbf{Sampling Error against Energy Evaluations by Peptide Length.} MD (grey, squares) and PBG 640M (petrol, circles). Lighter lines are shorter peptides.}}
  \label{fig:cost-vs-md-length-evaluations}
\end{figure}

\begin{figure}[H]
  \centering
  \begin{minipage}{0.75\linewidth}
  \includegraphics[width=\linewidth]{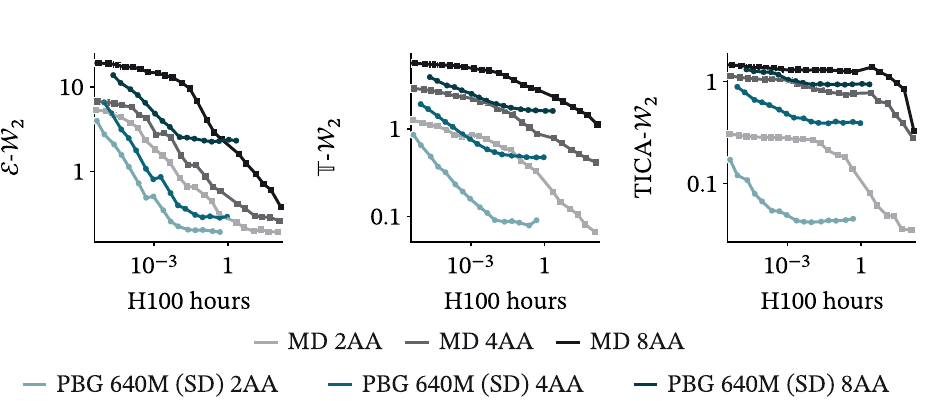}
  \end{minipage}
  \caption{\rebuttal{\textbf{Sampling Error against H100-Hours by Peptide Length.} As Figure~\ref{fig:cost-vs-md-length-evaluations}, against H100-hours per system.}}
  \label{fig:cost-vs-md-length-h100}
\end{figure}

\clearpage
\subsection{Increasing Sampling Rate with Speculative Decoding}

\paragraph{Sampling Rate Data.}
Sampling efficiency can be understood as the product of sampling quality and sampling rate, and measured in effective samples per second.
To model the sampling rate of PBGs and investigate the ability to speed up their sampling via speculative decoding~\cite{leviathan2023fastinferencetransformersspeculative}, we proceeded to observe their sampling rate at a fixed batch size of $512$.
We measured autoregressive (AR) sampling for 30M--640M PBG models trained on \textit{Many Peptides}, and speculative decoding (SD) for 85M--640M targets using a 15M drafter at the available proposal strides $K=1,\ldots,4$.

\begin{table}[H]
  \centering
  \caption{\textbf{Model Architectures and Inference Covariates.} $n_{\mathrm{layers}}$ is the layer count, $d_{\mathrm{model}}$ the hidden width, $d_{\mathrm{ff}}$ the SwiGLU intermediate width, and $H_{\mathrm Q}/H_{\mathrm{KV}}$ the query/KV head counts. All heads have width $d_{\mathrm{head}}=128$. $N_{4096}$ counts executed dense parameters with the 4096-row coordinate output head, excluding normalization scales and embedding lookups. $w_{4096}=2N_{4096}$ is the BF16 weight-read estimate per forward call, shared across the batch. $v_{\mathrm{KV}}$ is the BF16 KV size per cached token and sequence, summed over layers.}
  \label{tab:trained-architectures}
  \input{generated/sampling_model_architectures.tex}
\end{table}

\paragraph{Empirical Sampling Rate Model.}
We estimate batch sampling time using an architecture-based model of the decoding process for both autoregression (AR) and speculative decoding (SD). These estimates depend on model architecture, prompt and output length, and decoding method. For speculative decoding, they also depend on proposal stride and measured acceptance. We fit a shared empirical model:
\begin{equation}
\begin{aligned}
T_B&=a+\alpha W+\beta\left(V_1+V_{\mathrm{write}}+\frac{D_{\mathrm{read}}}{q^\gamma}\right)
       +\tau(J+\kappa U),\\
R&=B/T_B,\qquad S=T_B^{\mathrm{AR}}/T_B^{\mathrm{SD}}.
\end{aligned}
\label{eq:rate-model}
\end{equation}
Here $W$ estimates cumulative weight reads. $V_1$, $V_{\mathrm{write}}$, and $D_{\mathrm{read}}$ estimate cache traffic from single-token calls, target writes, and packed target reads, respectively. $J$ counts layer calls, and $U$ counts additional target layer-token positions during verification. Traffic is expressed in TB and execution counts in thousands. The call width is $q=1$ for AR and $q=K+1$ for SD. For SD, we infer a constant conditional acceptance probability from recorded accepted-to-proposed token ratios for each model.

\begin{figure}[htbp]
  \centering
  \includegraphics[width=\linewidth]{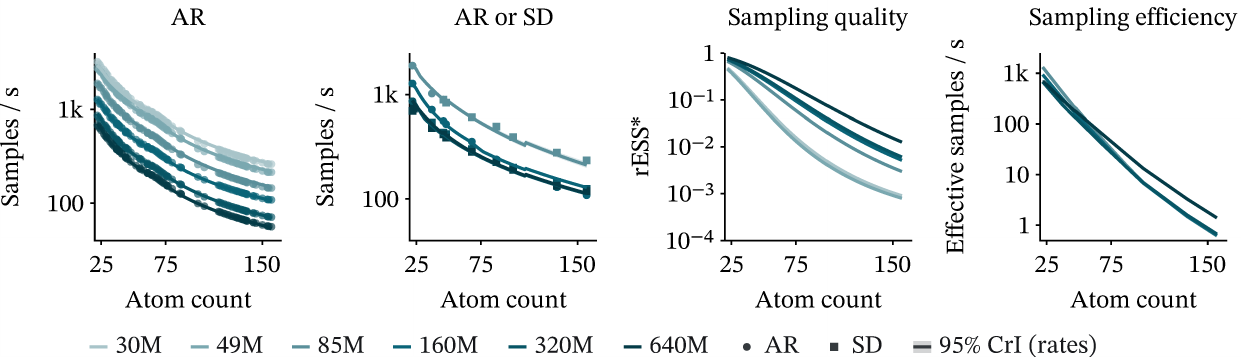}
  \caption{\textbf{Sampling Rate, Quality, and Efficiency.} From left to right: Measured autoregressive (AR) sampling rates and model estimates. Highest measured sampling rate using speculative-decoding (SD) strides with a 15M drafter across all strides. Refitted scaling-model estimates of \rESSast for large models that were previously part of the extrapolation holdout. Sampling efficiency of PBGs at optimal speculative decoding.}
  \label{fig:dipper-core-eff-speed-atoms}
\end{figure}

\paragraph{Model Specification.}
For each model--peptide--stride setting $i$, we convert the median rate $r_i$ across five repeats to batch latency $t_i=B/r_i$. We fit independent Gaussian residuals on log latency, with separate residual scales for AR and SD:
\begin{equation}
\log t_i\sim\mathcal N\!\left(\log T_{B,i},\sigma_{m(i)}^2\right),
\qquad m(i)\in\{\mathrm{AR},\mathrm{SD}\}.
\label{eq:rate-model-likelihood}
\end{equation}

Prior scales use seconds for $a$, seconds per TB for $\alpha$ and $\beta$, and seconds per thousand layer calls for $\tau$. The remaining parameters are dimensionless.

\input{generated/rate_model_specification.tex}

\paragraph{Rate Model Estimates.}
The empirical sampling rate model has reported mean absolute percentage errors below 10\% for AR rates, SD rates, and speedups (Tables~\ref{tab:rate-model-parameters} and~\ref{tab:rate-model-fit}).
Given the optimal sampling rates for a given model and peptide (Figure \ref{fig:dipper-core-eff-speed-atoms}, second from the left), and a model of sampling quality, the sampling efficiency, expressed as the $\rESSast/s$, was estimated (Figure \ref{fig:dipper-core-eff-speed-atoms}, third and fourth from the left).
Given that both sampling quality \textit{and} sampling rate decay with atom count, the sampling efficiency drops from ca.\ 1000 effective samples per second to just 1 effective sample per second as atom count increases from ca.\ 25 to 150 atoms.
This relationship is opening an opportunity for future work to improve sampling efficiency.

\begin{table}[htbp]
  \centering
  \begin{minipage}[t]{0.44\linewidth}
  \vspace{0pt}
  \caption{\textbf{Rate Model Parameters.} Values are posterior medians with 95\% CI.}
  \label{tab:rate-model-parameters}
  \centering
  \input{generated/rate_model_parameters.tex}
  \end{minipage}\hfill
  \begin{minipage}[t]{0.53\linewidth}
  \vspace{0pt}
  \caption{\textbf{Rate Model Fit.} Batch-latency $R_T^2$ and mean absolute percentage errors (MAPE) for rates and speedups.}
  \label{tab:rate-model-fit}
  \centering
  \input{generated/rate_model_fit.tex}
  \end{minipage}
\end{table}

\clearpage
\subsection{Solvation Free Energy Estimation}
\label{sec:asfe-estimation}
\begin{figure}[H]
  \centering
  \includegraphics[width=\linewidth]{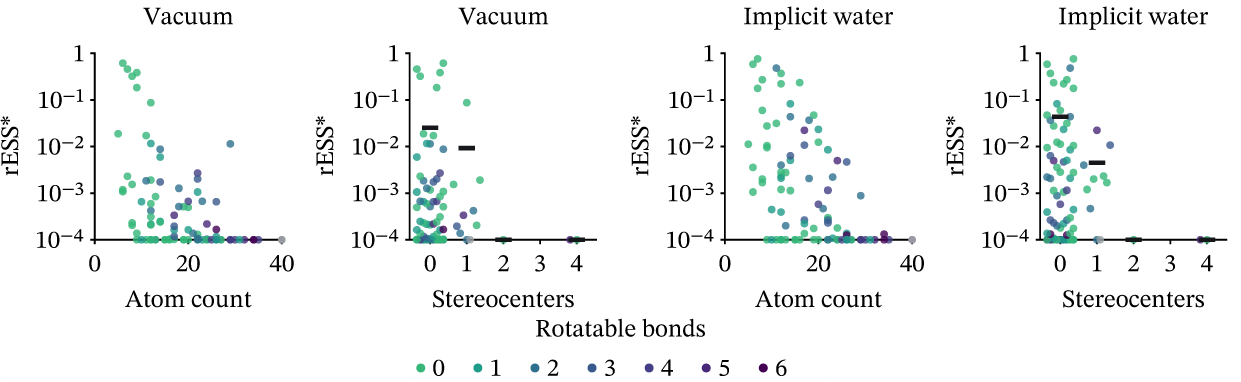}
  \caption{\textbf{FreeSolv Sampling Quality by Chemical Complexity.} Black bars show group means.}
  \label{fig:freesolv-sampling-diagnostics}
\end{figure}

\begin{figure}[htbp]
  \centering
  \includegraphics[width=\linewidth]{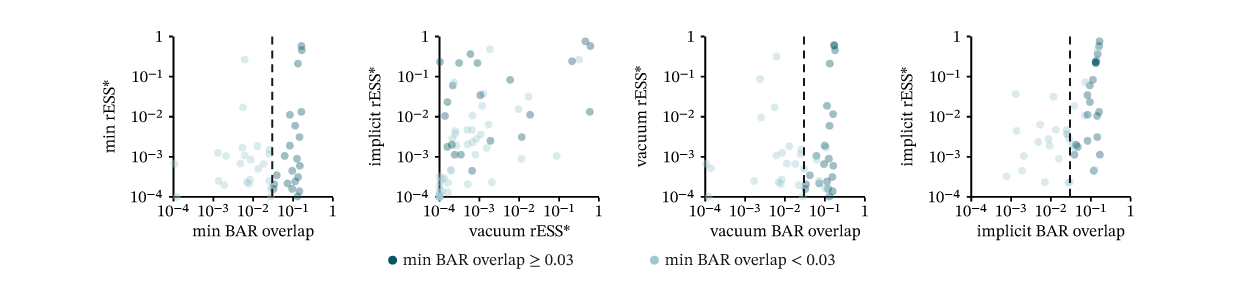}
  \caption{\textbf{FreeSolv Sampling Diagnostics.} Minimum and per-state forward \rESSast and BAR overlap. Dashed lines mark overlap $0.03$; values below the $10^{-4}$ display limits are omitted.}
  \label{fig:asfe-freesolv-supplement}
\end{figure}

\paragraph{Small Molecule Sampling.}
A prefix-conditioned Boltzmann generator, PBG-Joint, trained on a heterogeneous dataset of peptide and small molecule conformers showed comparable sampling quality on \textit{MDQM9-nc} holdout systems to dedicated models fine-tuned on this dataset (Table \ref{tab:mdqm9-native}).
Across both \textit{MDQM9-nc} (Figure \ref{fig:dipper-mdqm9-metric-atoms}) and \textit{FreeSolv} (Figures \ref{fig:freesolv-sampling-diagnostics} and \ref{fig:asfe-freesolv-supplement}) sampling quality on holdout systems is limited and decays with atom count, stereocenters and the number of rotatable bonds.

\sbox{\catalogprobebox}{%
  \renewcommand{\resizebox}[3]{#3}%
  \normalfont\normalsize
  \input{generated/dipper_mdqm9_table.tex}}
\setlength{\mdqmtablewidth}{\fpeval{min(1,\textwidth/\catalogmaxwd)}\wd\catalogprobebox}
\begin{figure}[H]
\begin{minipage}[t]{\mdqmtablewidth}
  \vspace{0pt}
  \centering
  \captionof{table}{\textbf{Transferable Sampling of Small Molecules in MDQM9-nc.} Shown are results from fine-tuned 47M and 85M PBG models and the PBG-Joint 160M model at 300K. Molecules are grouped into Easy ($n\le19$ atoms and $\le3$ rotatable bonds; 42 systems) and Hard (53 systems). Best values are \textbf{bold}; second-best are \underline{underlined}.}
  \label{tab:mdqm9-native}
  \resizebox{\linewidth}{!}{\usebox{\catalogprobebox}}
\end{minipage}\hfill
\begin{minipage}[t]{\dimexpr\textwidth-\mdqmtablewidth-1em\relax}
  \vspace{0pt}
  \centering
  \includegraphics[width=\linewidth]{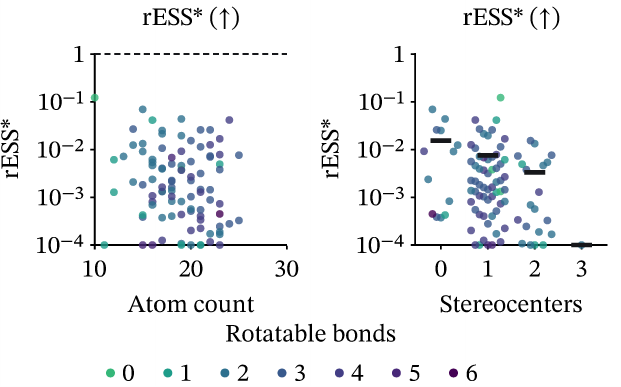}
  \captionof{figure}{\textbf{MDQM9-nc Sampling Quality by Chemical Complexity.} Black bars show group means.}
  \label{fig:dipper-mdqm9-metric-atoms}
\end{minipage}
\end{figure}

\begin{table}[H]
  \centering
  \caption{\textbf{Transferable Sampling of Small Molecules in FreeSolv.} Shown are results from the PBG-Joint 160M model at 300K.}
  \label{tab:freesolv-sampling}
  \input{generated/dipper_freesolv_table.tex}
\end{table}

\paragraph{Cutoff Offset of the Implicit-Water Energies.}
\label{par:asfe-cutoff}
The implicit-water energies use a 2\,nm cutoff, but the vacuum energies do not. Inside the cutoff, OpenMM shifts the Coulomb and generalized Born pair energies, so each implicit-water energy carries a nearly constant offset.
While the offset does not change the samples or normalized importance weights, it shifts the solvation free energy by the same amount. We measured it on implicit-water MD frames (median \FreeSolvCutoffOffsetMedian\,kcal/mol for FreeSolv and \PeptideCutoffOffsetMedian\,kcal/mol for peptides) and removed it from all solvation free energies in this paper.

\paragraph{Forcefield and Implicit Solvation Bias.}
To separate the error of PBG-Joint from that of the underlying solvation model, we also computed an MD-only reference with BAR between 5\,ns Langevin MD runs in the two potentials. Its mean absolute error against experiment was \FreeSolvMdReferenceMae\,kcal/mol (\FreeSolvMdReferenceCount{} molecules).

\newpage
\paragraph{Peptide Solvation Free Energy.}
\begin{wrapfigure}{R}{0.5\textwidth}
  \centering
  \includegraphics[width=\linewidth]{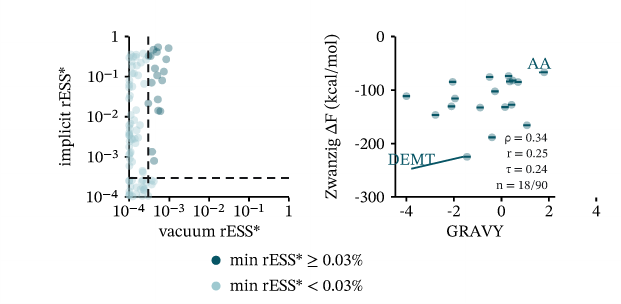}
  \caption{\textbf{Peptide Forward Zwanzig Estimates.} Vacuum versus implicit-water forward \rESSast (left) and unclipped forward $\Delta F$ versus GRAVY (right). Correlations require \rESSast $\ge0.03\%$ in both states. Free energies are in kcal/mol. Bars show propagated analytical standard errors.}
  \label{fig:asfe-forward}
  \label{fig:asfe-forward-peptides}
\end{wrapfigure}
Analogous to the previous results for small molecules, a prefix-conditioned Boltzmann generator, PBG-Joint, trained on a heterogeneous dataset of peptide and small molecule conformers showed comparable sampling quality to a dedicated fine-tuned model (Table \ref{tab:dipper-mdqm9-metric-atoms}).

\begin{figure}[htbp]
  \centering
  \includegraphics[width=0.5\linewidth]{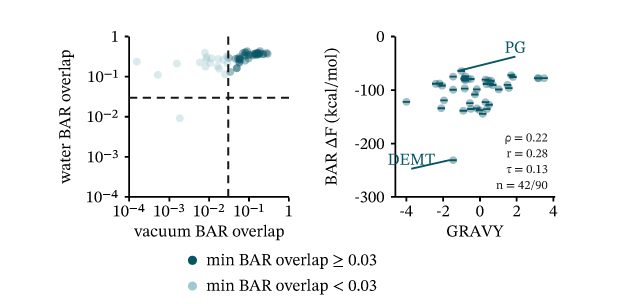}
  \caption{\textbf{Peptide BAR Estimates.} PBG-Joint 160M on the \AsfeTotalCount{} \textit{Many Peptides} test systems: vacuum versus implicit-water BAR overlap (left) and BAR $\Delta F$ versus GRAVY (right). Correlations require both BAR overlaps $\ge0.03$. Free energies are in kcal/mol. Error bars show propagated state uncertainty.}
  \label{fig:asfe-peptide-bar}
\end{figure}

\begin{figure}[htbp]
  \centering
  \includegraphics[width=\linewidth]{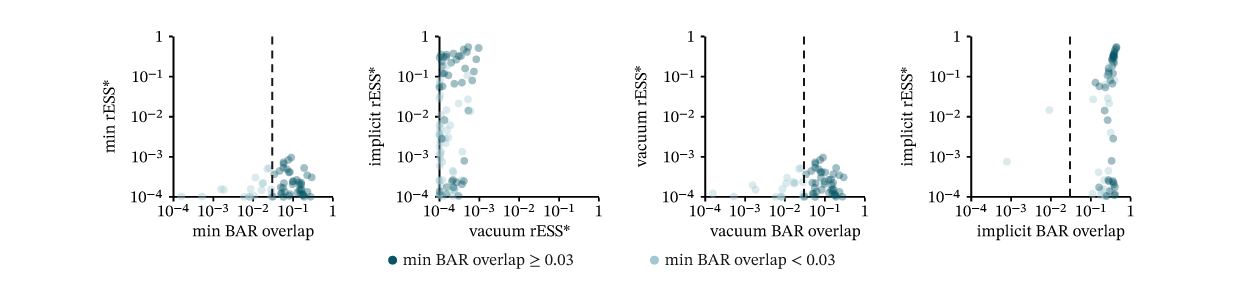}
  \caption{{\textbf{Peptide Sampling Diagnostics.} Minimum and per-state \rESSast and BAR overlap. Dashed lines mark overlap $0.03$; values below the $10^{-4}$ display limits are omitted.}}
  \label{fig:asfe-peptide-supplement}
\end{figure}

\rebuttal{Given that no experimental solvation free energies are publicly available for these peptides, we compared the estimates with GRAVY hydropathy scores instead.}
BAR estimates for the \textit{Many Peptides} test systems reached overlaps $\ge0.03$ in both states for \AsfeQualifiedCount{} of \AsfeTotalCount{} peptides and correlated weakly with Kyte--Doolittle GRAVY hydropathy scores~\cite{kyte1982simple} ($r=\AsfeGravyBarPearson$; Figure~\ref{fig:asfe-peptide-bar}, per-state overlaps in Figure~\ref{fig:asfe-peptide-supplement}).

In contrast to BAR estimates, forward-only estimation of solvation free energies using the Zwanzig equation does not require MD samples of the system of interest (Figure~\ref{fig:asfe-forward-peptides}).
Estimates of solvation free energy on \textit{Many Peptides} holdout systems correlated with GRAVY hydropathy scores.

\par\WFclear

\begin{table}[H]
  \centering
  \caption{\textbf{Vacuum Peptide Sampling across Training Conditions.} Implicit-water, vacuum-fine-tuned, and joint models evaluated at $300\,\mathrm{K}$. Superscripts mark partial coverage. Best values are \textbf{bold}; second-best are \underline{underlined}.}
  \label{tab:dipper-mdqm9-metric-atoms}
  \input{generated/dipper_vacuum_peptides_table.tex}
\end{table}

\begin{figure}[htbp]
  \centering
  \includegraphics[width=0.5\linewidth]{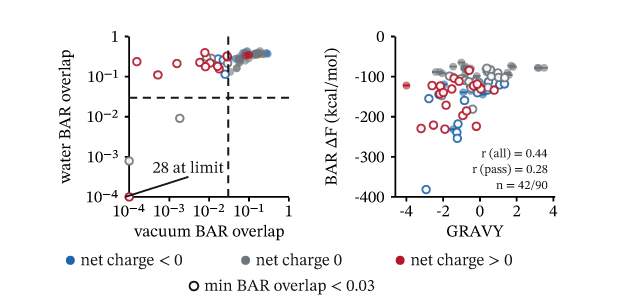}
  \caption{\rebuttal{\textbf{Peptide Solvation Free Energy by Net Charge.} Version of Figure~\ref{fig:asfe-peptide-bar} with all \AsfeTotalCount{} test peptides. Color marks the net charge (Lys and Arg $+1$, Asp and Glu $-1$, His neutral). Open circles mark peptides with a BAR overlap below $0.03$ in at least one state. Error bars show propagated state uncertainty and are shown only for the peptides that pass. Overlaps below $10^{-4}$ are drawn at the axis limit.}}
  \label{fig:asfe-peptide-charge}
\end{figure}

\newpage
\paragraph{\rebuttal{Charged Peptides.}}
\rebuttal{Fewer charged peptides than net-neutral peptides pass both overlap thresholds (\AsfeChargedQualifiedCount{} of \AsfeChargedCount{}, against \AsfeNeutralQualifiedCount{} of \AsfeNeutralCount{}; Table~\ref{tab:asfe-peptide-charge} and Figure~\ref{fig:asfe-peptide-charge}).}

\begin{table}[htbp]
  \centering
  \caption{\rebuttal{\textbf{Peptide Solvation Free Energy by Net Charge.} PBG-Joint 160M on the test peptides of Figure~\ref{fig:asfe-peptide-bar}, grouped by net charge.}}
  \label{tab:asfe-peptide-charge}
  \input{generated/asfe_peptide_charge_table.tex}
\end{table}

\clearpage
\subsection{Explicit Microsolvation}
\begin{figure}[H]
  \centering
  \includegraphics[width=\linewidth]{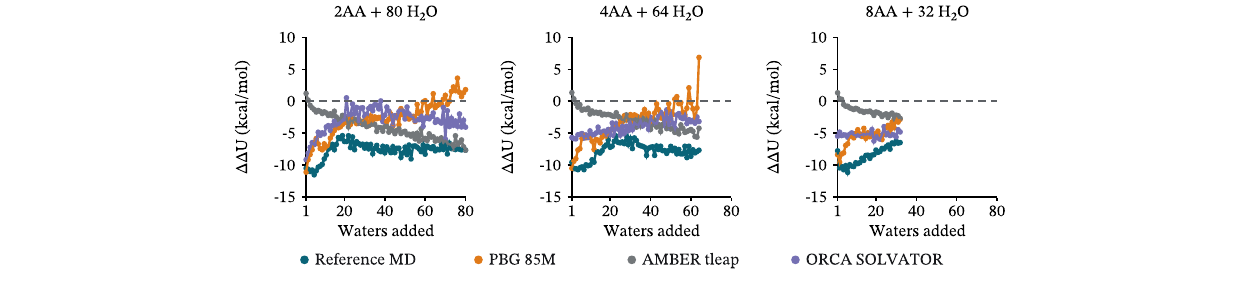}
  \caption{\textbf{Energy Change per Added Water.} Median potential energy differences  $\Delta U(k)=U(k)-U(0)$ of nearest neighboring water insertion. The dashed line marks zero energy change. MD reference and AMBER tleap waters are ordered by the distance from the water geometric center to the nearest solute atom. ORCA SOLVATOR and PBG waters follow addition order. Energies use the non-periodic Amber14/TIP3P reaction-field potential with a $2\,\mathrm{nm}$ cutoff.}
  \label{fig:dipper-solvated-water-ladder}
\end{figure}

Stepwise microsolvation (Figure~\ref{fig:dipper-solvated-water-ladder}) shows early plateauing of potential energy changes as waters from reference MD runs are added while microsolvation by a fine-tuned PBG 85M leads to steadily worse energetic changes with eventual positive potential energy contribution.

\clearpage
\section{Datasets}
\label{sec:datasets}

\paragraph{Many Peptides.}
The \textit{Many Peptides} dataset was used in its precollated webdataset format.
The reference trajectory of the octapeptide YFPHAGYT has a D-proline that is not in its published starting structure, a stereochemical artifact that also affects the training data (Appendix~\ref{sec:yfphagyt}).
We excluded this peptide as a result from the test set, leaving 29 octapeptide test systems in most evals.
The paragraph symbol (\P) marks evaluation summaries that are affected.
The literature rows in Table~\ref{tab:sota} use the same 29 octapeptides. Thus their octapeptide values differ from the 30-system means that \citet{tan2026amortizedsamplingtransferablenormalizing} and \citet{rehmann2025autoregressive} published.
Appendix~\ref{sec:yfphagyt} explains this artifact and repeats Table~\ref{tab:sota} on all 30 octapeptides with an adjusted stereochemistry filter.

\paragraph{MDQM9-nc.}
This dataset from \citet{mdqm9} was collated into a shuffled webdataset for training.
\publictext{The dataset is available on Hugging Face as \href{https://huggingface.co/datasets/niklastr/smiles_mdqm9nc}{\texttt{niklastr/smiles\_mdqm9nc}}.}{The dataset will be released on Hugging Face upon acceptance.}

\paragraph{Vacuum Peptides.}
We initialized vacuum simulations from conformers of
\textit{Many Peptides}, retaining its train/validation/test assignments. The simulations used Amber
ff14SB with constrained bonds to hydrogen and no nonbonded cutoff.
\publictext{The dataset is available on Hugging Face as \href{https://huggingface.co/datasets/niklastr/vacuum_peptides}{\texttt{niklastr/vacuum\_peptides}}.}{The dataset will be released on Hugging Face upon acceptance.}

\paragraph{Microsolvated Peptides.}
We placed peptide conformers from \textit{Many Peptides} in
periodic TIP3P water boxes with 1.0\,nm solvent padding. Simulations
used Amber ff14SB, constrained bonds to hydrogen, rigid water, and
particle-mesh Ewald electrostatics with a 0.9\,nm real-space cutoff.
Box dimensions remained fixed. We cropped each saved configuration
to the solute and 128 nearby complete water molecules. For training,
we ordered the retained waters by the minimum distance from each
water oxygen to a solute atom.
We retained train/validation/test assignments from \textit{Many Peptides}.
\rebuttal{Model training used only the net-neutral peptides of this dataset, as the water box of a charged peptide also contains one Na$^{+}$ or Cl$^{-}$ counterion per unit of net charge, leading to counterions enclosed within the nearest 128 water molecules in an irregular fashion.
Table~\ref{tab:counterion-audit} shows the fraction of frames in which the solute was in contact with a counterion. A cropped cluster of a charged peptide would therefore contain its counterion in some frames and not in others.
For \CounterionMixedCount{} of the \CounterionChargedCount{} charged test peptides, the counterion is inside the 128-water cluster in \CounterionMixedLowPercent--\CounterionMixedHighPercent{} of the frames.
For this reason, the microsolvated data contains only net-neutral peptides.
\publictext{The dataset is available on Hugging Face as \href{https://huggingface.co/datasets/niklastr/solvated_peptides}{\texttt{niklastr/solvated\_peptides}}.}{The dataset will be released on Hugging Face upon acceptance.}}

\begin{table}[htbp]
  \centering
  \caption{\rebuttal{\textbf{Counterions in the Explicit-Water MD of Charged Test Peptides.} Each periodic box of the \CounterionChargedCount{} charged test peptides holds one Na$^{+}$ or Cl$^{-}$ counterion per unit of net charge. Contact: the counterion is within $0.4\,\mathrm{nm}$ of a peptide heavy atom; a randomly placed point is in contact in \CounterionRandomContactPercent{} of frames. Inside cluster: the counterion is closer to the peptide than the 128th nearest water, so a 128-water cluster would contain it. Mixed: the counterion is inside the cluster in \CounterionMixedLowPercent--\CounterionMixedHighPercent{} of the frames. Percentages are medians over peptides, with the range in parentheses.}}
  \label{tab:counterion-audit}
  \input{generated/microsolvated_counterion_audit.tex}
\end{table}

\paragraph{FreeSolv.}
We generated vacuum, implicit-water, and explicit-water
ensembles for molecules from FreeSolv.
Vacuum simulations used
the original FreeSolv GAFF parameters and AM1-BCC charges, with
constrained bonds to hydrogen and no nonbonded cutoff. Implicit-water
simulations used GAFF2 with Antechamber AM1-BCC charges, OBC1 implicit solvent, a solvent dielectric of 78.5, and a 2.0\,nm cutoff. These
systems had no bond constraints. Explicit-water simulations used
GAFF2/AM1-BCC and TIP3P, with constrained bonds to hydrogen, rigid
water, and particle-mesh Ewald electrostatics with a 1.0\,nm real-space
cutoff.
The Lennard--Jones switching function began at 0.9\,nm.
The microsolvated export retained 128 complete waters per frame,
ordered by minimum oxygen-to-solute distance.
We retained train/validation/test assignments from \citet{mobley_guthrie_2014_freesolv}.
\publictext{The dataset is available on Hugging Face as \href{https://huggingface.co/datasets/niklastr/freesolv}{\texttt{niklastr/freesolv}}.}{The dataset will be released on Hugging Face upon acceptance.}

\paragraph{Many Cycles.}
We simulated head-to-tail cyclic peptides using Amber ff14SB
with OBC1 implicit solvent, a solvent dielectric of 78.5, a 1.0\,nm
cutoff, and constrained bonds to hydrogen.
The exported trajectories
used eight temperature states from 300 to 500\,K, a 2\,fs Langevin
timestep, and 10\,ps exchange and save intervals. Each exported run
contained 100\,ns per replica.
\publictext{The dataset is available on Hugging Face as \href{https://huggingface.co/datasets/niklastr/many_cycles}{\texttt{niklastr/many\_cycles}}.}{The dataset will be released on Hugging Face upon acceptance.}

\begin{table}[H]
  \centering
  \caption{\textbf{Overview of Training Datasets.} Tokens count three coordinate tokens per atom per conformer. Atom counts are min / median / max over distinct training systems.}
  \label{tab:datasets}
  \setlength{\tabcolsep}{3.5pt}
  \resizebox{\linewidth}{!}{%
  \begin{tabular}{@{}lrrrrccc@{}}
    \toprule
    Dataset & Conformers & Molecules & Conf./mol. & Tokens & Min & Median & Max \\
    \midrule
    Many Peptides
      & $4.34 \times 10^{8}$
      & 21{,}700
      & $2.00 \times 10^{4}$
      & $1.45 \times 10^{11}$
      & 17 & 117 & 178 \\
    MDQM9-nc
      & $1.969 \times 10^{8}$
      & 12{,}306
      & $1.60 \times 10^{4}$
      & $1.08 \times 10^{10}$
      & 3 & 18 & 29 \\
    Vacuum Peptides
      & $2.140 \times 10^{7}$
      & 11{,}852
      & $1.81 \times 10^{3}$
      & $7.09 \times 10^{9}$
      & 17 & 116 & 176 \\
    Microsolvated Peptides
      & $1.245 \times 10^{7}$
      & 7{,}340
      & $1.70 \times 10^{3}$
      & $1.85 \times 10^{10}$
      & 401 & 500 & 559 \\
    FreeSolv (vacuum)
      & $6.535 \times 10^{6}$
      & 508
      & $1.29 \times 10^{4}$
      & $3.56 \times 10^{8}$
      & 3 & 18 & 44 \\
    FreeSolv (implicit water)
      & $6.440 \times 10^{6}$
      & 508
      & $1.27 \times 10^{4}$
      & $3.51 \times 10^{8}$
      & 3 & 18 & 44 \\
    FreeSolv (microsolvated)
      & $2.158 \times 10^{6}$
      & 262
      & $8.24 \times 10^{3}$
      & $2.60 \times 10^{9}$
      & 389 & 402 & 428 \\
    \bottomrule
  \end{tabular}%
  }
\end{table}

\newpage
\section{Evaluation Metrics}

\paragraph{BAR bridge weight.}
\label{metrics:bar-overlap}
The sample-origin probabilities in~\eqref{eq:bar} define the BAR bridge weights. At the fitted free-energy difference $\widehat{\Delta F}$, each sample receives the probability of belonging to the other state,
\begin{equation}
  a_i^{A}
  = p_{\widehat{\Delta F}}(B\mid\mathbf{x}_i^A),
  \qquad
  a_j^{B}
  = 1-p_{\widehat{\Delta F}}(B\mid\mathbf{x}_j^B).
\end{equation}
Naturally, a weight approaching $0$ indicates confident assignment to the sample's source state. The BAR overlap reported is the smaller mean bridge weight,
\begin{equation}
  O_{\mathrm{BAR}}
  = \min\!\left(
      \frac{1}{n_A}\sum_{i=1}^{n_A}a_i^{A},
      \frac{1}{n_B}\sum_{j=1}^{n_B}a_j^{B}
    \right).
  \label{eq:bar-overlap}
\end{equation}
Thus $O_{\mathrm{BAR}}\in[0,1]$, with small values indicating little sampled overlap and warning of unreliable free-energy estimates. The metric depends on the retained sample counts: identical distributions give $O_{\mathrm{BAR}}=\min(n_A,n_B)/(n_A+n_B)$, which equals $1/2$ for equal counts.
For solvation free energies, we compute it on each Zero Free Energy Proposal leg and use the minimum of the vacuum and water values as the overall overlap diagnostic.
\citet{Klimovich2015GuidelinesFT} suggest that entries as low as $0.03$ can be tolerable with enough samples and sufficiently small uncertainty, while smaller values warrant caution.
We therefore require $O_{\mathrm{BAR}}\ge 0.03$ on both legs for the correlation analysis, while acknowledging that this screening criterion does not guarantee convergence.

\paragraph{Sliced Torus-Wasserstein.} \backlink{par:metrics}
\label{metrics:sliced}
We consider each conformation's vector $\boldsymbol\theta\in[0,2\pi)^d$ of $d$ dihedral angles. To identify them, we enumerate all bonded four-atom paths whose central bond joins two non-hydrogen atoms, counting a path and its reverse only once. The selection is not restricted to backbone angles or rotatable bonds.

We compare one-dimensional circular projections, called slices.
The integer vector $k=e_i$ selects angle $i$, while $k=e_i\pm e_j$
selects a pairwise sum or difference ($1\leq i<j\leq d$);
$e_i$ is the $i$th unit vector.
Let $K$ be the retained slice pool.
For each slice $k\in K$, define its projected coordinate
$z_k(\boldsymbol\theta)$, circle circumference $L_k$, and normalized
slice weight $a_k$:
\begin{equation}
  z_k(\boldsymbol\theta)
  = \frac{(k^\top\boldsymbol\theta)\bmod 2\pi}{\|k\|_2},
  \qquad L_k=\frac{2\pi}{\|k\|_2},
  \qquad
  a_k=\frac{(1+\|k\|_2^2)^{-(d/2+2)}}
  {\sum_{h\in K}(1+\|h\|_2^2)^{-(d/2+2)}}.
\end{equation}
Here $\|k\|_2$ is $1$ for individual angles and $\sqrt{2}$
for pairs. Let $\mu_k$ and $\nu_k$ be the generated and reference
distributions of $z_k$. Valid generated conformers receive the
normalized importance weights from~\eqref{eq:bg-snis};
reference frames receive equal weights. The reported quantity estimates
\begin{equation}
  \TgSWast
  \approx \left[\sum_{k\in K} a_k\,
  W_{2,\mathrm{circ}(L_k)}^2(\mu_k,\nu_k)\right]^{1/2}.
  \label{eq:sliced-torus-w2}
\end{equation}
Circular $W_2$ is the square root of the least mean squared distance needed to move probability mass between the two distributions, using the shorter path around the circle. Thus the metric is a weighted root-mean-square discrepancy in radians. The slice weights $a_k$ favor individual angles over angle pairs, increasingly so as $d$ grows.

We evaluate all retained individual-angle slices and use up to
256 pair draws for peptides, or 128 for small molecules.
Pair draws use replacement, with probabilities proportional to $a_k$.
The candidate pools contain at most 4096 and 1024 slices, respectively,
in deterministic order, with individual angles first.
Large weighted distributions are approximated by 1024 equally weighted
points selected at fixed weighted quantiles.

\newpage
\section{Methods}
\label{sec:experimental-methods}
\paragraph{Replica-exchange simulations.}
We generated the additional peptide and FreeSolv datasets using replica-exchange molecular dynamics in OpenMM with openmmtools.
Briefly, eight replicas covered geometrically spaced temperatures of 300.00, 322.71, 347.14, 373.42, 401.69, 432.10, 464.81, and 500.00\,K.
We ran Langevin dynamics with a 2\,fs timestep and a collision rate of $0.3\,\mathrm{ps}^{-1}$.
We attempted replica exchanges and saved configurations every 5,000 steps (10\,ps).
The peptide and FreeSolv runs
used 200 production intervals, corresponding to 2\,ns per replica
per simulation burst.
For training, simulation data was collated into pre-shuffled webdatasets.

\paragraph{Architecture.}
We use decoder-only transformers with causal grouped-query attention, pre-layer RMSNorm ($\epsilon=10^{-6}$), query/key normalization, and SwiGLU feed-forward blocks.
Input and output token embeddings were tied.
Table~\ref{tab:trained-architectures} lists the architectures used in the scaling and application experiments.
We refer to model size based on non-embedding parameters.
The smaller PBGs were randomly initialized.

\paragraph{Encoding.}
Peptides are specified by PEPTIDE HELM strings; small molecules use SMILES embedded in CHEM HELM strings.

At the default resolution, the implementation rounds coordinates to 4096 equally spaced grid points with spacing $0.8\,\mathrm{pm}$ and endpoints $\pm1.638\,\mathrm{nm}$ on each axis. This endpoint-inclusive implementation uses $b-1$ intervals between grid points. Evaluation reconstructs the selected grid points deterministically using midpoint dequantization.

Half of the RoPE frequency pairs encode atom indices and half encode token positions. The RoPE base is $10^3$.
Models also add a fixed sinusoidal coordinate-axis (x, y, or z dimension) encoding with base 10.
Conformations are greedily packed into rows of at most 1028 tokens, with attention restricted to each conformation's segment.

\paragraph{Training.}
We minimize teacher-forced next-token cross-entropy on coordinate tokens and the terminator, excluding prompt and padding tokens. Thus a complete $n$-atom conformation contributes $3n+1$ loss tokens when a terminator is present, and $3n$ otherwise. We report $D$ as delivered loss tokens, rather than padded sequence capacity or the nominal checkpoint name.
Dataset inventories are given in Table~\ref{tab:datasets}.

The dedicated 85M vacuum and microsolvation models continue from the pre-trained \textit{Many Peptides} checkpoint on their respective $300$--$500\,\mathrm{K}$ datasets and exhausted all data within these datasets, respectively. For microsolvation, only complete water molecules that fit within the context are retained.

\paragraph{Optimizer, Learning Rate, Scheduler.}
We used Muon for non-embedding weight matrices and AdamW for embeddings and remaining parameters~\cite{jordan2024muon,liu2025muonscalablellmtraining}. Muon uses Nesterov momentum 0.95 and five Newton--Schulz iterations. AdamW uses $(\beta_1,\beta_2)=(0.9,0.95)$ and $\epsilon=10^{-8}$. Generally training runs use a peak learning rate of $10^{-3}$ and decoupled weight decay 0.1.
Gradient clipping was initially performed with an initial global-norm limit of 10, then a per-tensor relative limit of 1.5 times its moving-average norm (decay 0.99) after 100 steps.

Warmup--stable--decay schedules~\cite{wsdscalinglaw} permit several training budgets to share a stable-phase trajectory. A cooldown typically added 20\% of the preceding token budget and reduced the learning rate to zero with a $1-\sqrt{t}$ profile, where $t\in[0,1]$ is cooldown progress.

\paragraph{Sampling and Evaluation.}
We draw $10^4$ proposals per test system at logit temperature 1.0 with KV-cached autoregressive decoding and deterministic grid-point reconstruction.
During evaluation we correct a fully mirrored stereochemical assignment by global reflection of the conformer, while retaining the initially assigned likelihood.
Remaining invalid stereoisomers are excluded (on average \StereoExcludedMeanPercent\% of samples, Appendix~\ref{sec:stereo}).
Importance weights use the target reduced potential and proposal likelihood. Starred metrics retain unclipped weights.

The fixed-solute microsolvation comparison uses the same 512 reference solute conformers per system for all methods and a non-periodic Amber14/TIP3P energy with a $2\,\mathrm{nm}$ reaction-field cutoff.

\paragraph{Bayesian Inference.}
\label{par:bayesian}
Bayesian inference was performed with NumPyro NUTS using four chains, 2000 warmup steps and 2000 retained draws per chain. The target acceptance was 0.99, and maximum tree depth was set to 12.

\paragraph{Hardware.}

\begin{table}[htbp]
  \centering
  \caption{\textbf{Selected model training configurations.} Recorded allocations for the runs producing the PBG checkpoints in Table~\ref{tab:sota}. Batch sizes count packed rows of at most 1028 tokens, not conformers. Each MI250X GPU exposes two compute devices, while each NVIDIA GPU exposes one.}
  \label{tab:sota-training-configuration}
  \input{generated/sota_training_configuration.tex}
\end{table}

The training implementation uses JAX with bfloat16 forward computation and float32 master parameters. Experiments use NVIDIA and AMD GPU clusters, including NVIDIA GH200 nodes on Arrhenius and AMD MI250X accelerators on LUMI-G. Hardware and batch allocation vary by run. Training is parallelized using a simple distributed data parallel approach. \rebuttal{Training of the \fittingcohortsize{} checkpoints of the scaling model took \ScalingFitGpuHours{} GPU-hours on RTX PRO 6000 and H100 NVL GPUs in \ScalingFitRunCount{} runs, with shared parent runs counted once.} The reported autoregressive and speculative-decoding throughput comparison uses an H100 NVL GPU at batch size 512.

\ifpaperworkshop
  \clearpage
  \input{checklist.tex}
\fi

\end{document}

%% file: generated/queries.tex
\newcommand{\DipperSpecEightyFiveMMeanSpeedup}{1.27}
\newcommand{\DipperSpecEightyFiveMMaxSpeedup}{1.64}

\newcommand{\DipperSpecSixFortyMMeanSpeedup}{1.55}
\newcommand{\DipperSpecSixFortyMMaxSpeedup}{2.14}

\newcommand{\SeedVarRatioRess}{7}

\newcommand{\SeedVarRatioEnergy}{2}

\newcommand{\SeedVarRatioTorus}{2}

\newcommand{\SeedVarRatioTica}{1}

\newcommand{\StereoExcludedMeanPercent}{1.2}

\newcommand{\StereoReflectedOfKeptMinPercent}{50.0}

\newcommand{\StereoLadderCentredMixedCount}{1}
\newcommand{\StereoLadderNoneMaxPercent}{2.0}

\newcommand{\StereoScalingExcludedComputeCorrelation}{$-0.81$}

\newcommand{\StereoPbgSamplesMillion}{18.69}
\newcommand{\StereoNonfiniteFrames}{1}

\newcommand{\DproTrainPeptides}{21{,}700}
\newcommand{\DproTrainProPeptides}{6{,}223}
\newcommand{\DproTrainAffected}{276}
\newcommand{\DproTrainAffectedPercent}{4.4}

\newcommand{\DproYfphagytPbgMin}{90.9}
\newcommand{\DproYfphagytPbgMax}{99.6}
\newcommand{\DproYfphagytRobinMin}{98.7}
\newcommand{\DproYfphagytRobinMax}{99.0}
\newcommand{\DproYfphagytLMax}{0.32}

\newcommand{\DproRulePrecision}{95}
\newcommand{\DproRuleRecall}{98}

\newcommand{\DproMinimizedAgree}{2{,}055}

\newcommand{\DproExampleSequence}{FPL}

\newcommand{\DproExampleStartEnergy}{8.3\times10^{8}}

\newcommand{\DipperScalingAblationHeldoutNllRmseMin}{\ensuremath{0.00277}}
\newcommand{\DipperScalingAblationHeldoutNllRmseMedian}{\ensuremath{0.00740}}
\newcommand{\DipperScalingAblationHeldoutNllRmseMax}{\ensuremath{0.0334}}

\newcommand{\DipperScalingAblationParityMin}{\ensuremath{3.52}}
\newcommand{\DipperScalingAblationParityMedian}{\ensuremath{4.84}}
\newcommand{\DipperScalingAblationParityMax}{\ensuremath{6.87}}
\newcommand{\DipperScalingAblationEnergyAtomMin}{\ensuremath{2.06}}
\newcommand{\DipperScalingAblationEnergyAtomMedian}{\ensuremath{2.17}}
\newcommand{\DipperScalingAblationEnergyAtomMax}{\ensuremath{2.39}}

\newcommand{\DipperScalingAblationTppCoreMin}{\ensuremath{21.2}}
\newcommand{\DipperScalingAblationTppCoreMedian}{\ensuremath{36.2}}
\newcommand{\DipperScalingAblationTppCoreMax}{\ensuremath{180}}
\newcommand{\DipperScalingAblationTppUnconstrained}{\ensuremath{4.8\times10^{4}}}
\newcommand{\DipperScalingAblationTppRobinMin}{\ensuremath{19.6}}
\newcommand{\DipperScalingAblationTppRobinMedian}{\ensuremath{34.8}}
\newcommand{\DipperScalingAblationTppRobinMax}{\ensuremath{301}}
\newcommand{\DipperScalingAblationRobinParamsMillion}{\ensuremath{132}}
\newcommand{\DipperScalingAblationNaiveTpp}{\ensuremath{63.4}}
\newcommand{\DipperScalingAblationNaiveTppMin}{\ensuremath{30.4}}

\newcommand{\DipperScalingAblationNaiveTppMax}{\ensuremath{168}}
\newcommand{\DipperScalingAblationUntiedTpp}{\ensuremath{44.8}}
\newcommand{\DipperScalingAblationUntiedTppMin}{\ensuremath{25.8}}

\newcommand{\DipperScalingAblationUntiedTppMax}{\ensuremath{53.5}}

\newcommand{\DipperScalingAblationSystemEnergyAtom}{\ensuremath{2.37}}
\newcommand{\DipperScalingAblationSystemEnergyAtomLow}{\ensuremath{2.03}}
\newcommand{\DipperScalingAblationSystemEnergyAtomHigh}{\ensuremath{2.72}}
\newcommand{\DipperScalingAblationSlicedAtomMin}{\ensuremath{0.753}}
\newcommand{\DipperScalingAblationSlicedAtomMedian}{\ensuremath{1.01}}
\newcommand{\DipperScalingAblationSlicedAtomMax}{\ensuremath{1.08}}

\newcommand{\DipperScalingAblationCensorOneEss}{\ensuremath{1.1}}

\newcommand{\DipperMdqmScalingCheckpointCount}{36\xspace}
\newcommand{\DipperMdqmScalingHeldoutCount}{4\xspace}
\newcommand{\DipperMdqmScalingSizeMin}{8M\xspace}
\newcommand{\DipperMdqmScalingSizeMax}{160M\xspace}
\newcommand{\DipperMdqmScalingTrainedTppMin}{\ensuremath{12}}
\newcommand{\DipperMdqmScalingTrainedTppMax}{\ensuremath{1{,}152}}

\newcommand{\DipperMdqmScalingFitSizeMin}{8M\xspace}
\newcommand{\DipperMdqmScalingFitSizeMax}{85M\xspace}
\newcommand{\DipperMdqmScalingAblationHeldoutNllRmseMin}{\ensuremath{0.00604}}
\newcommand{\DipperMdqmScalingAblationHeldoutNllRmseMedian}{\ensuremath{0.00962}}
\newcommand{\DipperMdqmScalingAblationHeldoutNllRmseMax}{\ensuremath{0.0220}}

\newcommand{\DipperMdqmScalingAblationTppMin}{\ensuremath{12.7}}
\newcommand{\DipperMdqmScalingAblationTppMedian}{\ensuremath{21.2}}
\newcommand{\DipperMdqmScalingAblationTppMax}{\ensuremath{3.7\times10^{4}}}
\newcommand{\DipperMdqmScalingAblationTppCoreMin}{\ensuremath{12.7}}
\newcommand{\DipperMdqmScalingAblationTppCoreMedian}{\ensuremath{21.1}}
\newcommand{\DipperMdqmScalingAblationTppCoreMax}{\ensuremath{73.5}}
\newcommand{\DipperMdqmScalingAblationSystemEnergyAtom}{\ensuremath{1.73}}
\newcommand{\DipperMdqmScalingAblationSystemEnergyAtomMin}{\ensuremath{1.71}}
\newcommand{\DipperMdqmScalingAblationSystemEnergyAtomMax}{\ensuremath{1.75}}
\newcommand{\DipperMdqmScalingAblationNaiveTpp}{\ensuremath{57.2}}
\newcommand{\DipperMdqmScalingAblationNaiveTppMin}{\ensuremath{47.9}}
\newcommand{\DipperMdqmScalingAblationNaiveTppMax}{\ensuremath{73.5}}
\newcommand{\DipperMdqmScalingAblationUntiedAdapterTpp}{\ensuremath{344}}
\newcommand{\DipperMdqmScalingAblationUntiedAdapterTppMin}{\ensuremath{17.4}}
\newcommand{\DipperMdqmScalingAblationUntiedAdapterTppMax}{\ensuremath{3.7\times10^{4}}}
\newcommand{\DipperMdqmScalingAblationUntiedTpp}{\ensuremath{21.0}}
\newcommand{\DipperMdqmScalingAblationUntiedTppMin}{\ensuremath{20.1}}
\newcommand{\DipperMdqmScalingAblationUntiedTppMax}{\ensuremath{21.1}}

\newcommand{\DipperScalingDstarTN}{40.5}
\newcommand{\DipperScalingDstarTNLow}{35.5}
\newcommand{\DipperScalingDstarTNHigh}{45.5}

\newcommand{\DipperScalingEnergyAtomExponent}{2.17}

\newcommand{\DipperScalingSlicedAtomExponent}{1.02}

\newcommand{\DipperScalingrESSAtomExponent}{1.79}

\newcommand{\fittingcohortsize}{130\xspace}
\newcommand{\fittingbinsizecount}{12\xspace}
\newcommand{\fittingmodelsizecount}{5\xspace}
\newcommand{\fittingtokenbudgetcount}{8\xspace}

\newcommand{\DipperScalingComputeParityExponent}{5.61}

\newcommand{\ScalingFitGpuHours}{1,259\xspace}
\newcommand{\ScalingFitRunCount}{134\xspace}

\newcommand{\AsfeTotalCount}{90}
\newcommand{\AsfeQualifiedCount}{42}
\newcommand{\AsfeGravyBarPearson}{\ensuremath{+0.28}}

\newcommand{\FreeSolvAsfeTotalCount}{94}
\newcommand{\FreeSolvAsfeQualifiedCount}{21}
\newcommand{\AsfeFreesolvPearson}{\ensuremath{+0.50}}

\newcommand{\FreeSolvBarMae}{4.9}

\newcommand{\FreeSolvZwanzigPearson}{\ensuremath{+0.64}}

\newcommand{\FreeSolvZwanzigMae}{4.2}

\newcommand{\FreeSolvMdReferenceCount}{93}
\newcommand{\FreeSolvMdReferenceMae}{2.5}

\newcommand{\FreeSolvCutoffOffsetMedian}{\ensuremath{+5.5}}

\newcommand{\AsfeNeutralCount}{46}
\newcommand{\AsfeChargedCount}{44}
\newcommand{\AsfeNeutralQualifiedCount}{29}
\newcommand{\AsfeChargedQualifiedCount}{13}

\newcommand{\PeptideCutoffOffsetMedian}{\ensuremath{+5.6}}

\newcommand{\CounterionChargedCount}{44}
\newcommand{\CounterionMixedCount}{37}
\newcommand{\CounterionRandomContactPercent}{12\%}
\newcommand{\CounterionMixedLowPercent}{10\%}
\newcommand{\CounterionMixedHighPercent}{90\%}

\newcommand{\BoxVolumeLogWeightSdEightAA}{0.16}

\newcommand{\BoxVolumeMeanShiftEightAA}{-0.51}
\newcommand{\BoxVolumePerfectRessMin}{0.965}
\newcommand{\BoxVolumeMaxChangeRessPercent}{9.4}
\newcommand{\BoxVolumeMaxChangeEnergyPercent}{4.0}
\newcommand{\BoxVolumeMaxChangeTorusPercent}{9.2}

\newcommand{\BoxVolumeModels}{13}

%% file: generated/alanine_dipeptide_solute_tokens_tikz.tex
\begin{figure}[H]
\centering
\resizebox{\linewidth}{!}{%
\begin{tikzpicture}[
  font=\small,
  solutebond/.style={line width=0.62pt,black!75},
  C/.style={circle,draw=black!70,fill=gray!65,minimum size=2.2mm,inner sep=0pt},
  N/.style={circle,draw=black!70,fill=blue!70,minimum size=2.2mm,inner sep=0pt},
  O/.style={circle,draw=black!70,fill=red!75,minimum size=2.2mm,inner sep=0pt},
  S/.style={circle,draw=black!70,fill=yellow,minimum size=2.2mm,inner sep=0pt},
  H/.style={circle,draw=black!45,fill=white,minimum size=1.05mm,inner sep=0pt},
  soluteC/.style={C,draw=black,line width=0.45pt},
  soluteN/.style={N,draw=black,line width=0.45pt},
  soluteO/.style={O,draw=black,line width=0.45pt},
  soluteS/.style={S,draw=black,line width=0.45pt},
  soluteH/.style={H,draw=black,line width=0.35pt},
  tokfont/.style={font=\ttfamily\fontsize{7pt}{8pt}\selectfont},
  tokbox/.style={
    tokfont,
    rounded corners=0.7pt,
    draw=black!50,
    line width=0.35pt,
    inner sep=0.35pt,
    minimum width=4.0mm,
    minimum height=4.0mm,
    align=center,
    anchor=west
  },
  promptTok/.style={tokbox,fill=cyan!12,text=teal!80!black,draw=teal!35!black!25},
  coordTok/.style={tokbox,fill=blue!70,text=white},
  soluteTokC/.style={tokbox,fill=gray!65},
  soluteTokN/.style={tokbox,fill=blue!70},
  soluteTokO/.style={tokbox,fill=red!75},
  soluteTokS/.style={tokbox,fill=yellow},
  soluteTokH/.style={tokbox,fill=white,draw=black!55}
]
\begin{scope}[shift={(-4.6,0.12)}]
\coordinate (a1) at (-0.726,-0.807);
\coordinate (a2) at (-1.191,-1.007);
\coordinate (a3) at (-0.593,-1.003);
\coordinate (a4) at (-0.294,-0.863);
\coordinate (a5) at (-0.786,-0.193);
\coordinate (a6) at (-0.505,0.085);
\coordinate (a7) at (-1.593,-0.087);
\coordinate (a8) at (-1.563,0.382);
\coordinate (a9) at (-1.906,-0.315);
\coordinate (a10) at (-1.853,-0.287);
\coordinate (a11) at (-1.352,-0.006);
\coordinate (a12) at (-0.504,-0.041);
\coordinate (a13) at (-0.950,0.003);
\coordinate (a14) at (0.204,0.013);
\coordinate (a15) at (0.527,-0.024);
\coordinate (a16) at (0.573,0.164);
\coordinate (a17) at (0.226,0.065);
\coordinate (a18) at (1.182,-0.175);
\coordinate (a19) at (1.310,-0.125);
\coordinate (a20) at (1.640,-0.025);
\coordinate (a21) at (1.009,-0.823);
\coordinate (a22) at (0.864,-0.878);
\coordinate (a23) at (1.473,-1.061);
\coordinate (a24) at (0.333,-1.073);
\coordinate (a25) at (0.272,-0.987);
\coordinate (a26) at (-0.174,-1.382);
\coordinate (a27) at (0.778,0.825);
\coordinate (a28) at (0.981,1.142);
\coordinate (a29) at (0.778,1.019);
\draw[solutebond] (a1) -- (a2);
\draw[solutebond] (a1) -- (a3);
\draw[solutebond] (a1) -- (a4);
\draw[solutebond] (a1) -- (a5);
\draw[solutebond] (a5) -- (a6);
\draw[solutebond] (a5) -- (a7);
\draw[solutebond] (a5) -- (a12);
\draw[solutebond] (a7) -- (a8);
\draw[solutebond] (a7) -- (a9);
\draw[solutebond] (a7) -- (a10);
\draw[solutebond] (a10) -- (a11);
\draw[solutebond] (a12) -- (a13);
\draw[solutebond] (a12) -- (a14);
\draw[solutebond] (a14) -- (a15);
\draw[solutebond] (a14) -- (a16);
\draw[solutebond] (a16) -- (a17);
\draw[solutebond] (a16) -- (a18);
\draw[solutebond] (a16) -- (a27);
\draw[solutebond] (a18) -- (a19);
\draw[solutebond] (a18) -- (a20);
\draw[solutebond] (a18) -- (a21);
\draw[solutebond] (a21) -- (a22);
\draw[solutebond] (a21) -- (a23);
\draw[solutebond] (a21) -- (a24);
\draw[solutebond] (a24) -- (a25);
\draw[solutebond] (a24) -- (a26);
\draw[solutebond] (a27) -- (a28);
\draw[solutebond] (a27) -- (a29);
\node[soluteO] at (a25) {};
\node[soluteO] at (a26) {};
\node[soluteC] at (a24) {};
\node[soluteH] at (a23) {};
\node[soluteC] at (a21) {};
\node[soluteH] at (a19) {};
\node[soluteH] at (a22) {};
\node[soluteC] at (a18) {};
\node[soluteH] at (a17) {};
\node[soluteH] at (a3) {};
\node[soluteH] at (a20) {};
\node[soluteC] at (a16) {};
\node[soluteO] at (a13) {};
\node[soluteH] at (a2) {};
\node[soluteN] at (a1) {};
\node[soluteN] at (a14) {};
\node[soluteO] at (a29) {};
\node[soluteC] at (a12) {};
\node[soluteH] at (a9) {};
\node[soluteH] at (a4) {};
\node[soluteH] at (a15) {};
\node[soluteC] at (a27) {};
\node[soluteC] at (a5) {};
\node[soluteC] at (a7) {};
\node[soluteO] at (a28) {};
\node[soluteH] at (a6) {};
\node[soluteS] at (a10) {};
\node[soluteH] at (a8) {};
\node[soluteH] at (a11) {};
\end{scope}
\newdimen\tokx
\newdimen\toky
\newdimen\tokleft
\newdimen\tokright
\newdimen\toksep
\newdimen\tokrow
\global\tokleft=-2.35cm
\global\tokright=9.65cm
\global\toksep=0.40mm
\global\tokrow=5.0mm
\global\tokx=\tokleft
\global\toky=1.00cm
\def\addtok#1#2#3{%
  \ifdim\dimexpr\tokx+#3\relax>\tokright
    \global\tokx=\tokleft
    \global\advance\toky by -\tokrow
  \fi
  \node[#1,minimum width=#3] at (\tokx,\toky) {#2};
  \global\advance\tokx by #3
  \global\advance\tokx by \toksep
}
\def\tokgap{\global\advance\tokx by 1.6mm}
\def\xyztok#1{%
  \ifdim\dimexpr\tokx+12.8mm\relax>\tokright
    \global\tokx=\tokleft
    \global\advance\toky by -\tokrow
  \fi
  \addtok{#1}{x}{4.0mm}%
  \addtok{#1}{y}{4.0mm}%
  \addtok{#1}{z}{4.0mm}%
}
\addtok{promptTok}{$<$env$>$}{8.0mm}
\tokgap
\addtok{promptTok}{3}{4.0mm}
\addtok{promptTok}{0}{4.0mm}
\addtok{promptTok}{0}{4.0mm}
\addtok{promptTok}{K}{4.0mm}
\addtok{promptTok}{,}{4.0mm}
\addtok{promptTok}{water}{9.0mm}
\addtok{promptTok}{,}{4.0mm}
\addtok{promptTok}{ff14sb}{10.5mm}
\addtok{promptTok}{-}{4.0mm}
\addtok{promptTok}{obc1}{8.0mm}
\addtok{promptTok}{-}{4.0mm}
\addtok{promptTok}{2}{4.0mm}
\addtok{promptTok}{nm}{4.0mm}
\addtok{promptTok}{$<$/env$>$}{9.0mm}
\tokgap
\addtok{promptTok}{$<$helm$>$}{9.0mm}
\tokgap
\addtok{promptTok}{PEPTIDE}{12.0mm}
\addtok{promptTok}{1}{4.0mm}
\addtok{promptTok}{\{}{4.0mm}
\addtok{promptTok}{[}{4.0mm}
\addtok{promptTok}{N}{4.0mm}
\addtok{promptTok}{[}{4.0mm}
\addtok{promptTok}{C}{4.0mm}
\addtok{promptTok}{@@}{4.0mm}
\addtok{promptTok}{H}{4.0mm}
\addtok{promptTok}{]}{4.0mm}
\addtok{promptTok}{(}{4.0mm}
\addtok{promptTok}{C}{4.0mm}
\addtok{promptTok}{S}{4.0mm}
\addtok{promptTok}{)}{4.0mm}
\addtok{promptTok}{C}{4.0mm}
\addtok{promptTok}{(=}{4.0mm}
\addtok{promptTok}{O}{4.0mm}
\addtok{promptTok}{)}{4.0mm}
\addtok{promptTok}{[}{4.0mm}
\addtok{promptTok}{*}{4.0mm}
\addtok{promptTok}{:}{4.0mm}
\addtok{promptTok}{2}{4.0mm}
\addtok{promptTok}{]}{4.0mm}
\addtok{promptTok}{]}{4.0mm}
\addtok{promptTok}{.}{4.0mm}
\addtok{promptTok}{E}{4.0mm}
\addtok{promptTok}{\}}{4.0mm}
\addtok{promptTok}{\$\$\$\$}{8.0mm}
\addtok{promptTok}{$<$/helm$>$}{10.0mm}
\tokgap
\addtok{coordTok}{$<$coord$>$}{10.5mm}
\tokgap
\xyztok{soluteTokN}
\xyztok{soluteTokH}
\xyztok{soluteTokH}
\xyztok{soluteTokH}
\xyztok{soluteTokC}
\xyztok{soluteTokH}
\xyztok{soluteTokC}
\xyztok{soluteTokH}
\xyztok{soluteTokH}
\xyztok{soluteTokS}
\xyztok{soluteTokH}
\xyztok{soluteTokC}
\xyztok{soluteTokO}
\xyztok{soluteTokN}
\xyztok{soluteTokH}
\xyztok{soluteTokC}
\xyztok{soluteTokH}
\xyztok{soluteTokC}
\xyztok{soluteTokH}
\xyztok{soluteTokH}
\xyztok{soluteTokC}
\xyztok{soluteTokH}
\xyztok{soluteTokH}
\xyztok{soluteTokC}
\xyztok{soluteTokO}
\xyztok{soluteTokO}
\xyztok{soluteTokC}
\xyztok{soluteTokO}
\xyztok{soluteTokO}
\addtok{coordTok}{$<$/coord$>$}{11.5mm}
\end{tikzpicture}%
}
\caption{\textbf{Prefix-Conditioned Boltzmann Generators} represent molecular environments, chemical graphs, and atomic coordinates as discrete tokens, enabling flexible sampling across molecular systems and environments through prefix conditioning. Sampling quality exhibits empirical scaling relationships.
}
\label{fig:dbg-ar-generation}
\end{figure}

%% file: generated/sota_plot.tex
\resizebox{\linewidth}{!}{%
\begin{tabular}{lcccccccccccc}
\toprule
 & \multicolumn{4}{c}{$2\mathrm{AA}$ (30 systems)} & \multicolumn{4}{c}{$4\mathrm{AA}$ (30 systems)} & \multicolumn{4}{c}{$8\mathrm{AA}$ (29 systems)\,\P} \\
\cmidrule(lr){2-5} \cmidrule(lr){6-9} \cmidrule(lr){10-13}
Model & $\mathrm{rESS}\uparrow$ & $\mathcal{E}\text{-}\mathcal{W}_2\downarrow$ & $\mathbb{T}\text{-}\mathcal{W}_2\downarrow$ & $\mathrm{TICA}\text{-}\mathcal{W}_2\downarrow$ & $\mathrm{rESS}\uparrow$ & $\mathcal{E}\text{-}\mathcal{W}_2\downarrow$ & $\mathbb{T}\text{-}\mathcal{W}_2\downarrow$ & $\mathrm{TICA}\text{-}\mathcal{W}_2\downarrow$ & $\mathrm{rESS}\uparrow$ & $\mathcal{E}\text{-}\mathcal{W}_2\downarrow$ & $\mathbb{T}\text{-}\mathcal{W}_2\downarrow$ & $\mathrm{TICA}\text{-}\mathcal{W}_2\downarrow$ \\
\midrule

$\text{TarFlow}$ & $0.134$ & $0.452$ & $0.193$ & $\text{---}$ & $0.045$ & $1.260$ & $0.924$ & $0.492$ & $0.008$ & $11.272$ & $2.644$ & $1.059$ \\

$\text{Prose}$ & $0.191$ & $0.371$ & $0.210$ & $\text{---}$ & $0.071$ & $0.932$ & $0.752$ & $\mathbf{0.367}$ & $0.011$ & $9.817$ & $2.451$ & $1.026$ \\

$\text{ROBIN SMC}$ & $\text{---}$ & $\text{---}$ & $\text{---}$ & $\text{---}$ & $\text{---}$ & $1.079$ & $0.874$ & $0.463$ & $\text{---}$ & $4.238$ & $2.300$ & $0.996$ \\

$\text{ROBIN}\ (T=0.9)$ & $\text{---}$ & $\text{---}$ & $\text{---}$ & $\text{---}$ & $0.031$ & $1.168$ & $0.886$ & $0.471$ & $0.013$ & $4.295$ & $2.314$ & $0.957$ \\
\midrule
$\text{ROBIN}\ (T=1.0)$ & $0.036\,{\scriptstyle \pm1}$ & $0.689\,{\scriptstyle \pm23}$ & $0.242\,{\scriptstyle \pm14}$ & $0.062\,{\scriptstyle \pm5}$ & $0.031\,{\scriptstyle {<}1}$ & $1.210\,{\scriptstyle \pm76}$ & $0.813\,{\scriptstyle \pm11}$ & $0.454\,{\scriptstyle \pm25}$ & $0.015\,{\scriptstyle \pm1}$ & $6.035\,{\scriptstyle \pm211}$ & $2.244\,{\scriptstyle \pm10}$ & $\underline{0.956}\,{\scriptstyle \pm15}$ \\
\midrule
$\text{PBG 30M}\ (D/N = 1760)$ & $0.451\,{\scriptstyle \pm2}$ & $0.288\,{\scriptstyle \pm22}$ & $0.117\,{\scriptstyle \pm6}$ & $0.043\,{\scriptstyle \pm3}$ & $0.181\,{\scriptstyle \pm1}$ & $0.725\,{\scriptstyle \pm67}$ & $0.634\,{\scriptstyle \pm6}$ & $0.414\,{\scriptstyle \pm21}$ & $0.021\,{\scriptstyle \pm1}$ & $6.859\,{\scriptstyle \pm157}$ & $2.162\,{\scriptstyle \pm13}$ & $1.029\,{\scriptstyle \pm25}$ \\

$\text{PBG 49M}\ (D/N = 416)$ & $0.420\,{\scriptstyle \pm2}$ & $0.293\,{\scriptstyle {<}1}$ & $0.123\,{\scriptstyle \pm10}$ & $\mathbf{0.041}\,{\scriptstyle \pm1}$ & $0.161\,{\scriptstyle \pm1}$ & $0.749\,{\scriptstyle \pm32}$ & $0.639\,{\scriptstyle \pm5}$ & $0.417\,{\scriptstyle \pm27}$ & $0.019\,{\scriptstyle {<}1}$ & $6.811\,{\scriptstyle \pm235}$ & $2.178\,{\scriptstyle \pm10}$ & $1.035\,{\scriptstyle \pm15}$ \\

$\text{PBG 85M}\ (D/N = 768)$ & $\underline{0.573}\,{\scriptstyle \pm2}$ & $\underline{0.264}\,{\scriptstyle \pm10}$ & $0.111\,{\scriptstyle \pm3}$ & $0.043\,{\scriptstyle \pm1}$ & $\underline{0.294}\,{\scriptstyle \pm1}$ & $\underline{0.496}\,{\scriptstyle \pm19}$ & $0.571\,{\scriptstyle \pm7}$ & $\underline{0.374}\,{\scriptstyle \pm8}$ & $\underline{0.041}\,{\scriptstyle \pm1}$ & $4.114\,{\scriptstyle \pm149}$ & $1.983\,{\scriptstyle \pm7}$ & $1.001\,{\scriptstyle \pm17}$ \\

$\text{PBG 160M}\ (D/N = 352)$ & $0.448\,{\scriptstyle \pm2}$ & $0.289\,{\scriptstyle \pm12}$ & $0.114\,{\scriptstyle \pm1}$ & $0.043\,{\scriptstyle \pm4}$ & $0.216\,{\scriptstyle \pm9}$ & $0.538\,{\scriptstyle \pm31}$ & $0.587\,{\scriptstyle \pm4}$ & $0.403\,{\scriptstyle \pm23}$ & $0.036\,{\scriptstyle \pm1}$ & $\underline{3.766}\,{\scriptstyle \pm140}$ & $1.981\,{\scriptstyle \pm15}$ & $1.013\,{\scriptstyle \pm18}$ \\

$\text{PBG 320M}\ (D/N = 86)$ & $0.501\,{\scriptstyle \pm1}$ & $0.274\,{\scriptstyle \pm9}$ & $\underline{0.107}\,{\scriptstyle \pm10}$ & $0.042\,{\scriptstyle {<}1}$ & $0.260\,{\scriptstyle \pm2}$ & $0.500\,{\scriptstyle \pm48}$ & $\underline{0.569}\,{\scriptstyle \pm6}$ & $0.374\,{\scriptstyle \pm19}$ & $0.040\,{\scriptstyle {<}1}$ & $3.809\,{\scriptstyle \pm323}$ & $\underline{1.971}\,{\scriptstyle \pm12}$ & $0.980\,{\scriptstyle \pm6}$ \\

$\text{PBG 640M}\ (D/N = 48)$ & $\mathbf{0.657}\,{\scriptstyle \pm2}$ & $\mathbf{0.259}\,{\scriptstyle \pm4}$ & $\mathbf{0.107}\,{\scriptstyle \pm8}$ & $0.043\,{\scriptstyle \pm2}$ & $\mathbf{0.399}\,{\scriptstyle \pm1}$ & $\mathbf{0.438}\,{\scriptstyle \pm29}$ & $\mathbf{0.545}\,{\scriptstyle \pm3}$ & $0.413\,{\scriptstyle \pm7}$ & $\mathbf{0.078}\,{\scriptstyle \pm2}$ & $\mathbf{2.631}\,{\scriptstyle \pm102}$ & $\mathbf{1.823}\,{\scriptstyle \pm6}$ & $\mathbf{0.955}\,{\scriptstyle \pm11}$ \\
\midrule
$\text{PBG-Joint 160M}\ (D/N = 198)$ & $0.439\,{\scriptstyle \pm1}$ & $0.278\,{\scriptstyle \pm14}$ & $0.114\,{\scriptstyle \pm3}$ & $\underline{0.041}\,{\scriptstyle \pm2}$ & $0.198\,{\scriptstyle \pm2}$ & $0.580\,{\scriptstyle \pm5}$ & $0.589\,{\scriptstyle \pm5}$ & $0.378\,{\scriptstyle \pm18}$ & $0.032\,{\scriptstyle \pm1}$ & $4.096\,{\scriptstyle \pm488}$ & $1.979\,{\scriptstyle \pm12}$ & $0.957\,{\scriptstyle \pm12}$ \\
\bottomrule
\end{tabular}%
}

%% file: generated/dipper_solvated_solvator_comparison.tex
\resizebox{\linewidth}{!}{%
\begin{tabular}{@{}lrrrrrrrrr@{}}
\toprule
 & \multicolumn{3}{c}{$2\mathrm{AA}+\ 80\,\mathrm{H_2O}$ (19\ net\ neutral\ systems)} & \multicolumn{3}{c}{$4\mathrm{AA}+\ 64\,\mathrm{H_2O}$ (13\ net\ neutral\ systems)} & \multicolumn{3}{c}{$8\mathrm{AA}+\ 32\,\mathrm{H_2O}$ (14\ net\ neutral\ systems)} \\
\cmidrule(lr){2-4} \cmidrule(lr){5-7} \cmidrule(lr){8-10}
Method & Avg. min $\downarrow$ & Avg. median $\downarrow$ & Avg. samples/s $\uparrow$ & Avg. min $\downarrow$ & Avg. median $\downarrow$ & Avg. samples/s $\uparrow$ & Avg. min $\downarrow$ & Avg. median $\downarrow$ & Avg. samples/s $\uparrow$ \\
 & {\scriptsize kcal/mol} & {\scriptsize kcal/mol} & & {\scriptsize kcal/mol} & {\scriptsize kcal/mol} & & {\scriptsize kcal/mol} & {\scriptsize kcal/mol} & \\
\midrule
Reference MD & $-659.8$ & $-592.0$ & \text{---} & $-489.4$ & $-417.6$ & \text{---} & $-241.7$ & $-154.0$ & \text{---} \\
\midrule
AMBER tleap & $\mathbf{-412.4}\,{\scriptstyle \pm19}$ & $\mathbf{-322.4}\,{\scriptstyle \pm1}$ & $\underline{10.79}$ & $-199.7\,{\scriptstyle \pm37}$ & $-106.6\,{\scriptstyle \pm4}$ & $\underline{9.228}$ & $-53.36\,{\scriptstyle \pm524}$ & $81.78\,{\scriptstyle \pm27}$ & $\underline{7.496}$ \\
ORCA SOLVATOR & $-369.8\,{\scriptstyle \pm13}$ & $\underline{-217.4}\,{\scriptstyle \pm3}$ & $0.4477$ & $\underline{-287.6}\,{\scriptstyle \pm9}$ & $\mathbf{-172.3}\,{\scriptstyle \pm1}$ & $0.4913$ & $\mathbf{-160.8}\,{\scriptstyle \pm11}$ & $\underline{-40.32}\,{\scriptstyle \pm64}$ & $0.6494$ \\
PBG 85M & $\underline{-406.5}\,{\scriptstyle \pm23}$ & $-188.9\,{\scriptstyle \pm52}$ & $\mathbf{14.14}$ & $\mathbf{-294.7}\,{\scriptstyle \pm48}$ & $\underline{-126.8}\,{\scriptstyle \pm13}$ & $\mathbf{15.03}$ & $\underline{-158.3}\,{\scriptstyle \pm32}$ & $\mathbf{-50.75}\,{\scriptstyle \pm57}$ & $\mathbf{16.59}$ \\
\bottomrule
\end{tabular}%
}

%% file: generated/sota_modern_plot.tex
\resizebox{\linewidth}{!}{%
\begin{tabular}{lccccccccc}
\toprule
 & \multicolumn{3}{c}{$2\mathrm{AA}$ (30 systems)} & \multicolumn{3}{c}{$4\mathrm{AA}$ (30 systems)} & \multicolumn{3}{c}{$8\mathrm{AA}$ (29 systems)\,\P} \\
\cmidrule(lr){2-4} \cmidrule(lr){5-7} \cmidrule(lr){8-10}
Model & $\mathrm{rESS}^{\ast}\uparrow$ & $\mathcal{E}\text{-}\mathcal{W}_2^{\ast}\downarrow$ & $\mathbb{T}\mathrm{g}\text{-}\mathcal{SW}_2^{\ast}\downarrow$ & $\mathrm{rESS}^{\ast}\uparrow$ & $\mathcal{E}\text{-}\mathcal{W}_2^{\ast}\downarrow$ & $\mathbb{T}\mathrm{g}\text{-}\mathcal{SW}_2^{\ast}\downarrow$ & $\mathrm{rESS}^{\ast}\uparrow$ & $\mathcal{E}\text{-}\mathcal{W}_2^{\ast}\downarrow$ & $\mathbb{T}\mathrm{g}\text{-}\mathcal{SW}_2^{\ast}\downarrow$ \\
\midrule

$\text{ROBIN}\ (T=1.0)$ & $0.002\,{\scriptstyle \pm1}$ & $2.294\,{\scriptstyle \pm513}$ & $0.441\,{\scriptstyle \pm72}$ & $0.001\,{\scriptstyle \pm1}$ & $3.774\,{\scriptstyle \pm507}$ & $0.470\,{\scriptstyle \pm47}$ & $0.001\,{\scriptstyle {<}1}$ & $8.506\,{\scriptstyle \pm218}$ & $0.639\,{\scriptstyle \pm50}$ \\
\midrule
$\text{PBG 30M}\ (D/N = 1760)$ & $0.322\,{\scriptstyle \pm26}$ & $0.261\,{\scriptstyle \pm14}$ & $0.090\,{\scriptstyle \pm5}$ & $0.058\,{\scriptstyle \pm10}$ & $0.827\,{\scriptstyle \pm107}$ & $0.153\,{\scriptstyle \pm7}$ & $0.002\,{\scriptstyle \pm1}$ & $5.832\,{\scriptstyle \pm919}$ & $0.495\,{\scriptstyle \pm70}$ \\

$\text{PBG 49M}\ (D/N = 416)$ & $0.305\,{\scriptstyle \pm18}$ & $0.257\,{\scriptstyle \pm11}$ & $0.088\,{\scriptstyle \pm5}$ & $0.047\,{\scriptstyle \pm14}$ & $0.730\,{\scriptstyle \pm48}$ & $0.147\,{\scriptstyle \pm15}$ & $0.002\,{\scriptstyle {<}1}$ & $6.300\,{\scriptstyle \pm717}$ & $0.506\,{\scriptstyle \pm23}$ \\

$\text{PBG 85M}\ (D/N = 768)$ & $\underline{0.494}\,{\scriptstyle \pm22}$ & $\underline{0.227}\,{\scriptstyle \pm13}$ & $\underline{0.081}\,{\scriptstyle \pm2}$ & $\underline{0.150}\,{\scriptstyle \pm48}$ & $\underline{0.513}\,{\scriptstyle \pm36}$ & $\underline{0.115}\,{\scriptstyle \pm4}$ & $\underline{0.004}\,{\scriptstyle {<}1}$ & $\underline{3.843}\,{\scriptstyle \pm382}$ & $0.373\,{\scriptstyle \pm34}$ \\

$\text{PBG 160M}\ (D/N = 352)$ & $0.336\,{\scriptstyle \pm65}$ & $0.311\,{\scriptstyle \pm45}$ & $0.091\,{\scriptstyle \pm7}$ & $0.085\,{\scriptstyle \pm46}$ & $0.646\,{\scriptstyle \pm96}$ & $0.130\,{\scriptstyle \pm18}$ & $0.003\,{\scriptstyle \pm1}$ & $4.665\,{\scriptstyle \pm507}$ & $0.389\,{\scriptstyle \pm34}$ \\

$\text{PBG 320M}\ (D/N = 86)$ & $0.416\,{\scriptstyle \pm13}$ & $0.249\,{\scriptstyle \pm10}$ & $0.084\,{\scriptstyle \pm4}$ & $0.111\,{\scriptstyle \pm40}$ & $0.615\,{\scriptstyle \pm59}$ & $0.117\,{\scriptstyle \pm5}$ & $0.004\,{\scriptstyle \pm1}$ & $3.981\,{\scriptstyle \pm563}$ & $\underline{0.360}\,{\scriptstyle \pm30}$ \\

$\text{PBG 640M}\ (D/N = 48)$ & $\mathbf{0.605}\,{\scriptstyle \pm8}$ & $\mathbf{0.220}\,{\scriptstyle \pm11}$ & $\mathbf{0.077}\,{\scriptstyle \pm2}$ & $\mathbf{0.284}\,{\scriptstyle \pm34}$ & $\mathbf{0.460}\,{\scriptstyle \pm23}$ & $\mathbf{0.099}\,{\scriptstyle \pm4}$ & $\mathbf{0.008}\,{\scriptstyle \pm2}$ & $\mathbf{2.802}\,{\scriptstyle \pm415}$ & $\mathbf{0.270}\,{\scriptstyle \pm22}$ \\
\midrule
$\text{PBG-Joint 160M}\ (D/N = 198)$ & $0.272\,{\scriptstyle \pm28}$ & $0.326\,{\scriptstyle \pm58}$ & $0.103\,{\scriptstyle \pm10}$ & $0.046\,{\scriptstyle \pm12}$ & $0.866\,{\scriptstyle \pm187}$ & $0.146\,{\scriptstyle \pm2}$ & $0.002\,{\scriptstyle {<}1}$ & $5.196\,{\scriptstyle \pm446}$ & $0.449\,{\scriptstyle \pm34}$ \\
\bottomrule
\end{tabular}%
}

%% file: generated/joint_scaling_plot.tex
\resizebox{\linewidth}{!}{%
\begin{tabular}{llc}
\toprule
Group & Parameter & Posterior median $\pm$ half 95\% CI \\
\midrule

$\text{core }G$ & $G_0$ & $0.0740 \pm 0.0117$ \\

 & $r$ & $40.5 \pm 5.0$ \\

 & $\alpha$ & $0.227 \pm 0.0269$ \\
\midrule
$\text{train NLL}$ & $E$ & $3.526 \pm 0.0117$ \\

 & $\sigma_L$ & $0.00331 \pm 7.63\times 10^{-4}$ \\

 & $\nu_L$ & $2.699 \pm 0.568$ \\
\midrule
$\log(\mathcal{E}\text{-}\mathcal{W}_2^{\ast})$ & $a_E$ & $8.951 \pm 0.698$ \\

 & $p_E$ & $2.173 \pm 0.0207$ \\

 & $k_E$ & $3.408 \pm 0.410$ \\

 & $\sigma_E$ & $0.513 \pm 0.0111$ \\

 & $\nu_E$ & $2.482 \pm 0.111$ \\
\midrule
$\log(\mathbb{T}\mathrm{g}\text{-}\mathcal{SW}_2^{\ast})$ & $a_S$ & $4.375 \pm 0.480$ \\

 & $p_S$ & $1.018 \pm 0.0157$ \\

 & $k_S$ & $2.317 \pm 0.279$ \\

 & $\sigma_S$ & $0.401 \pm 0.00932$ \\

 & $\nu_S$ & $4.019 \pm 0.316$ \\
\midrule
$-\ln(\mathrm{rESS}^{\ast})$ & $a_R$ & $10.80 \pm 0.880$ \\

 & $p_R$ & $1.786 \pm 0.0227$ \\

 & $k_R$ & $4.391 \pm 0.518$ \\

 & $\sigma_R$ & $0.760 \pm 0.0193$ \\

 & $\nu_R$ & $2.471 \pm 0.132$ \\
\bottomrule
\end{tabular}
}

%% file: generated/joint_scaling_r2.tex
\resizebox{\linewidth}{!}{%
\begin{tabular}{lc}
\toprule
Response & $R^2$ \\
\midrule
$\text{train NLL}$ & $0.9747$ \\
$\log(\mathcal{E}\text{-}\mathcal{W}_2^{\ast})$ & $0.6524$ \\
$\log(\mathbb{T}\mathrm{g}\text{-}\mathcal{SW}_2^{\ast})$ & $0.5681$ \\
$-\ln(\mathrm{rESS}^{\ast})$ & $0.6728$ \\
\bottomrule
\end{tabular}
}

%% file: generated/dipper_scaling_ablation_table.tex
\resizebox{\linewidth}{!}{%
\begin{tabular}{@{}llrrrrrlrrr@{}}
\toprule
& & & & & & & \multicolumn{4}{c}{Held-out checkpoints: RMSE} \\
\cmidrule(l){8-11}
Row & Model & $\alpha$ & $\beta$ & $D^\ast/N$ & $C^\ast_{\mathcal{E}\text{-}\mathcal{W}_2}$ exponent & $p_E$ & NLL & $\mathcal{E}\text{-}\mathcal{W}_2^{\ast}$ & $\mathbb{T}\mathrm{g}\text{-}\mathcal{SW}_2^{\ast}$ & $\mathrm{rESS}^{\ast}$ \\
\midrule
A0 & NLL only, $\alpha\neq\beta$, priors on $A,B$, $\gamma=0$ & 1.13\,{\scriptsize[0.942, 1.35]} & 0.815\,{\scriptsize[0.649, 1.00]} & 63.4\,{\scriptsize[41.5, 109]} & --- & --- & \underline{0.00370} & --- & --- & --- \\
A1 & + metric adapters & 1.22\,{\scriptsize[1.14, 1.30]} & 0.148\,{\scriptsize[0.125, 0.177]} & $4.8\times10^{4}$ & 4.96\,{\scriptsize[4.84, 5.08]} & 2.09\,{\scriptsize[2.07, 2.11]} & 0.00848 & 1.70 & \textbf{0.648} & \underline{1.95} \\
A2 & + tied exponents $\alpha=\beta$ & 0.293\,{\scriptsize[0.260, 0.330]} & $=\alpha$ & 34.9\,{\scriptsize[30.2, 39.6]} & 4.32\,{\scriptsize[4.22, 4.42]} & 2.09\,{\scriptsize[2.07, 2.11]} & 0.01915 & 1.89 & 0.740 & 2.45 \\
A3 & + prior $r\sim\mathcal N^{+}(20,10)$ on $D^\ast/N$ & 0.268\,{\scriptsize[0.240, 0.301]} & $=\alpha$ & 29.6\,{\scriptsize[26.0, 33.2]} & 4.30\,{\scriptsize[4.20, 4.41]} & 2.09\,{\scriptsize[2.07, 2.11]} & 0.02075 & 1.89 & 0.741 & 2.46 \\
A4 & + compute tilting $\gamma=0.5$ (default) & 0.227\,{\scriptsize[0.201, 0.255]} & $=\alpha$ & 40.5\,{\scriptsize[35.5, 45.5]} & 5.61\,{\scriptsize[5.46, 5.78]} & 2.17\,{\scriptsize[2.15, 2.19]} & 0.00614 & 1.80 & 0.725 & 2.17 \\
A5 & + test-system effects & 0.217\,{\scriptsize[0.193, 0.242]} & $=\alpha$ & 43.9\,{\scriptsize[38.9, 48.9]} & 6.13\,{\scriptsize[5.23, 7.03]} & 2.37\,{\scriptsize[2.03, 2.72]} & 0.00608 & \textbf{1.13} & \underline{0.672} & \textbf{1.93} \\
A6 & + training-lineage effects & 0.371\,{\scriptsize[0.342, 0.402]} & $=\alpha$ & 26.0\,{\scriptsize[20.9, 31.4]} & 4.78\,{\scriptsize[4.06, 5.52]} & 2.38\,{\scriptsize[2.03, 2.72]} & 0.00704 & \underline{1.22} & 0.717 & 2.04 \\
\midrule
D1 & Default with $\alpha\neq\beta$ & 0.106\,{\scriptsize[0.0933, 0.120]} & 0.454\,{\scriptsize[0.422, 0.486]} & 44.8\,{\scriptsize[41.6, 48.1]} & 5.39\,{\scriptsize[5.22, 5.58]} & 2.17\,{\scriptsize[2.15, 2.19]} & 0.00771 & 1.79 & 0.718 & 2.16 \\
D2 & Default with priors on $A,B$ instead of $r$ & 0.337\,{\scriptsize[0.296, 0.380]} & $=\alpha$ & 101\,{\scriptsize[86.6, 116]} & 5.90\,{\scriptsize[5.72, 6.10]} & 2.17\,{\scriptsize[2.15, 2.19]} & \textbf{0.00356} & 1.78 & 0.705 & 2.11 \\
D3 & Default with $\gamma=0$ & 0.268\,{\scriptsize[0.240, 0.301]} & $=\alpha$ & 29.6\,{\scriptsize[26.0, 33.2]} & 4.30\,{\scriptsize[4.20, 4.41]} & 2.09\,{\scriptsize[2.07, 2.11]} & 0.02075 & 1.89 & 0.741 & 2.46 \\
\midrule
F1 & Exponential atom-count form & 0.222\,{\scriptsize[0.198, 0.250]} & $=\alpha$ & 40.4\,{\scriptsize[35.5, 45.3]} & --- & --- & 0.00617 & 1.79 & 0.721 & 2.10 \\
F2 & Power law plus curvature & 0.224\,{\scriptsize[0.198, 0.254]} & $=\alpha$ & 40.7\,{\scriptsize[35.7, 45.8]} & --- & 2.08\,{\scriptsize[2.06, 2.10]} & 0.00616 & 1.79 & 0.722 & 2.12 \\
F3 & Extensive, $p_E=p_R=1$ fixed & 0.219\,{\scriptsize[0.193, 0.250]} & $=\alpha$ & 37.3\,{\scriptsize[32.3, 42.4]} & 2.46\,{\scriptsize[2.37, 2.55]} & $=1$ & 0.00630 & 1.98 & 0.726 & 2.27 \\
C1 & Censored rESS, $\mathrm{ESS}\le1.1$ & 0.229\,{\scriptsize[0.202, 0.259]} & $=\alpha$ & 40.2\,{\scriptsize[35.1, 45.1]} & 5.61\,{\scriptsize[5.45, 5.78]} & 2.17\,{\scriptsize[2.15, 2.19]} & 0.00614 & 1.80 & 0.725 & 2.18 \\
\bottomrule
\end{tabular}%
}

%% file: generated/dipper_scaling_ablation_loso_table.tex
\resizebox{\linewidth}{!}{%
\begin{tabular}{@{}lrrrrrlrrrlrrr@{}}
\toprule
& & & & & & \multicolumn{4}{c}{5-fold holdout: RMSE} & \multicolumn{4}{c}{Extrapolation checkpoints: RMSE} \\
\cmidrule(lr){7-10}\cmidrule(l){11-14}
Row & $\alpha$ & $\beta$ & $D^\ast/N$ & $C^\ast_{\mathcal{E}\text{-}\mathcal{W}_2}$ exponent & $p_E$ & NLL & $\mathcal{E}\text{-}\mathcal{W}_2^{\ast}$ & $\mathbb{T}\mathrm{g}\text{-}\mathcal{SW}_2^{\ast}$ & $\mathrm{rESS}^{\ast}$ & NLL & $\mathcal{E}\text{-}\mathcal{W}_2^{\ast}$ & $\mathbb{T}\mathrm{g}\text{-}\mathcal{SW}_2^{\ast}$ & $\mathrm{rESS}^{\ast}$ \\
\midrule
A0 & 1.20\,{\scriptsize[0.772, 1.95]} & 0.771\,{\scriptsize[0.653, 0.837]} & 83.4\,{\scriptsize[30.4, 168]} & --- & --- & 0.02640 & --- & --- & --- & \textbf{0.00506} & --- & --- & --- \\
A1 & 0.683\,{\scriptsize[0.137, 1.18]} & 0.274\,{\scriptsize[0.126, 0.434]} & $1.8\times10^{4}$ & 4.76\,{\scriptsize[4.38, 4.99]} & 2.09\,{\scriptsize[2.07, 2.12]} & 0.02109 & 1.22 & \textbf{0.516} & \underline{1.22} & 0.01073 & 1.76 & \textbf{0.678} & 2.17 \\
A2 & 0.330\,{\scriptsize[0.242, 0.408]} & $=\alpha$ & 40.8\,{\scriptsize[21.6, 70.2]} & 4.31\,{\scriptsize[3.53, 4.94]} & 2.09\,{\scriptsize[2.06, 2.11]} & 0.02187 & 1.23 & 0.522 & 1.26 & 0.01838 & 1.91 & 0.751 & 2.50 \\
A3 & 0.298\,{\scriptsize[0.228, 0.389]} & $=\alpha$ & 30.7\,{\scriptsize[21.2, 45.2]} & 4.26\,{\scriptsize[3.52, 4.77]} & 2.09\,{\scriptsize[2.06, 2.11]} & \underline{0.02102} & 1.22 & \underline{0.520} & 1.24 & 0.02036 & 1.92 & 0.755 & 2.52 \\
A4 & 0.271\,{\scriptsize[0.205, 0.429]} & $=\alpha$ & 36.8\,{\scriptsize[29.2, 46.6]} & 5.34\,{\scriptsize[4.29, 6.12]} & 2.17\,{\scriptsize[2.16, 2.18]} & 0.02436 & 1.28 & 0.578 & 1.34 & 0.00864 & 1.85 & 0.758 & 2.28 \\
A5 & 0.259\,{\scriptsize[0.193, 0.414]} & $=\alpha$ & 39.5\,{\scriptsize[30.7, 51.0]} & 5.84\,{\scriptsize[4.67, 6.69]} & 2.37\,{\scriptsize[2.36, 2.39]} & 0.02448 & \underline{0.861} & 0.588 & 1.31 & 0.00855 & \textbf{1.19} & \underline{0.707} & \textbf{2.04} \\
A6 & 0.381\,{\scriptsize[0.318, 0.469]} & $=\alpha$ & 25.3\,{\scriptsize[24.4, 26.0]} & 4.72\,{\scriptsize[4.08, 5.13]} & 2.37\,{\scriptsize[2.37, 2.38]} & \textbf{0.01955} & \textbf{0.780} & 0.564 & \textbf{1.14} & 0.00795 & \underline{1.25} & 0.733 & \underline{2.10} \\
\midrule
D1 & 0.147\,{\scriptsize[0.108, 0.227]} & 0.440\,{\scriptsize[0.293, 0.517]} & 40.3\,{\scriptsize[25.8, 53.5]} & 5.13\,{\scriptsize[4.23, 6.11]} & 2.17\,{\scriptsize[2.16, 2.18]} & 0.02397 & 1.29 & 0.585 & 1.34 & 0.01011 & 1.84 & 0.754 & 2.28 \\
D2 & 0.377\,{\scriptsize[0.267, 0.481]} & $=\alpha$ & 96.7\,{\scriptsize[35.1, 180]} & 5.64\,{\scriptsize[4.39, 6.87]} & 2.17\,{\scriptsize[2.16, 2.18]} & 0.02440 & 1.30 & 0.588 & 1.40 & \underline{0.00601} & 1.83 & 0.738 & 2.22 \\
D3 & 0.298\,{\scriptsize[0.228, 0.389]} & $=\alpha$ & 30.7\,{\scriptsize[21.2, 45.2]} & 4.26\,{\scriptsize[3.52, 4.77]} & 2.09\,{\scriptsize[2.06, 2.11]} & \underline{0.02102} & 1.22 & \underline{0.520} & 1.24 & 0.02036 & 1.92 & 0.755 & 2.52 \\
\midrule
F1 & 0.261\,{\scriptsize[0.203, 0.400]} & $=\alpha$ & 36.7\,{\scriptsize[28.3, 46.6]} & --- & --- & 0.02437 & 1.27 & 0.584 & 1.27 & 0.00868 & 1.84 & 0.752 & 2.19 \\
F2 & 0.265\,{\scriptsize[0.203, 0.413]} & $=\alpha$ & 36.9\,{\scriptsize[28.5, 47.1]} & --- & 2.08\,{\scriptsize[2.06, 2.09]} & 0.02438 & 1.27 & 0.581 & 1.29 & 0.00870 & 1.84 & 0.755 & 2.22 \\
F3 & 0.256\,{\scriptsize[0.207, 0.390]} & $=\alpha$ & 34.5\,{\scriptsize[29.4, 40.9]} & 2.37\,{\scriptsize[2.01, 2.63]} & $=1$ & 0.02430 & 1.43 & 0.577 & 1.25 & 0.00872 & 2.01 & 0.756 & 2.35 \\
C1 & 0.272\,{\scriptsize[0.206, 0.432]} & $=\alpha$ & 36.5\,{\scriptsize[29.0, 46.1]} & 5.33\,{\scriptsize[4.29, 6.12]} & 2.17\,{\scriptsize[2.16, 2.18]} & 0.02436 & 1.28 & 0.578 & 1.34 & 0.00870 & 1.85 & 0.758 & 2.29 \\
\bottomrule
\end{tabular}%
}

%% file: generated/dipper_mdqm9_scaling_ablation_table.tex
\resizebox{\linewidth}{!}{%
\begin{tabular}{@{}llrrrrlrrr@{}}
\toprule
& & & & & & \multicolumn{4}{c}{Held-out checkpoints: RMSE} \\
\cmidrule(l){7-10}
Row & Model & $\alpha$ & $\beta$ & $D^\ast/N$ & $p_E$ & NLL & $\mathcal{E}\text{-}\mathcal{W}_2^{\ast}$ & $\mathbb{T}\mathrm{g}\text{-}\mathcal{SW}_2^{\ast}$ & $\mathrm{rESS}^{\ast}$ \\
\midrule
A0 & NLL only, $\alpha\neq\beta$, priors on $A,B$, $\gamma=0$ & 0.709\,{\scriptsize[0.362, 1.16]} & 0.551\,{\scriptsize[0.180, 1.35]} & 57.2\,{\scriptsize[16.1, 1297]} & --- & 0.01031 & --- & --- & --- \\
A1 & + metric adapters & 0.889\,{\scriptsize[0.619, 1.17]} & 0.172\,{\scriptsize[0.0973, 0.322]} & 344\,{\scriptsize[35.9, 9890]} & $-0.475$\,{\scriptsize[$-0.745$, $-0.204$]} & 0.00927 & \underline{1.76} & 0.451 & 1.99 \\
A2 & + tied exponents $\alpha=\beta$ & 0.457\,{\scriptsize[0.276, 0.724]} & $=\alpha$ & 22.1\,{\scriptsize[7.57, 40.3]} & $-0.467$\,{\scriptsize[$-0.739$, $-0.191$]} & 0.01353 & \underline{1.76} & 0.450 & 2.01 \\
A3 & + prior $r\sim\mathcal N^{+}(20,10)$ on $D^\ast/N$ & 0.422\,{\scriptsize[0.309, 0.573]} & $=\alpha$ & 20.1\,{\scriptsize[14.2, 26.0]} & $-0.468$\,{\scriptsize[$-0.736$, $-0.206$]} & 0.01426 & \underline{1.76} & 0.450 & 2.00 \\
A4 & + compute tilting $\gamma=0.5$ (default) & 0.403\,{\scriptsize[0.279, 0.573]} & $=\alpha$ & 21.0\,{\scriptsize[15.2, 27.1]} & $-0.555$\,{\scriptsize[$-0.822$, $-0.292$]} & 0.00981 & \underline{1.76} & \underline{0.449} & 1.99 \\
A5 & + test-system effects & 0.491\,{\scriptsize[0.373, 0.641]} & $=\alpha$ & 21.9\,{\scriptsize[16.1, 27.7]} & 1.73\,{\scriptsize[0.816, 2.64]} & 0.00907 & \textbf{1.58} & \textbf{0.361} & \textbf{1.81} \\
A6 & + training-lineage effects & --- & --- & --- & --- & --- & --- & --- & --- \\
\midrule
D1 & Default with $\alpha\neq\beta$ & 0.638\,{\scriptsize[0.459, 0.832]} & 0.138\,{\scriptsize[0.0781, 0.280]} & 21.0\,{\scriptsize[14.9, 27.0]} & $-0.558$\,{\scriptsize[$-0.821$, $-0.299$]} & \textbf{0.00802} & \underline{1.76} & \underline{0.449} & \underline{1.98} \\
D2 & Default with priors on $A,B$ instead of $r$ & 0.591\,{\scriptsize[0.315, 0.949]} & $=\alpha$ & 38.1\,{\scriptsize[13.4, 61.2]} & $-0.559$\,{\scriptsize[$-0.822$, $-0.293$]} & \underline{0.00811} & \underline{1.76} & 0.450 & 1.99 \\
D3 & Default with $\gamma=0$ & 0.422\,{\scriptsize[0.309, 0.573]} & $=\alpha$ & 20.1\,{\scriptsize[14.2, 26.0]} & $-0.468$\,{\scriptsize[$-0.736$, $-0.206$]} & 0.01426 & \underline{1.76} & 0.450 & 2.00 \\
\bottomrule
\end{tabular}%
}

%% file: generated/dipper_mdqm9_scaling_ablation_loso_table.tex
\resizebox{\linewidth}{!}{%
\begin{tabular}{@{}lrrrrlrrrlrrr@{}}
\toprule
& & & & & \multicolumn{4}{c}{5-fold holdout: RMSE} & \multicolumn{4}{c}{Extrapolation checkpoints: RMSE} \\
\cmidrule(lr){6-9}\cmidrule(l){10-13}
Row & $\alpha$ & $\beta$ & $D^\ast/N$ & $p_E$ & NLL & $\mathcal{E}\text{-}\mathcal{W}_2^{\ast}$ & $\mathbb{T}\mathrm{g}\text{-}\mathcal{SW}_2^{\ast}$ & $\mathrm{rESS}^{\ast}$ & NLL & $\mathcal{E}\text{-}\mathcal{W}_2^{\ast}$ & $\mathbb{T}\mathrm{g}\text{-}\mathcal{SW}_2^{\ast}$ & $\mathrm{rESS}^{\ast}$ \\
\midrule
A0 & 0.744\,{\scriptsize[0.569, 0.921]} & 0.600\,{\scriptsize[0.485, 0.803]} & 56.3\,{\scriptsize[47.9, 73.5]} & --- & \textbf{0.01450} & --- & --- & --- & 0.01084 & --- & --- & --- \\
A1 & 0.938\,{\scriptsize[0.446, 1.59]} & 0.200\,{\scriptsize[0.147, 0.278]} & $1.0\times10^{4}$ & $-0.427$\,{\scriptsize[$-0.506$, $-0.363$]} & 0.02103 & \underline{2.55} & 0.451 & 1.79 & 0.01108 & \underline{1.76} & 0.452 & 2.01 \\
A2 & 0.478\,{\scriptsize[0.336, 0.662]} & $=\alpha$ & 23.6\,{\scriptsize[12.7, 40.2]} & $-0.424$\,{\scriptsize[$-0.498$, $-0.366$]} & 0.01526 & \underline{2.55} & 0.446 & 1.77 & 0.01417 & \underline{1.76} & 0.451 & 2.03 \\
A3 & 0.417\,{\scriptsize[0.371, 0.461]} & $=\alpha$ & 20.0\,{\scriptsize[18.8, 21.2]} & $-0.425$\,{\scriptsize[$-0.499$, $-0.366$]} & \underline{0.01498} & \underline{2.55} & \underline{0.445} & 1.77 & 0.01503 & \underline{1.76} & 0.451 & 2.02 \\
A4 & 0.391\,{\scriptsize[0.341, 0.456]} & $=\alpha$ & 20.5\,{\scriptsize[19.7, 21.4]} & $-0.511$\,{\scriptsize[$-0.598$, $-0.450$]} & 0.01623 & \underline{2.55} & 0.446 & 1.77 & 0.01052 & \underline{1.76} & \underline{0.450} & 2.00 \\
A5 & 0.468\,{\scriptsize[0.381, 0.518]} & $=\alpha$ & 21.0\,{\scriptsize[18.7, 23.2]} & 1.73\,{\scriptsize[1.71, 1.75]} & 0.01607 & \textbf{1.19} & \textbf{0.364} & \textbf{1.49} & 0.00986 & \textbf{1.60} & \textbf{0.359} & \textbf{1.85} \\
A6 & --- & --- & --- & --- & --- & --- & --- & --- & --- & --- & --- & --- \\
\midrule
D1 & 0.591\,{\scriptsize[0.452, 0.689]} & 0.203\,{\scriptsize[0.130, 0.310]} & 20.7\,{\scriptsize[20.1, 21.1]} & $-0.511$\,{\scriptsize[$-0.600$, $-0.450$]} & 0.01578 & \underline{2.55} & \underline{0.445} & \underline{1.76} & \textbf{0.00924} & \underline{1.76} & \underline{0.450} & \underline{1.99} \\
D2 & 0.554\,{\scriptsize[0.393, 0.846]} & $=\alpha$ & 34.9\,{\scriptsize[17.9, 60.3]} & $-0.509$\,{\scriptsize[$-0.596$, $-0.452$]} & 0.01579 & \underline{2.55} & 0.446 & 1.78 & \underline{0.00929} & \underline{1.76} & 0.451 & 2.01 \\
D3 & 0.417\,{\scriptsize[0.371, 0.461]} & $=\alpha$ & 20.0\,{\scriptsize[18.8, 21.2]} & $-0.425$\,{\scriptsize[$-0.499$, $-0.366$]} & \underline{0.01498} & \underline{2.55} & \underline{0.445} & 1.77 & 0.01503 & \underline{1.76} & 0.451 & 2.02 \\
\bottomrule
\end{tabular}%
}

%% file: generated/sota_training_seed_plot.tex
\resizebox{\linewidth}{!}{%
\begin{tabular}{lcccccccccccc}
\toprule
 & \multicolumn{4}{c}{$2\mathrm{AA}$ (30 systems)} & \multicolumn{4}{c}{$4\mathrm{AA}$ (30 systems)} & \multicolumn{4}{c}{$8\mathrm{AA}$ (29 systems)\,\P} \\
\cmidrule(lr){2-5} \cmidrule(lr){6-9} \cmidrule(lr){10-13}
Model & $\mathrm{rESS}\uparrow$ & $\mathcal{E}\text{-}\mathcal{W}_2\downarrow$ & $\mathbb{T}\text{-}\mathcal{W}_2\downarrow$ & $\mathrm{TICA}\text{-}\mathcal{W}_2\downarrow$ & $\mathrm{rESS}\uparrow$ & $\mathcal{E}\text{-}\mathcal{W}_2\downarrow$ & $\mathbb{T}\text{-}\mathcal{W}_2\downarrow$ & $\mathrm{TICA}\text{-}\mathcal{W}_2\downarrow$ & $\mathrm{rESS}\uparrow$ & $\mathcal{E}\text{-}\mathcal{W}_2\downarrow$ & $\mathbb{T}\text{-}\mathcal{W}_2\downarrow$ & $\mathrm{TICA}\text{-}\mathcal{W}_2\downarrow$ \\
\midrule

$\text{PBG 30M, seed 42}$ & $0.057\,{\scriptstyle \pm2}$ & $0.911\,{\scriptstyle \pm20}$ & $0.284\,{\scriptstyle \pm11}$ & $0.070\,{\scriptstyle \pm3}$ & $0.015\,{\scriptstyle {<}1}$ & $3.747\,{\scriptstyle \pm39}$ & $1.076\,{\scriptstyle \pm21}$ & $0.564\,{\scriptstyle \pm12}$ & $0.003\,{\scriptstyle {<}1}$ & $26.045\,{\scriptstyle \pm631}$ & $3.021\,{\scriptstyle \pm22}$ & $1.085\,{\scriptstyle \pm38}$ \\

$\text{PBG 30M, seed 43}$ & $0.065\,{\scriptstyle \pm1}$ & $0.823\,{\scriptstyle \pm26}$ & $0.292\,{\scriptstyle \pm11}$ & $0.070\,{\scriptstyle \pm7}$ & $0.018\,{\scriptstyle {<}1}$ & $3.424\,{\scriptstyle \pm65}$ & $1.053\,{\scriptstyle \pm10}$ & $0.547\,{\scriptstyle \pm14}$ & $0.003\,{\scriptstyle {<}1}$ & $24.693\,{\scriptstyle \pm362}$ & $2.991\,{\scriptstyle \pm37}$ & $1.079\,{\scriptstyle \pm38}$ \\

$\text{PBG 30M, seed 44}$ & $0.053\,{\scriptstyle {<}1}$ & $0.930\,{\scriptstyle \pm31}$ & $0.301\,{\scriptstyle \pm10}$ & $0.074\,{\scriptstyle \pm6}$ & $0.015\,{\scriptstyle {<}1}$ & $4.164\,{\scriptstyle \pm118}$ & $1.091\,{\scriptstyle \pm9}$ & $0.558\,{\scriptstyle \pm20}$ & $0.003\,{\scriptstyle {<}1}$ & $26.486\,{\scriptstyle \pm1212}$ & $2.992\,{\scriptstyle \pm73}$ & $1.108\,{\scriptstyle \pm32}$ \\
\midrule
$\text{PBG 49M, seed 42}$ & $0.137\,{\scriptstyle \pm2}$ & $0.572\,{\scriptstyle \pm45}$ & $0.209\,{\scriptstyle \pm8}$ & $0.057\,{\scriptstyle \pm4}$ & $0.038\,{\scriptstyle {<}1}$ & $2.070\,{\scriptstyle \pm89}$ & $0.898\,{\scriptstyle \pm7}$ & $0.483\,{\scriptstyle \pm10}$ & $0.005\,{\scriptstyle {<}1}$ & $16.429\,{\scriptstyle \pm524}$ & $2.684\,{\scriptstyle \pm26}$ & $1.053\,{\scriptstyle \pm14}$ \\

$\text{PBG 49M, seed 43}$ & $0.124\,{\scriptstyle \pm1}$ & $0.565\,{\scriptstyle \pm24}$ & $0.220\,{\scriptstyle \pm3}$ & $0.062\,{\scriptstyle \pm1}$ & $0.033\,{\scriptstyle \pm1}$ & $2.239\,{\scriptstyle \pm143}$ & $0.905\,{\scriptstyle \pm13}$ & $0.489\,{\scriptstyle \pm9}$ & $0.005\,{\scriptstyle {<}1}$ & $17.310\,{\scriptstyle \pm228}$ & $2.710\,{\scriptstyle \pm32}$ & $1.026\,{\scriptstyle \pm20}$ \\

$\text{PBG 49M, seed 44}$ & $0.144\,{\scriptstyle \pm1}$ & $0.530\,{\scriptstyle \pm22}$ & $0.217\,{\scriptstyle \pm9}$ & $0.056\,{\scriptstyle \pm7}$ & $0.038\,{\scriptstyle \pm1}$ & $2.003\,{\scriptstyle \pm61}$ & $0.867\,{\scriptstyle \pm10}$ & $0.476\,{\scriptstyle \pm24}$ & $0.005\,{\scriptstyle {<}1}$ & $16.313\,{\scriptstyle \pm485}$ & $2.670\,{\scriptstyle \pm23}$ & $1.048\,{\scriptstyle \pm13}$ \\
\midrule
$\text{PBG 85M, seed 42}$ & $0.288\,{\scriptstyle \pm3}$ & $0.351\,{\scriptstyle \pm19}$ & $0.149\,{\scriptstyle \pm3}$ & $0.044\,{\scriptstyle \pm2}$ & $0.094\,{\scriptstyle \pm1}$ & $1.077\,{\scriptstyle \pm62}$ & $0.707\,{\scriptstyle \pm4}$ & $0.409\,{\scriptstyle \pm14}$ & $0.011\,{\scriptstyle {<}1}$ & $10.227\,{\scriptstyle \pm167}$ & $2.344\,{\scriptstyle \pm18}$ & $1.024\,{\scriptstyle \pm20}$ \\

$\text{PBG 85M, seed 43}$ & $0.274\,{\scriptstyle \pm2}$ & $0.388\,{\scriptstyle \pm25}$ & $0.154\,{\scriptstyle \pm4}$ & $0.045\,{\scriptstyle \pm1}$ & $0.086\,{\scriptstyle {<}1}$ & $1.288\,{\scriptstyle \pm56}$ & $0.729\,{\scriptstyle \pm6}$ & $0.419\,{\scriptstyle \pm13}$ & $0.010\,{\scriptstyle {<}1}$ & $10.402\,{\scriptstyle \pm391}$ & $2.379\,{\scriptstyle \pm8}$ & $1.021\,{\scriptstyle \pm29}$ \\
\midrule
$\text{PBG 160M, seed 42}$ & $0.429\,{\scriptstyle \pm1}$ & $0.286\,{\scriptstyle \pm8}$ & $\underline{0.114}\,{\scriptstyle \pm8}$ & $0.044\,{\scriptstyle \pm1}$ & $0.184\,{\scriptstyle \pm1}$ & $0.656\,{\scriptstyle \pm28}$ & $0.614\,{\scriptstyle \pm10}$ & $0.389\,{\scriptstyle \pm15}$ & $0.024\,{\scriptstyle \pm1}$ & $6.049\,{\scriptstyle \pm489}$ & $2.111\,{\scriptstyle \pm11}$ & $0.983\,{\scriptstyle \pm40}$ \\

$\text{PBG 160M, seed 43}$ & $0.433\,{\scriptstyle {<}1}$ & $0.275\,{\scriptstyle \pm3}$ & $0.117\,{\scriptstyle \pm6}$ & $\mathbf{0.042}\,{\scriptstyle \pm4}$ & $0.178\,{\scriptstyle \pm2}$ & $0.693\,{\scriptstyle \pm43}$ & $0.607\,{\scriptstyle \pm13}$ & $\underline{0.369}\,{\scriptstyle \pm25}$ & $0.024\,{\scriptstyle \pm1}$ & $5.809\,{\scriptstyle \pm237}$ & $2.116\,{\scriptstyle \pm3}$ & $0.983\,{\scriptstyle \pm13}$ \\
\midrule
$\text{PBG 320M, seed 42}$ & $\underline{0.481}\,{\scriptstyle \pm2}$ & $\underline{0.268}\,{\scriptstyle \pm9}$ & $0.117\,{\scriptstyle \pm4}$ & $\underline{0.043}\,{\scriptstyle {<}1}$ & $\underline{0.239}\,{\scriptstyle \pm1}$ & $\underline{0.536}\,{\scriptstyle \pm12}$ & $\underline{0.574}\,{\scriptstyle \pm5}$ & $\mathbf{0.351}\,{\scriptstyle \pm9}$ & $\underline{0.035}\,{\scriptstyle \pm1}$ & $\underline{4.588}\,{\scriptstyle \pm81}$ & $\underline{2.005}\,{\scriptstyle \pm9}$ & $\mathbf{0.939}\,{\scriptstyle \pm22}$ \\

$\text{PBG 320M, seed 43}$ & $0.466\,{\scriptstyle \pm2}$ & $0.280\,{\scriptstyle \pm16}$ & $0.115\,{\scriptstyle \pm4}$ & $0.043\,{\scriptstyle \pm1}$ & $0.223\,{\scriptstyle \pm1}$ & $0.553\,{\scriptstyle \pm6}$ & $0.587\,{\scriptstyle \pm10}$ & $0.384\,{\scriptstyle \pm9}$ & $0.033\,{\scriptstyle \pm1}$ & $5.091\,{\scriptstyle \pm118}$ & $2.046\,{\scriptstyle \pm14}$ & $0.990\,{\scriptstyle \pm29}$ \\
\midrule
$\text{PBG 640M, seed 42}$ & $\mathbf{0.657}\,{\scriptstyle \pm2}$ & $\mathbf{0.259}\,{\scriptstyle \pm4}$ & $\mathbf{0.107}\,{\scriptstyle \pm8}$ & $0.043\,{\scriptstyle \pm2}$ & $\mathbf{0.399}\,{\scriptstyle \pm1}$ & $\mathbf{0.438}\,{\scriptstyle \pm29}$ & $\mathbf{0.545}\,{\scriptstyle \pm3}$ & $0.413\,{\scriptstyle \pm7}$ & $\mathbf{0.078}\,{\scriptstyle \pm2}$ & $\mathbf{2.631}\,{\scriptstyle \pm102}$ & $\mathbf{1.823}\,{\scriptstyle \pm6}$ & $\underline{0.955}\,{\scriptstyle \pm11}$ \\
\bottomrule
\end{tabular}%
}

%% file: generated/sota_stereo_validity.tex
\resizebox{\ifdim\width>\linewidth\linewidth\else\width\fi}{!}{%
\begin{tabular}{lcccccccccccc}
\toprule
\% of samples & \multicolumn{4}{c}{$2\mathrm{AA}$ (30 systems)} & \multicolumn{4}{c}{$4\mathrm{AA}$ (30 systems)} & \multicolumn{4}{c}{$8\mathrm{AA}$ (29 systems)\,\P} \\
\cmidrule(lr){2-5} \cmidrule(lr){6-9} \cmidrule(lr){10-13}
Model & Gen. & Refl. & Excl. & Worst & Gen. & Refl. & Excl. & Worst & Gen. & Refl. & Excl. & Worst \\
\midrule
ROBIN ($T=1.0$) & 99.7 & 0.0 & 0.3 & 0.4 & 99.8 & 0.0 & 0.2 & 0.3 & 99.7 & 0.0 & 0.3 & 0.8 \\
\midrule
PBG 30M & 49.9 & 49.9 & 0.2 & 0.5 & 49.8 & 49.8 & 0.4 & 1.0 & 49.5 & 49.6 & 1.0 & 1.9 \\
PBG 49M & 49.9 & 49.9 & 0.2 & 0.5 & 49.8 & 49.8 & 0.4 & 1.2 & 49.4 & 49.5 & 1.2 & 2.7 \\
PBG 85M & 49.9 & 50.0 & 0.1 & 0.3 & 49.9 & 49.8 & 0.2 & 0.8 & 49.8 & 49.7 & 0.5 & 1.3 \\
PBG 160M & 49.5 & 49.5 & 1.0 & 1.8 & 48.5 & 48.6 & 2.9 & 4.0 & 46.3 & 46.2 & 7.5 & 9.3 \\
PBG 320M & 49.7 & 49.6 & 0.7 & 1.1 & 49.5 & 49.5 & 1.0 & 1.5 & 49.2 & 49.2 & 1.7 & 2.5 \\
PBG 640M & 49.9 & 50.0 & 0.1 & 0.1 & 50.0 & 49.9 & 0.1 & 0.2 & 50.0 & 49.8 & 0.2 & 0.5 \\
\midrule
PBG-Joint 160M & 49.7 & 49.6 & 0.8 & 1.4 & 48.8 & 48.9 & 2.4 & 3.6 & 46.8 & 46.8 & 6.3 & 7.7 \\
\bottomrule
\end{tabular}%
}

%% file: generated/dipper_scaling_stereo.tex
\resizebox{\ifdim\width>\linewidth\linewidth\else\width\fi}{!}{%
\begin{tabular}{lrcccc}
\toprule
Size & Ckpts & $D/N$ & Refl.\ (\%) & Gen.\ (\%) & Excl.\ (\%) \\
\midrule
8M & 7 & 191--192 & 30.6--41.2 & 30.6--41.1 & 17.7--38.8 \\
15M & 13 & 96--193 & 34.7--46.9 & 34.6--47.0 & 6.1--30.7 \\
30M & 36 & 24--1760 & 41.1--49.8 & 41.1--49.7 & 0.5--17.7 \\
49M & 42 & 12--863 & 42.5--49.9 & 42.5--49.8 & 0.5--15.0 \\
85M & 32 & 12--768 & 0.1--49.8 & 45.5--99.5 & 0.3--9.1 \\
\bottomrule
\end{tabular}%
}

%% file: generated/sota_stereo_by_stereocenters.tex
\resizebox{\ifdim\width>\linewidth\linewidth\else\width\fi}{!}{%
\begin{tabular}{rrlcccc}
\toprule
Centers & Systems & Lengths & Gen.\ (\%) & Refl.\ (\%) & Excl.\ (\%) & Worst (\%) \\
\midrule
1 & 3 & 2AA & 50.0 & 50.0 & 0.00 & 0.00 \\
2 & 23 & 2AA, 4AA & 49.7 & 49.8 & 0.47 & 0.61 \\
3 & 8 & 2AA, 4AA & 49.7 & 49.6 & 0.66 & 0.86 \\
4 & 16 & 4AA & 49.5 & 49.5 & 1.06 & 1.47 \\
5 & 9 & 4AA & 49.4 & 49.4 & 1.21 & 1.76 \\
6 & 1 & 4AA & 49.4 & 49.5 & 1.15 & 1.15 \\
7 & 6 & 8AA & 48.8 & 48.7 & 2.51 & 3.25 \\
8 & 11 & 8AA & 48.7 & 48.7 & 2.59 & 3.39 \\
9 & 8 & 8AA & 48.7 & 48.7 & 2.55 & 2.87 \\
10 & 2 & 8AA & 48.5 & 48.6 & 2.90 & 3.07 \\
11 & 2 & 8AA & 48.5 & 48.5 & 3.00 & 3.29 \\
\bottomrule
\end{tabular}%
}

%% file: generated/robin_pbg_ladder_stereo.tex
\resizebox{\ifdim\width>\linewidth\linewidth\else\width\fi}{!}{%
\begin{tabular}{llcccccc}
\toprule
 & & \multicolumn{2}{c}{30M} & \multicolumn{2}{c}{85M} & \multicolumn{2}{c}{160M} \\
\cmidrule(lr){3-4} \cmidrule(lr){5-6} \cmidrule(lr){7-8}
Rung & Change & Refl. & Excl. & Refl. & Excl. & Refl. & Excl. \\
\midrule
L0 & ROBIN protocol & 15.2 & 69.0 & 20.9 & 44.3 & 0.8 & 5.6 \\
L1 & Scalable Muon & 0.8 & 6.4 & 49.2 & 1.7 & 0.8 & 1.3 \\
L2 & WSD schedule & 45.3 & 9.5 & 48.9 & 2.2 & 49.5 & 1.1 \\
L3 & bfloat16 precision & 46.2 & 7.7 & 48.7 & 2.6 & 0.3 & 1.9 \\
L4 & Packed batches & 1.9 & 7.9 & 48.7 & 2.5 & 0.4 & 1.5 \\
L5 & Token-mean loss & 0.8 & 13.2 & 48.6 & 2.8 & 0.5 & 2.2 \\
L6 & 0.8 pm quantization & 0.5 & 7.4 & 0.3 & 2.0 & 0.2 & 1.2 \\
L7 & Random translation & 48.5 & 3.1 & 49.7 & 0.6 & 49.5 & 1.2 \\
L8 & HELM prompt (PBG) & 48.2 & 3.5 & 49.4 & 1.2 & 49.0 & 2.1 \\
\bottomrule
\end{tabular}%
}

%% file: generated/many_peptides_dpro_prev_residue.tex
\resizebox{\linewidth}{!}{%
\begin{tabular}{lrrrrrrr}
\toprule
 & & \multicolumn{3}{c}{Published data} & \multicolumn{3}{c}{Reproduced} \\
\cmidrule(lr){3-5} \cmidrule(lr){6-8}
Residue before Pro & Prolines & Clash & D-Pro & D-Pro (\%) & Prolines & D-Pro & D-Pro (\%) \\
\midrule
Phe & 284 & 284 & 270 & 95.1 & 284 & 270 & 95.1 \\
Tyr & 296 & 296 & 6 & 2.0 & 296 & 6 & 2.0 \\
Trp & 323 & 323 & 0 & 0.0 & 323 & 0 & 0.0 \\
His & 289 & 289 & 0 & 0.0 & 289 & 0 & 0.0 \\
Asn & 326 & 326 & 0 & 0.0 & 326 & 0 & 0.0 \\
Other or N-terminus & 5{,}685 & 0 & 0 & 0.0 & 534 & 0 & 0.0 \\
\midrule
All & 7{,}203 & 1{,}518 & 276 & 3.8 & 2{,}052 & 276 & 13.5 \\
\bottomrule
\end{tabular}%
}

%% file: generated/sota_30_plot.tex
\resizebox{\linewidth}{!}{%
\begin{tabular}{lcccccccccccc}
\toprule
 & \multicolumn{4}{c}{$2\mathrm{AA}$ (30 systems)} & \multicolumn{4}{c}{$4\mathrm{AA}$ (30 systems)} & \multicolumn{4}{c}{$8\mathrm{AA}$ (30 systems)} \\
\cmidrule(lr){2-5} \cmidrule(lr){6-9} \cmidrule(lr){10-13}
Model & $\mathrm{rESS}\uparrow$ & $\mathcal{E}\text{-}\mathcal{W}_2\downarrow$ & $\mathbb{T}\text{-}\mathcal{W}_2\downarrow$ & $\mathrm{TICA}\text{-}\mathcal{W}_2\downarrow$ & $\mathrm{rESS}\uparrow$ & $\mathcal{E}\text{-}\mathcal{W}_2\downarrow$ & $\mathbb{T}\text{-}\mathcal{W}_2\downarrow$ & $\mathrm{TICA}\text{-}\mathcal{W}_2\downarrow$ & $\mathrm{rESS}\uparrow$ & $\mathcal{E}\text{-}\mathcal{W}_2\downarrow$ & $\mathbb{T}\text{-}\mathcal{W}_2\downarrow$ & $\mathrm{TICA}\text{-}\mathcal{W}_2\downarrow$ \\
\midrule

$\text{TarFlow}$ & $0.134$ & $0.452$ & $0.193$ & $\text{---}$ & $0.045$ & $1.260$ & $0.924$ & $0.492$ & $0.008$ & $11.298$ & $2.733$ & $1.087$ \\

$\text{Prose}$ & $0.191$ & $0.371$ & $0.210$ & $\text{---}$ & $0.071$ & $0.932$ & $0.752$ & $\mathbf{0.367}$ & $0.011$ & $10.038$ & $2.456$ & $0.988$ \\

$\text{ROBIN SMC}$ & $\text{---}$ & $\text{---}$ & $\text{---}$ & $\text{---}$ & $\text{---}$ & $1.079$ & $0.874$ & $0.463$ & $\text{---}$ & $4.263$ & $2.315$ & $0.977$ \\

$\text{ROBIN}\ (T=0.9)$ & $\text{---}$ & $\text{---}$ & $\text{---}$ & $\text{---}$ & $0.031$ & $1.168$ & $0.886$ & $0.471$ & $0.013$ & $4.251$ & $2.325$ & $\mathbf{0.943}$ \\
\midrule
$\text{ROBIN}\ (T=1.0)$ & $0.036\,{\scriptstyle \pm1}$ & $0.689\,{\scriptstyle \pm23}$ & $0.242\,{\scriptstyle \pm14}$ & $0.062\,{\scriptstyle \pm5}$ & $0.031\,{\scriptstyle {<}1}$ & $1.210\,{\scriptstyle \pm76}$ & $0.813\,{\scriptstyle \pm11}$ & $0.454\,{\scriptstyle \pm25}$ & $0.014\,{\scriptstyle {<}1}$ & $6.033\,{\scriptstyle \pm146}$ & $2.250\,{\scriptstyle \pm12}$ & $0.947\,{\scriptstyle \pm9}$ \\
\midrule
$\text{PBG 30M}\ (D/N = 1760)$ & $0.451\,{\scriptstyle \pm2}$ & $0.288\,{\scriptstyle \pm22}$ & $0.117\,{\scriptstyle \pm6}$ & $0.043\,{\scriptstyle \pm3}$ & $0.181\,{\scriptstyle \pm1}$ & $0.725\,{\scriptstyle \pm67}$ & $0.634\,{\scriptstyle \pm6}$ & $0.414\,{\scriptstyle \pm21}$ & $0.021\,{\scriptstyle \pm1}$ & $6.790\,{\scriptstyle \pm134}$ & $2.170\,{\scriptstyle \pm10}$ & $1.018\,{\scriptstyle \pm29}$ \\

$\text{PBG 49M}\ (D/N = 416)$ & $0.420\,{\scriptstyle \pm2}$ & $0.293\,{\scriptstyle {<}1}$ & $0.123\,{\scriptstyle \pm10}$ & $\mathbf{0.041}\,{\scriptstyle \pm1}$ & $0.161\,{\scriptstyle \pm1}$ & $0.749\,{\scriptstyle \pm32}$ & $0.639\,{\scriptstyle \pm5}$ & $0.417\,{\scriptstyle \pm27}$ & $0.018\,{\scriptstyle {<}1}$ & $6.721\,{\scriptstyle \pm173}$ & $2.182\,{\scriptstyle \pm13}$ & $1.030\,{\scriptstyle \pm15}$ \\

$\text{PBG 85M}\ (D/N = 768)$ & $\underline{0.573}\,{\scriptstyle \pm2}$ & $\underline{0.264}\,{\scriptstyle \pm10}$ & $0.111\,{\scriptstyle \pm3}$ & $0.043\,{\scriptstyle \pm1}$ & $\underline{0.294}\,{\scriptstyle \pm1}$ & $\underline{0.496}\,{\scriptstyle \pm19}$ & $0.571\,{\scriptstyle \pm7}$ & $\underline{0.374}\,{\scriptstyle \pm8}$ & $\underline{0.040}\,{\scriptstyle {<}1}$ & $4.097\,{\scriptstyle \pm185}$ & $1.985\,{\scriptstyle \pm6}$ & $0.993\,{\scriptstyle \pm17}$ \\

$\text{PBG 160M}\ (D/N = 352)$ & $0.448\,{\scriptstyle \pm2}$ & $0.289\,{\scriptstyle \pm12}$ & $0.114\,{\scriptstyle \pm1}$ & $0.043\,{\scriptstyle \pm4}$ & $0.216\,{\scriptstyle \pm9}$ & $0.538\,{\scriptstyle \pm31}$ & $0.587\,{\scriptstyle \pm4}$ & $0.403\,{\scriptstyle \pm23}$ & $0.035\,{\scriptstyle {<}1}$ & $\underline{3.714}\,{\scriptstyle \pm123}$ & $1.987\,{\scriptstyle \pm15}$ & $1.005\,{\scriptstyle \pm15}$ \\

$\text{PBG 320M}\ (D/N = 86)$ & $0.501\,{\scriptstyle \pm1}$ & $0.274\,{\scriptstyle \pm9}$ & $\underline{0.107}\,{\scriptstyle \pm10}$ & $0.042\,{\scriptstyle {<}1}$ & $0.260\,{\scriptstyle \pm2}$ & $0.500\,{\scriptstyle \pm48}$ & $\underline{0.569}\,{\scriptstyle \pm6}$ & $0.374\,{\scriptstyle \pm19}$ & $0.040\,{\scriptstyle {<}1}$ & $3.790\,{\scriptstyle \pm318}$ & $\underline{1.973}\,{\scriptstyle \pm11}$ & $0.969\,{\scriptstyle \pm8}$ \\

$\text{PBG 640M}\ (D/N = 48)$ & $\mathbf{0.657}\,{\scriptstyle \pm2}$ & $\mathbf{0.259}\,{\scriptstyle \pm4}$ & $\mathbf{0.107}\,{\scriptstyle \pm8}$ & $0.043\,{\scriptstyle \pm2}$ & $\mathbf{0.399}\,{\scriptstyle \pm1}$ & $\mathbf{0.438}\,{\scriptstyle \pm29}$ & $\mathbf{0.545}\,{\scriptstyle \pm3}$ & $0.413\,{\scriptstyle \pm7}$ & $\mathbf{0.077}\,{\scriptstyle \pm2}$ & $\mathbf{2.597}\,{\scriptstyle \pm117}$ & $\mathbf{1.824}\,{\scriptstyle \pm4}$ & $\underline{0.944}\,{\scriptstyle \pm8}$ \\
\midrule
$\text{PBG-Joint 160M}\ (D/N = 198)$ & $0.439\,{\scriptstyle \pm1}$ & $0.278\,{\scriptstyle \pm14}$ & $0.114\,{\scriptstyle \pm3}$ & $\underline{0.041}\,{\scriptstyle \pm2}$ & $0.198\,{\scriptstyle \pm2}$ & $0.580\,{\scriptstyle \pm5}$ & $0.589\,{\scriptstyle \pm5}$ & $0.378\,{\scriptstyle \pm18}$ & $0.032\,{\scriptstyle \pm1}$ & $4.065\,{\scriptstyle \pm475}$ & $1.981\,{\scriptstyle \pm11}$ & $0.947\,{\scriptstyle \pm11}$ \\
\bottomrule
\end{tabular}%
}

%% file: generated/robin_pbg_ladder.tex
\resizebox{\linewidth}{!}{%
\begin{tabular}{lllccccccccc}
\toprule
 & & & \multicolumn{3}{c}{$2\mathrm{AA}$ (30 systems)} & \multicolumn{3}{c}{$4\mathrm{AA}$ (30 systems)} & \multicolumn{3}{c}{$8\mathrm{AA}$ (29 systems)} \\
\cmidrule(lr){4-6} \cmidrule(lr){7-9} \cmidrule(lr){10-12}
Rung & Change & Size & $\mathrm{rESS}^{\ast}\uparrow$ & $\mathcal{E}\text{-}\mathcal{W}_2^{\ast}\downarrow$ & $\mathbb{T}\mathrm{g}\text{-}\mathcal{SW}_2^{\ast}\downarrow$ & $\mathrm{rESS}^{\ast}\uparrow$ & $\mathcal{E}\text{-}\mathcal{W}_2^{\ast}\downarrow$ & $\mathbb{T}\mathrm{g}\text{-}\mathcal{SW}_2^{\ast}\downarrow$ & $\mathrm{rESS}^{\ast}\uparrow$ & $\mathcal{E}\text{-}\mathcal{W}_2^{\ast}\downarrow$ & $\mathbb{T}\mathrm{g}\text{-}\mathcal{SW}_2^{\ast}\downarrow$ \\
\midrule
L0 & ROBIN protocol & 30M & $0.000\,{\scriptstyle {<}1}$ & $56.058\,{\scriptstyle \pm280}$ & $1.295\,{\scriptstyle \pm13}$ & $0.000\,{\scriptstyle {<}1}$ & $185.174\,{\scriptstyle \pm2295}$ & $1.364\,{\scriptstyle \pm20}$ & \text{---} & \text{---} & \text{---} \\
 &  & 85M & $0.001\,{\scriptstyle {<}1}$ & $5.564\,{\scriptstyle \pm560}$ & $0.852\,{\scriptstyle \pm97}$ & $0.000\,{\scriptstyle {<}1}$ & $11.030\,{\scriptstyle \pm657}$ & $0.941\,{\scriptstyle \pm51}$ & $0.000\,{\scriptstyle {<}1}$ & $53.260\,{\scriptstyle \pm2270}$ & $1.180\,{\scriptstyle \pm19}$ \\
 &  & 160M & $0.002\,{\scriptstyle {<}1}$ & $3.382\,{\scriptstyle \pm418}$ & $0.607\,{\scriptstyle \pm34}$ & $0.001\,{\scriptstyle {<}1}$ & $6.329\,{\scriptstyle \pm271}$ & $0.709\,{\scriptstyle \pm39}$ & $0.000\,{\scriptstyle {<}1}$ & $21.077\,{\scriptstyle \pm943}$ & $1.002\,{\scriptstyle \pm48}$ \\
\addlinespace
L1 & + Scalable Muon & 30M & $0.001\,{\scriptstyle {<}1}$ & $3.904\,{\scriptstyle \pm391}$ & $0.632\,{\scriptstyle \pm19}$ & $0.001\,{\scriptstyle {<}1}$ & $7.270\,{\scriptstyle \pm1263}$ & $0.783\,{\scriptstyle \pm84}$ & $0.000\,{\scriptstyle {<}1}$ & $21.949\,{\scriptstyle \pm2060}$ & $1.021\,{\scriptstyle \pm42}$ \\
 &  & 85M & $0.001\,{\scriptstyle {<}1}$ & $4.064\,{\scriptstyle \pm312}$ & $0.669\,{\scriptstyle \pm58}$ & $0.001\,{\scriptstyle {<}1}$ & $6.696\,{\scriptstyle \pm548}$ & $0.687\,{\scriptstyle \pm34}$ & $0.000\,{\scriptstyle {<}1}$ & $16.220\,{\scriptstyle \pm1249}$ & $0.877\,{\scriptstyle \pm49}$ \\
 &  & 160M & $0.002\,{\scriptstyle {<}1}$ & $3.816\,{\scriptstyle \pm784}$ & $0.649\,{\scriptstyle \pm81}$ & $0.001\,{\scriptstyle {<}1}$ & $5.496\,{\scriptstyle \pm84}$ & $0.670\,{\scriptstyle \pm40}$ & $0.001\,{\scriptstyle {<}1}$ & $11.248\,{\scriptstyle \pm809}$ & $0.730\,{\scriptstyle \pm26}$ \\
\addlinespace
L2 & + WSD schedule & 30M & $0.001\,{\scriptstyle {<}1}$ & $3.963\,{\scriptstyle \pm504}$ & $0.672\,{\scriptstyle \pm52}$ & $0.001\,{\scriptstyle {<}1}$ & $7.005\,{\scriptstyle \pm912}$ & $0.778\,{\scriptstyle \pm33}$ & $0.000\,{\scriptstyle {<}1}$ & $22.348\,{\scriptstyle \pm1171}$ & $1.021\,{\scriptstyle \pm27}$ \\
 &  & 85M & $0.001\,{\scriptstyle {<}1}$ & $4.001\,{\scriptstyle \pm744}$ & $0.650\,{\scriptstyle \pm99}$ & $0.001\,{\scriptstyle {<}1}$ & $5.755\,{\scriptstyle \pm142}$ & $0.665\,{\scriptstyle \pm19}$ & $0.000\,{\scriptstyle {<}1}$ & $15.767\,{\scriptstyle \pm2867}$ & $0.853\,{\scriptstyle \pm49}$ \\
 &  & 160M & $0.002\,{\scriptstyle {<}1}$ & $3.605\,{\scriptstyle \pm216}$ & $0.635\,{\scriptstyle \pm22}$ & $0.001\,{\scriptstyle {<}1}$ & $5.652\,{\scriptstyle \pm784}$ & $0.634\,{\scriptstyle \pm70}$ & $0.001\,{\scriptstyle {<}1}$ & $11.470\,{\scriptstyle \pm1203}$ & $0.753\,{\scriptstyle \pm15}$ \\
\addlinespace
L3 & + bfloat16 precision & 30M & $0.001\,{\scriptstyle {<}1}$ & $4.189\,{\scriptstyle \pm164}$ & $0.751\,{\scriptstyle \pm58}$ & $0.001\,{\scriptstyle {<}1}$ & $7.503\,{\scriptstyle \pm810}$ & $0.766\,{\scriptstyle \pm25}$ & $0.000\,{\scriptstyle {<}1}$ & $23.400\,{\scriptstyle \pm571}$ & $0.988\,{\scriptstyle \pm71}$ \\
 &  & 85M & $0.002\,{\scriptstyle {<}1}$ & $3.791\,{\scriptstyle \pm198}$ & $0.661\,{\scriptstyle \pm13}$ & $0.001\,{\scriptstyle {<}1}$ & $5.671\,{\scriptstyle \pm1066}$ & $0.667\,{\scriptstyle \pm106}$ & $0.001\,{\scriptstyle {<}1}$ & $16.076\,{\scriptstyle \pm2306}$ & $0.852\,{\scriptstyle \pm22}$ \\
 &  & 160M & $0.002\,{\scriptstyle {<}1}$ & $3.818\,{\scriptstyle \pm115}$ & $0.645\,{\scriptstyle \pm20}$ & $0.002\,{\scriptstyle {<}1}$ & $4.984\,{\scriptstyle \pm334}$ & $0.589\,{\scriptstyle \pm20}$ & $0.001\,{\scriptstyle {<}1}$ & $12.859\,{\scriptstyle \pm516}$ & $0.782\,{\scriptstyle \pm38}$ \\
\addlinespace
L4 & + Packed batches & 30M & $0.001\,{\scriptstyle {<}1}$ & $4.210\,{\scriptstyle \pm336}$ & $0.681\,{\scriptstyle \pm48}$ & $0.001\,{\scriptstyle {<}1}$ & $7.929\,{\scriptstyle \pm429}$ & $0.823\,{\scriptstyle \pm74}$ & $0.000\,{\scriptstyle {<}1}$ & $23.716\,{\scriptstyle \pm778}$ & $1.006\,{\scriptstyle \pm28}$ \\
 &  & 85M & $0.001\,{\scriptstyle {<}1}$ & $3.639\,{\scriptstyle \pm251}$ & $0.674\,{\scriptstyle \pm29}$ & $0.001\,{\scriptstyle {<}1}$ & $6.555\,{\scriptstyle \pm447}$ & $0.688\,{\scriptstyle \pm62}$ & $0.001\,{\scriptstyle {<}1}$ & $15.597\,{\scriptstyle \pm520}$ & $0.820\,{\scriptstyle \pm42}$ \\
 &  & 160M & $0.002\,{\scriptstyle {<}1}$ & $3.604\,{\scriptstyle \pm298}$ & $0.621\,{\scriptstyle \pm37}$ & $0.002\,{\scriptstyle {<}1}$ & $4.941\,{\scriptstyle \pm933}$ & $0.598\,{\scriptstyle \pm79}$ & $0.001\,{\scriptstyle {<}1}$ & $11.936\,{\scriptstyle \pm109}$ & $0.770\,{\scriptstyle \pm27}$ \\
\addlinespace
L5 & + Token-mean loss & 30M & $0.001\,{\scriptstyle {<}1}$ & $3.777\,{\scriptstyle \pm610}$ & $0.638\,{\scriptstyle \pm48}$ & $0.001\,{\scriptstyle {<}1}$ & $8.009\,{\scriptstyle \pm918}$ & $0.834\,{\scriptstyle \pm43}$ & $0.000\,{\scriptstyle {<}1}$ & $25.122\,{\scriptstyle \pm4040}$ & $1.009\,{\scriptstyle \pm46}$ \\
 &  & 85M & $0.001\,{\scriptstyle {<}1}$ & $3.590\,{\scriptstyle \pm273}$ & $0.627\,{\scriptstyle \pm51}$ & $0.001\,{\scriptstyle {<}1}$ & $6.206\,{\scriptstyle \pm1083}$ & $0.669\,{\scriptstyle \pm20}$ & $0.000\,{\scriptstyle {<}1}$ & $16.404\,{\scriptstyle \pm256}$ & $0.854\,{\scriptstyle \pm26}$ \\
 &  & 160M & $0.002\,{\scriptstyle {<}1}$ & $4.084\,{\scriptstyle \pm508}$ & $0.663\,{\scriptstyle \pm71}$ & $0.001\,{\scriptstyle {<}1}$ & $5.639\,{\scriptstyle \pm810}$ & $0.615\,{\scriptstyle \pm38}$ & $0.001\,{\scriptstyle {<}1}$ & $11.593\,{\scriptstyle \pm1352}$ & $0.736\,{\scriptstyle \pm57}$ \\
\addlinespace
L6 & + 0.8 pm quantization & 30M & $0.001\,{\scriptstyle {<}1}$ & $4.000\,{\scriptstyle \pm289}$ & $0.724\,{\scriptstyle \pm64}$ & $0.001\,{\scriptstyle {<}1}$ & $7.701\,{\scriptstyle \pm297}$ & $0.753\,{\scriptstyle \pm30}$ & $0.000\,{\scriptstyle {<}1}$ & $23.112\,{\scriptstyle \pm739}$ & $0.964\,{\scriptstyle \pm58}$ \\
 &  & 85M & $0.001\,{\scriptstyle {<}1}$ & $3.753\,{\scriptstyle \pm619}$ & $0.678\,{\scriptstyle \pm41}$ & $0.001\,{\scriptstyle {<}1}$ & $5.945\,{\scriptstyle \pm167}$ & $0.644\,{\scriptstyle \pm37}$ & $0.001\,{\scriptstyle {<}1}$ & $15.380\,{\scriptstyle \pm1122}$ & $0.825\,{\scriptstyle \pm15}$ \\
 &  & 160M & $0.002\,{\scriptstyle \pm1}$ & $3.504\,{\scriptstyle \pm296}$ & $0.580\,{\scriptstyle \pm69}$ & $0.002\,{\scriptstyle {<}1}$ & $6.213\,{\scriptstyle \pm809}$ & $0.660\,{\scriptstyle \pm41}$ & $0.001\,{\scriptstyle {<}1}$ & $11.453\,{\scriptstyle \pm353}$ & $0.710\,{\scriptstyle \pm54}$ \\
\addlinespace
L7 & + Random translation & 30M & $0.010\,{\scriptstyle \pm3}$ & $2.321\,{\scriptstyle \pm359}$ & $0.423\,{\scriptstyle \pm85}$ & $0.002\,{\scriptstyle {<}1}$ & $5.434\,{\scriptstyle \pm597}$ & $0.581\,{\scriptstyle \pm25}$ & $0.000\,{\scriptstyle {<}1}$ & $21.789\,{\scriptstyle \pm815}$ & $0.910\,{\scriptstyle \pm34}$ \\
 &  & 85M & $0.062\,{\scriptstyle \pm15}$ & $2.812\,{\scriptstyle \pm303}$ & $0.465\,{\scriptstyle \pm111}$ & $0.009\,{\scriptstyle \pm5}$ & $4.608\,{\scriptstyle \pm504}$ & $0.560\,{\scriptstyle \pm49}$ & $0.001\,{\scriptstyle {<}1}$ & $12.322\,{\scriptstyle \pm1040}$ & $0.708\,{\scriptstyle \pm60}$ \\
 &  & 160M & $0.191\,{\scriptstyle \pm17}$ & $2.345\,{\scriptstyle \pm263}$ & $0.410\,{\scriptstyle \pm20}$ & $0.045\,{\scriptstyle \pm10}$ & $3.634\,{\scriptstyle \pm147}$ & $0.437\,{\scriptstyle \pm16}$ & $0.003\,{\scriptstyle \pm1}$ & $8.510\,{\scriptstyle \pm429}$ & $0.564\,{\scriptstyle \pm50}$ \\
\addlinespace
L8 & + HELM prompt (PBG) & 30M & $0.014\,{\scriptstyle \pm3}$ & $2.048\,{\scriptstyle \pm525}$ & $0.386\,{\scriptstyle \pm55}$ & $0.002\,{\scriptstyle {<}1}$ & $5.418\,{\scriptstyle \pm774}$ & $0.607\,{\scriptstyle \pm73}$ & $0.000\,{\scriptstyle {<}1}$ & $20.645\,{\scriptstyle \pm2307}$ & $0.897\,{\scriptstyle \pm87}$ \\
 &  & 85M & $0.120\,{\scriptstyle \pm14}$ & $1.245\,{\scriptstyle \pm91}$ & $0.253\,{\scriptstyle \pm23}$ & $0.019\,{\scriptstyle \pm1}$ & $3.619\,{\scriptstyle \pm702}$ & $0.440\,{\scriptstyle \pm56}$ & $0.001\,{\scriptstyle {<}1}$ & $10.627\,{\scriptstyle \pm915}$ & $0.653\,{\scriptstyle \pm33}$ \\
 &  & 160M & $0.199\,{\scriptstyle \pm64}$ & $2.394\,{\scriptstyle \pm1393}$ & $0.379\,{\scriptstyle \pm146}$ & $0.048\,{\scriptstyle \pm10}$ & $3.915\,{\scriptstyle \pm1736}$ & $0.412\,{\scriptstyle \pm101}$ & $0.003\,{\scriptstyle {<}1}$ & $8.632\,{\scriptstyle \pm1521}$ & $0.570\,{\scriptstyle \pm33}$ \\
\bottomrule
\end{tabular}%
}

%% file: generated/joint_matched_peptide_control.tex
\resizebox{\linewidth}{!}{%
\begin{tabular}{lcccccccccccc}
\toprule
 & \multicolumn{4}{c}{$2\mathrm{AA}$ (30 systems)} & \multicolumn{4}{c}{$4\mathrm{AA}$ (30 systems)} & \multicolumn{4}{c}{$8\mathrm{AA}$ (29 systems)\,\P} \\
\cmidrule(lr){2-5} \cmidrule(lr){6-9} \cmidrule(lr){10-13}
Model & $\mathrm{rESS}\uparrow$ & $\mathcal{E}\text{-}\mathcal{W}_2\downarrow$ & $\mathbb{T}\text{-}\mathcal{W}_2\downarrow$ & $\mathrm{TICA}\text{-}\mathcal{W}_2\downarrow$ & $\mathrm{rESS}\uparrow$ & $\mathcal{E}\text{-}\mathcal{W}_2\downarrow$ & $\mathbb{T}\text{-}\mathcal{W}_2\downarrow$ & $\mathrm{TICA}\text{-}\mathcal{W}_2\downarrow$ & $\mathrm{rESS}\uparrow$ & $\mathcal{E}\text{-}\mathcal{W}_2\downarrow$ & $\mathbb{T}\text{-}\mathcal{W}_2\downarrow$ & $\mathrm{TICA}\text{-}\mathcal{W}_2\downarrow$ \\
\midrule
$\text{PBG-Joint 160M}\ (D/N = 198)$ & $0.439\,{\scriptstyle \pm1}$ & $\mathbf{0.278}\,{\scriptstyle \pm14}$ & $\mathbf{0.114}\,{\scriptstyle \pm3}$ & $\mathbf{0.041}\,{\scriptstyle \pm2}$ & $0.198\,{\scriptstyle \pm2}$ & $0.580\,{\scriptstyle \pm5}$ & $\mathbf{0.589}\,{\scriptstyle \pm5}$ & $\mathbf{0.378}\,{\scriptstyle \pm18}$ & $0.032\,{\scriptstyle \pm1}$ & $4.096\,{\scriptstyle \pm488}$ & $\mathbf{1.979}\,{\scriptstyle \pm12}$ & $\mathbf{0.957}\,{\scriptstyle \pm12}$ \\
$\text{PBG 160M}\ (D/N = 193)$ & $\mathbf{0.442}\,{\scriptstyle \pm1}$ & $0.280\,{\scriptstyle \pm14}$ & $0.116\,{\scriptstyle \pm4}$ & $0.044\,{\scriptstyle \pm2}$ & $\mathbf{0.211}\,{\scriptstyle \pm4}$ & $\mathbf{0.529}\,{\scriptstyle \pm31}$ & $0.592\,{\scriptstyle \pm5}$ & $0.417\,{\scriptstyle \pm12}$ & $\mathbf{0.034}\,{\scriptstyle \pm1}$ & $\mathbf{4.046}\,{\scriptstyle \pm160}$ & $1.994\,{\scriptstyle \pm5}$ & $1.022\,{\scriptstyle \pm7}$ \\
\midrule
PBG-Joint better & $9/30$ & $18/30$ & $14/30$ & $15/30$ & $3/30$ & $9/30$ & $18/30$ & $21/30$ & $11/29$ & $16/29$ & $16/29$ & $18/29$ \\
\bottomrule
\end{tabular}%
}

%% file: generated/box_volume_correction.tex
\resizebox{\linewidth}{!}{%
\begin{tabular}{lrlccccccccc}
\toprule
 & & & \multicolumn{3}{c}{$2\mathrm{AA}$ (30 systems)} & \multicolumn{3}{c}{$4\mathrm{AA}$ (30 systems)} & \multicolumn{3}{c}{$8\mathrm{AA}$ (29 systems)} \\
\cmidrule(lr){4-6} \cmidrule(lr){7-9} \cmidrule(lr){10-12}
Model & $D/N$ & Weights & $\mathrm{rESS}^{\ast}\uparrow$ & $\mathcal{E}\text{-}\mathcal{W}_2^{\ast}\downarrow$ & $\mathbb{T}\mathrm{g}\text{-}\mathcal{SW}_2^{\ast}\downarrow$ & $\mathrm{rESS}^{\ast}\uparrow$ & $\mathcal{E}\text{-}\mathcal{W}_2^{\ast}\downarrow$ & $\mathbb{T}\mathrm{g}\text{-}\mathcal{SW}_2^{\ast}\downarrow$ & $\mathrm{rESS}^{\ast}\uparrow$ & $\mathcal{E}\text{-}\mathcal{W}_2^{\ast}\downarrow$ & $\mathbb{T}\mathrm{g}\text{-}\mathcal{SW}_2^{\ast}\downarrow$ \\
\midrule
Reference MD & --- & standard & $0.997$ & $0.031$ & $0.043$ & $0.992$ & $0.103$ & $0.048$ & $0.977$ & $0.518$ & $0.057$ \\
\midrule
PBG 30M & 1760 & standard & $0.319\,{\scriptstyle \pm18}$ & $0.516\,{\scriptstyle \pm157}$ & $0.128\,{\scriptstyle \pm32}$ & $0.068\,{\scriptstyle \pm6}$ & $2.112\,{\scriptstyle \pm340}$ & $0.256\,{\scriptstyle \pm10}$ & $0.003\,{\scriptstyle \pm1}$ & $7.448\,{\scriptstyle \pm174}$ & $0.533\,{\scriptstyle \pm48}$ \\
 &  & corrected & $0.326\,{\scriptstyle \pm19}$ & $0.507\,{\scriptstyle \pm161}$ & $0.120\,{\scriptstyle \pm33}$ & $0.072\,{\scriptstyle \pm6}$ & $2.071\,{\scriptstyle \pm341}$ & $0.249\,{\scriptstyle \pm7}$ & $0.003\,{\scriptstyle \pm1}$ & $7.502\,{\scriptstyle \pm258}$ & $0.527\,{\scriptstyle \pm50}$ \\
\addlinespace
PBG 49M & 416 & standard & $0.304\,{\scriptstyle \pm18}$ & $0.391\,{\scriptstyle \pm191}$ & $0.104\,{\scriptstyle \pm30}$ & $0.057\,{\scriptstyle \pm10}$ & $1.379\,{\scriptstyle \pm220}$ & $0.203\,{\scriptstyle \pm29}$ & $0.002\,{\scriptstyle \pm1}$ & $8.284\,{\scriptstyle \pm791}$ & $0.564\,{\scriptstyle \pm23}$ \\
 &  & corrected & $0.310\,{\scriptstyle \pm17}$ & $0.382\,{\scriptstyle \pm187}$ & $0.095\,{\scriptstyle \pm30}$ & $0.060\,{\scriptstyle \pm9}$ & $1.354\,{\scriptstyle \pm226}$ & $0.198\,{\scriptstyle \pm27}$ & $0.003\,{\scriptstyle \pm1}$ & $8.286\,{\scriptstyle \pm855}$ & $0.558\,{\scriptstyle \pm20}$ \\
\addlinespace
PBG 85M & 768 & standard & $0.454\,{\scriptstyle \pm15}$ & $0.700\,{\scriptstyle \pm98}$ & $0.143\,{\scriptstyle \pm18}$ & $0.151\,{\scriptstyle \pm16}$ & $0.967\,{\scriptstyle \pm387}$ & $0.158\,{\scriptstyle \pm47}$ & $0.007\,{\scriptstyle \pm1}$ & $5.546\,{\scriptstyle \pm828}$ & $0.425\,{\scriptstyle \pm11}$ \\
 &  & corrected & $0.461\,{\scriptstyle \pm15}$ & $0.695\,{\scriptstyle \pm99}$ & $0.133\,{\scriptstyle \pm18}$ & $0.160\,{\scriptstyle \pm16}$ & $0.929\,{\scriptstyle \pm386}$ & $0.151\,{\scriptstyle \pm45}$ & $0.007\,{\scriptstyle \pm1}$ & $5.619\,{\scriptstyle \pm710}$ & $0.418\,{\scriptstyle \pm17}$ \\
\addlinespace
PBG 160M & 352 & standard & $0.285\,{\scriptstyle \pm52}$ & $1.722\,{\scriptstyle \pm432}$ & $0.282\,{\scriptstyle \pm47}$ & $0.094\,{\scriptstyle \pm20}$ & $2.573\,{\scriptstyle \pm956}$ & $0.304\,{\scriptstyle \pm60}$ & $0.005\,{\scriptstyle \pm1}$ & $6.110\,{\scriptstyle \pm193}$ & $0.485\,{\scriptstyle \pm26}$ \\
 &  & corrected & $0.287\,{\scriptstyle \pm52}$ & $1.718\,{\scriptstyle \pm444}$ & $0.275\,{\scriptstyle \pm49}$ & $0.098\,{\scriptstyle \pm21}$ & $2.581\,{\scriptstyle \pm978}$ & $0.299\,{\scriptstyle \pm59}$ & $0.006\,{\scriptstyle \pm1}$ & $6.167\,{\scriptstyle \pm224}$ & $0.479\,{\scriptstyle \pm28}$ \\
\addlinespace
PBG 320M & 86 & standard & $0.382\,{\scriptstyle \pm22}$ & $0.587\,{\scriptstyle \pm329}$ & $0.135\,{\scriptstyle \pm61}$ & $0.111\,{\scriptstyle \pm19}$ & $1.317\,{\scriptstyle \pm228}$ & $0.210\,{\scriptstyle \pm22}$ & $0.007\,{\scriptstyle \pm1}$ & $5.045\,{\scriptstyle \pm686}$ & $0.424\,{\scriptstyle \pm17}$ \\
 &  & corrected & $0.386\,{\scriptstyle \pm23}$ & $0.589\,{\scriptstyle \pm338}$ & $0.125\,{\scriptstyle \pm62}$ & $0.115\,{\scriptstyle \pm18}$ & $1.286\,{\scriptstyle \pm214}$ & $0.205\,{\scriptstyle \pm22}$ & $0.008\,{\scriptstyle \pm1}$ & $5.044\,{\scriptstyle \pm763}$ & $0.414\,{\scriptstyle \pm21}$ \\
\addlinespace
PBG 640M & 48 & standard & $0.572\,{\scriptstyle \pm37}$ & $0.414\,{\scriptstyle \pm204}$ & $0.109\,{\scriptstyle \pm42}$ & $0.253\,{\scriptstyle \pm12}$ & $1.162\,{\scriptstyle \pm580}$ & $0.168\,{\scriptstyle \pm52}$ & $0.016\,{\scriptstyle \pm5}$ & $3.867\,{\scriptstyle \pm635}$ & $0.356\,{\scriptstyle \pm55}$ \\
 &  & corrected & $0.576\,{\scriptstyle \pm38}$ & $0.410\,{\scriptstyle \pm212}$ & $0.099\,{\scriptstyle \pm45}$ & $0.262\,{\scriptstyle \pm13}$ & $1.152\,{\scriptstyle \pm547}$ & $0.161\,{\scriptstyle \pm50}$ & $0.017\,{\scriptstyle \pm5}$ & $3.890\,{\scriptstyle \pm591}$ & $0.349\,{\scriptstyle \pm53}$ \\
\addlinespace
PBG-Joint 160M & 198 & standard & $0.239\,{\scriptstyle \pm42}$ & $2.045\,{\scriptstyle \pm809}$ & $0.353\,{\scriptstyle \pm113}$ & $0.063\,{\scriptstyle \pm12}$ & $3.111\,{\scriptstyle \pm656}$ & $0.401\,{\scriptstyle \pm34}$ & $0.004\,{\scriptstyle {<}1}$ & $7.698\,{\scriptstyle \pm757}$ & $0.579\,{\scriptstyle \pm30}$ \\
 &  & corrected & $0.241\,{\scriptstyle \pm42}$ & $2.044\,{\scriptstyle \pm813}$ & $0.347\,{\scriptstyle \pm113}$ & $0.065\,{\scriptstyle \pm12}$ & $3.093\,{\scriptstyle \pm609}$ & $0.396\,{\scriptstyle \pm30}$ & $0.004\,{\scriptstyle {<}1}$ & $7.836\,{\scriptstyle \pm790}$ & $0.581\,{\scriptstyle \pm32}$ \\
\addlinespace
Ladder L7 30M & 48 & standard & $0.010\,{\scriptstyle \pm3}$ & $2.321\,{\scriptstyle \pm359}$ & $0.423\,{\scriptstyle \pm85}$ & $0.002\,{\scriptstyle {<}1}$ & $5.434\,{\scriptstyle \pm597}$ & $0.581\,{\scriptstyle \pm25}$ & $0.000\,{\scriptstyle {<}1}$ & $21.789\,{\scriptstyle \pm815}$ & $0.910\,{\scriptstyle \pm34}$ \\
 &  & corrected & $0.011\,{\scriptstyle \pm3}$ & $2.305\,{\scriptstyle \pm366}$ & $0.420\,{\scriptstyle \pm84}$ & $0.002\,{\scriptstyle {<}1}$ & $5.428\,{\scriptstyle \pm577}$ & $0.580\,{\scriptstyle \pm24}$ & $0.000\,{\scriptstyle {<}1}$ & $22.025\,{\scriptstyle \pm997}$ & $0.910\,{\scriptstyle \pm31}$ \\
\addlinespace
Ladder L8 30M & 48 & standard & $0.014\,{\scriptstyle \pm3}$ & $2.048\,{\scriptstyle \pm525}$ & $0.386\,{\scriptstyle \pm55}$ & $0.002\,{\scriptstyle {<}1}$ & $5.418\,{\scriptstyle \pm774}$ & $0.607\,{\scriptstyle \pm73}$ & $0.000\,{\scriptstyle {<}1}$ & $20.645\,{\scriptstyle \pm2307}$ & $0.897\,{\scriptstyle \pm87}$ \\
 &  & corrected & $0.014\,{\scriptstyle \pm3}$ & $2.027\,{\scriptstyle \pm523}$ & $0.383\,{\scriptstyle \pm53}$ & $0.002\,{\scriptstyle {<}1}$ & $5.373\,{\scriptstyle \pm788}$ & $0.602\,{\scriptstyle \pm73}$ & $0.000\,{\scriptstyle {<}1}$ & $20.680\,{\scriptstyle \pm2357}$ & $0.890\,{\scriptstyle \pm88}$ \\
\addlinespace
Ladder L7 85M & 48 & standard & $0.062\,{\scriptstyle \pm15}$ & $2.812\,{\scriptstyle \pm303}$ & $0.465\,{\scriptstyle \pm111}$ & $0.009\,{\scriptstyle \pm5}$ & $4.608\,{\scriptstyle \pm504}$ & $0.560\,{\scriptstyle \pm49}$ & $0.001\,{\scriptstyle {<}1}$ & $12.322\,{\scriptstyle \pm1040}$ & $0.708\,{\scriptstyle \pm60}$ \\
 &  & corrected & $0.064\,{\scriptstyle \pm16}$ & $2.813\,{\scriptstyle \pm324}$ & $0.461\,{\scriptstyle \pm114}$ & $0.010\,{\scriptstyle \pm5}$ & $4.578\,{\scriptstyle \pm495}$ & $0.559\,{\scriptstyle \pm51}$ & $0.001\,{\scriptstyle {<}1}$ & $12.508\,{\scriptstyle \pm930}$ & $0.704\,{\scriptstyle \pm52}$ \\
\addlinespace
Ladder L8 85M & 48 & standard & $0.120\,{\scriptstyle \pm14}$ & $1.245\,{\scriptstyle \pm91}$ & $0.253\,{\scriptstyle \pm23}$ & $0.019\,{\scriptstyle \pm1}$ & $3.619\,{\scriptstyle \pm702}$ & $0.440\,{\scriptstyle \pm56}$ & $0.001\,{\scriptstyle {<}1}$ & $10.627\,{\scriptstyle \pm915}$ & $0.653\,{\scriptstyle \pm33}$ \\
 &  & corrected & $0.124\,{\scriptstyle \pm15}$ & $1.238\,{\scriptstyle \pm95}$ & $0.246\,{\scriptstyle \pm25}$ & $0.020\,{\scriptstyle \pm1}$ & $3.607\,{\scriptstyle \pm722}$ & $0.438\,{\scriptstyle \pm58}$ & $0.001\,{\scriptstyle {<}1}$ & $10.671\,{\scriptstyle \pm851}$ & $0.640\,{\scriptstyle \pm34}$ \\
\addlinespace
Ladder L7 160M & 48 & standard & $0.191\,{\scriptstyle \pm17}$ & $2.345\,{\scriptstyle \pm263}$ & $0.410\,{\scriptstyle \pm20}$ & $0.045\,{\scriptstyle \pm10}$ & $3.634\,{\scriptstyle \pm147}$ & $0.437\,{\scriptstyle \pm16}$ & $0.003\,{\scriptstyle \pm1}$ & $8.510\,{\scriptstyle \pm429}$ & $0.564\,{\scriptstyle \pm50}$ \\
 &  & corrected & $0.194\,{\scriptstyle \pm17}$ & $2.354\,{\scriptstyle \pm249}$ & $0.405\,{\scriptstyle \pm22}$ & $0.048\,{\scriptstyle \pm10}$ & $3.621\,{\scriptstyle \pm143}$ & $0.433\,{\scriptstyle \pm16}$ & $0.003\,{\scriptstyle \pm1}$ & $8.501\,{\scriptstyle \pm348}$ & $0.555\,{\scriptstyle \pm48}$ \\
\addlinespace
Ladder L8 160M & 48 & standard & $0.199\,{\scriptstyle \pm64}$ & $2.394\,{\scriptstyle \pm1393}$ & $0.379\,{\scriptstyle \pm146}$ & $0.048\,{\scriptstyle \pm10}$ & $3.915\,{\scriptstyle \pm1736}$ & $0.412\,{\scriptstyle \pm101}$ & $0.003\,{\scriptstyle {<}1}$ & $8.632\,{\scriptstyle \pm1521}$ & $0.570\,{\scriptstyle \pm33}$ \\
 &  & corrected & $0.203\,{\scriptstyle \pm65}$ & $2.383\,{\scriptstyle \pm1395}$ & $0.371\,{\scriptstyle \pm145}$ & $0.051\,{\scriptstyle \pm10}$ & $3.940\,{\scriptstyle \pm1738}$ & $0.411\,{\scriptstyle \pm104}$ & $0.003\,{\scriptstyle {<}1}$ & $8.792\,{\scriptstyle \pm1539}$ & $0.571\,{\scriptstyle \pm28}$ \\
\bottomrule
\end{tabular}%
}

%% file: generated/sampling_model_architectures.tex
\begingroup
\small
\setlength{\tabcolsep}{2pt}
\begin{tabular*}{\linewidth}{@{\extracolsep{\fill}}lrrrrrrr@{}}
\toprule
Model & $n_{\mathrm{layers}}$ & $d_{\mathrm{model}}$ & $d_{\mathrm{ff}}$ & $H_{\mathrm Q}/H_{\mathrm{KV}}$ & $N_{4096}\,[10^6]$ & $w_{4096}\,[\mathrm{MB}]$ & $v_{\mathrm{KV}}\,[\mathrm{KiB}]$ \\
\midrule
PBG 8M & 4 & 384 & 1024 & 6/3 & 9.830 & 19.661 & 6 \\
PBG 15M & 4 & 512 & 1408 & 8/4 & 17.039 & 34.079 & 8 \\
PBG 30M & 8 & 512 & 1408 & 8/4 & 31.982 & 63.963 & 16 \\
PBG 49M & 8 & 640 & 1920 & 10/5 & 51.773 & 103.547 & 20 \\
PBG 85M & 10 & 768 & 2176 & 12/6 & 88.670 & 177.340 & 30 \\
PBG 160M & 24 & 768 & 2112 & 6/3 & 162.398 & 324.796 & 36 \\
PBG 320M & 28 & 1024 & 2816 & 8/4 & 334.496 & 668.991 & 56 \\
PBG 640M & 24 & 1536 & 4224 & 12/6 & 643.301 & 1286.603 & 72 \\
\bottomrule
\end{tabular*}
\endgroup

%% file: generated/rate_model_specification.tex
\begingroup\small
\noindent\begin{tabular}{p{0.25\linewidth}p{0.69\linewidth}}
\toprule
Parameter & Prior \\
\midrule
$a$ & $\mathcal N^{+}(0,0.1^2)$ \\
$\log \alpha$ & $\mathcal N(0,1^2)$ \\
$\log \beta$ & $\mathcal N(0,1^2)$ \\
$\log \tau$ & $\mathcal N(\log 0.1,1^2)$ \\
$\log \gamma$ & $\mathcal N(0,0.75^2)$ \\
$\log \kappa$ & $\mathcal N(0,0.75^2)$ \\
$\sigma_{\mathrm{AR}}$ & $\mathcal N^{+}(0,0.25^2)$ \\
$\sigma_{\mathrm{SD}}$ & $\mathcal N^{+}(0,0.25^2)$ \\
\bottomrule
\end{tabular}
\endgroup

%% file: generated/rate_model_parameters.tex
\begingroup\small
\setlength{\tabcolsep}{2pt}
\begin{tabular*}{\linewidth}{@{\extracolsep{\fill}}lrr@{}}
\toprule
Parameter & Median & 95\% interval \\
\midrule
$a$ & $0.0463$ & $[0.0439,\,0.0487]$ \\
$\alpha$ & $2.819$ & $[2.740,\,2.898]$ \\
$\beta$ & $0.964$ & $[0.955,\,0.972]$ \\
$\tau$ & $0.0948$ & $[0.0917,\,0.0980]$ \\
$\gamma$ & $2.095$ & $[1.550,\,3.006]$ \\
$\kappa$ & $2.208$ & $[2.075,\,2.345]$ \\
$\sigma_{\mathrm{AR}}$ & $0.0249$ & $[0.0235,\,0.0265]$ \\
$\sigma_{\mathrm{SD}}$ & $0.1160$ & $[0.1029,\,0.1322]$ \\
\bottomrule
\end{tabular*}
\endgroup

%% file: generated/rate_model_fit.tex
\begingroup\small
\setlength{\tabcolsep}{2pt}
\begin{tabular*}{\linewidth}{@{\extracolsep{\fill}}lrrr@{}}
\toprule
Metric & AR & SD & Speedup \\
\midrule
$R_T^2$ & $0.9982$ & $0.9359$ & --- \\
MAPE (\%) & $2.02$ & $9.57$ & $9.75$ \\
\bottomrule
\end{tabular*}
\endgroup

%% file: generated/dipper_mdqm9_table.tex
\resizebox{\linewidth}{!}{%
\begin{tabular}{@{}lcccccc@{}}
\toprule
 & \multicolumn{3}{c}{Easy (42 systems)} & \multicolumn{3}{c}{Hard (53 systems)} \\
\cmidrule(lr){2-4} \cmidrule(lr){5-7}
Model & $\mathrm{rESS}^{\ast}\uparrow$ & $\mathcal{E}\text{-}\mathcal{W}_2^{\ast}\downarrow$ & $\mathbb{T}\mathrm{g}\text{-}\mathcal{SW}_2^{\ast}\downarrow$ & $\mathrm{rESS}^{\ast}\uparrow$ & $\mathcal{E}\text{-}\mathcal{W}_2^{\ast}\downarrow$ & $\mathbb{T}\mathrm{g}\text{-}\mathcal{SW}_2^{\ast}\downarrow$ \\
\midrule
\text{PBG 47M} & $0.004$ & $3.975$ & $0.652$ & $0.001$ & $5.046$ & $0.952$ \\
\text{PBG 85M} & $\mathbf{0.018}$ & $\underline{3.168}$ & $\underline{0.596}$ & $\mathbf{0.005}$ & $\mathbf{4.706}$ & $\mathbf{0.781}$ \\
\text{PBG-Joint 160M} & $\underline{0.006}$ & $\mathbf{2.693}$ & $\mathbf{0.593}$ & $\underline{0.002}$ & $\underline{4.723}$ & $\underline{0.823}$ \\
\bottomrule
\end{tabular}%
}

%% file: generated/dipper_freesolv_table.tex
\resizebox{\linewidth}{!}{%
\begin{tabular}{@{}lccccccccc@{}}
\toprule
 & \multicolumn{3}{c}{$\leq 12$ atoms (31)} & \multicolumn{3}{c}{$13\text{--}24$ atoms (40)} & \multicolumn{3}{c}{$\geq 25$ atoms (23)} \\
\cmidrule(lr){2-4} \cmidrule(lr){5-7} \cmidrule(lr){8-10}
Environment & $\mathrm{rESS}^{\ast}\uparrow$ & $\mathcal{E}\text{-}\mathcal{W}_2^{\ast}\downarrow$ & $\mathbb{T}\mathrm{g}\text{-}\mathcal{SW}_2^{\ast}\downarrow$ & $\mathrm{rESS}^{\ast}\uparrow$ & $\mathcal{E}\text{-}\mathcal{W}_2^{\ast}\downarrow$ & $\mathbb{T}\mathrm{g}\text{-}\mathcal{SW}_2^{\ast}\downarrow$ & $\mathrm{rESS}^{\ast}\uparrow$ & $\mathcal{E}\text{-}\mathcal{W}_2^{\ast}\downarrow$ & $\mathbb{T}\mathrm{g}\text{-}\mathcal{SW}_2^{\ast}\downarrow$ \\
\midrule
Vacuum & $0.00069\,{\scriptstyle \pm35}$ & $2.194\,{\scriptstyle \pm84}$ & $0.697^{(30/31)}\,{\scriptstyle \pm124}$ & $0.00017\,{\scriptstyle \pm5}$ & $8.938\,{\scriptstyle \pm802}$ & $1.076\,{\scriptstyle \pm76}$ & $0.00010\,{\scriptstyle {<}1}$ & $27254\,{\scriptstyle \pm3455}$ & $1.723\,{\scriptstyle \pm16}$ \\
Implicit water & $0.00512\,{\scriptstyle \pm304}$ & $0.610\,{\scriptstyle \pm69}$ & $0.559^{(30/31)}\,{\scriptstyle \pm57}$ & $0.00045\,{\scriptstyle \pm21}$ & $4.394\,{\scriptstyle \pm714}$ & $0.946\,{\scriptstyle \pm33}$ & $0.00010\,{\scriptstyle {<}1}$ & $13544\,{\scriptstyle \pm4421}$ & $1.674\,{\scriptstyle \pm65}$ \\
\bottomrule
\end{tabular}%
}

%% file: generated/dipper_vacuum_peptides_table.tex
\resizebox{\linewidth}{!}{%
\begin{tabular}{lccccccccccc}
\toprule
 & & \multicolumn{3}{c}{$2\mathrm{AA}$ (30 systems)} & \multicolumn{3}{c}{$4\mathrm{AA}$ (30 systems)} & \multicolumn{3}{c}{$8\mathrm{AA}$ (30 systems)} \\
\cmidrule(lr){3-5} \cmidrule(lr){6-8} \cmidrule(lr){9-11}
Model & $T$ & $\mathrm{rESS}^{\ast}\uparrow$ & $\mathcal{E}\text{-}\mathcal{W}_2^{\ast}\downarrow$ & $\mathbb{T}\mathrm{g}\text{-}\mathcal{SW}_2^{\ast}\downarrow$ & $\mathrm{rESS}^{\ast}\uparrow$ & $\mathcal{E}\text{-}\mathcal{W}_2^{\ast}\downarrow$ & $\mathbb{T}\mathrm{g}\text{-}\mathcal{SW}_2^{\ast}\downarrow$ & $\mathrm{rESS}^{\ast}\uparrow$ & $\mathcal{E}\text{-}\mathcal{W}_2^{\ast}\downarrow$ & $\mathbb{T}\mathrm{g}\text{-}\mathcal{SW}_2^{\ast}\downarrow$ \\
\midrule
\text{PBG 85M (Implicit Water)} & $300\,\mathrm{K}$ & $0.00010$ & $52.544$ & $1.512$ & $0.00010$ & $65.228$ & $1.361$ & $0.00010$ & $143.966$ & $1.380$ \\
\text{PBG 85M Vacuum} & $300\,\mathrm{K}$ & $\mathbf{0.00020}$ & $\mathbf{7.192}$ & $\mathbf{1.115}$ & $\underline{0.00013}$ & $\mathbf{9.628}$ & $\underline{1.086}$ & $\underline{0.00011}$ & $\mathbf{39.858}$ & $\mathbf{1.275}$ \\
\text{PBG-Joint 160M} & $300\,\mathrm{K}$ & $\underline{0.00018}\,{\scriptstyle \pm4}$ & $\underline{7.737}\,{\scriptstyle \pm1707}$ & $\underline{1.126}\,{\scriptstyle \pm80}$ & $\mathbf{0.00018}\,{\scriptstyle \pm4}$ & $\underline{10.427}\,{\scriptstyle \pm825}$ & $\mathbf{1.022}\,{\scriptstyle \pm46}$ & $\mathbf{0.00012}\,{\scriptstyle \pm1}$ & $\underline{47.773}\,{\scriptstyle \pm2267}$ & $\underline{1.278}\,{\scriptstyle \pm51}$ \\
\bottomrule
\end{tabular}%
}

%% file: generated/asfe_peptide_charge_table.tex
\resizebox{\linewidth}{!}{%
\begin{tabular}{lrcccccccccc}
\toprule
 & & \multicolumn{3}{c}{BAR} & \multicolumn{3}{c}{Implicit water ($310\,\mathrm{K}$)} & \multicolumn{3}{c}{Vacuum ($300\,\mathrm{K}$)} \\
\cmidrule(lr){3-5} \cmidrule(lr){6-8} \cmidrule(lr){9-11}
Net charge & $n$ & Pass (fraction) & $r$ (GRAVY) & SE (kcal/mol) & $\mathrm{rESS}^{\ast}\uparrow$ & $\mathcal{E}\text{-}\mathcal{W}_2^{\ast}\downarrow$ & $\mathbb{T}\mathrm{g}\text{-}\mathcal{SW}_2^{\ast}\downarrow$ & $\mathrm{rESS}^{\ast}\uparrow$ & $\mathcal{E}\text{-}\mathcal{W}_2^{\ast}\downarrow$ & $\mathbb{T}\mathrm{g}\text{-}\mathcal{SW}_2^{\ast}\downarrow$ \\
\midrule
$-3$ & $1$ & $0$ ($0.00$) & \text{---} & \text{---} & $0.00042$ & $8.156$ & $0.747$ & $0.00018$ & $50.951$ & $1.231$ \\
$-2$ & $4$ & $1$ ($0.25$) & \text{---} & $0.043$ & $0.14998$ & $0.613$ & $0.129$ & $0.00023$ & $15.107$ & $1.037$ \\
$-1$ & $17$ & $9$ ($0.53$) & $-0.42$ & $0.043$ & $0.03251$ & $1.302$ & $0.152$ & $0.00021$ & $14.192$ & $1.147$ \\
$0$ & $46$ & $29$ ($0.63$) & $0.16$ & $0.030$ & $0.02463$ & $0.969$ & $0.176$ & $0.00014$ & $10.864$ & $1.140$ \\
$+1$ & $15$ & $3$ ($0.20$) & \text{---} & $0.027$ & $0.00137$ & $4.482$ & $0.493$ & $0.00013$ & $10.979$ & $1.190$ \\
$+2$ & $7$ & $0$ ($0.00$) & \text{---} & \text{---} & $0.00248$ & $3.516$ & $0.382$ & $0.00011$ & $30.588$ & $1.402$ \\
\midrule
All & $90$ & $42$ ($0.47$) & $0.28$ & $0.032$ & $0.01068$ & $2.148$ & $0.242$ & $0.00015$ & $13.375$ & $1.153$ \\
\bottomrule
\end{tabular}%
}

%% file: generated/microsolvated_counterion_audit.tex
\resizebox{\linewidth}{!}{%
\begin{tabular}{lrccc}
\toprule
Counterion & Peptides & Contact & Inside cluster & Mixed \\
\midrule
Na$^{+}$ (net charge $-3$ to $-1$) & $22$ & 45\% (17--88\%) & 65\% (24--96\%) & 20 of 22 \\
Cl$^{-}$ (net charge $+1$ to $+2$) & $22$ & 7\% (2--15\%) & 23\% (3--41\%) & 17 of 22 \\
\midrule
All charged & $44$ & 16\% (2--88\%) & 33\% (3--96\%) & 37 of 44 \\
\bottomrule
\end{tabular}%
}

%% file: generated/sota_training_configuration.tex
\begingroup
\small
\setlength{\tabcolsep}{2pt}
\begin{tabular*}{\linewidth}{@{\extracolsep{\fill}}llrrrrrrrrr@{}}
\toprule
Model & GPU type & Nodes & GPUs & Devices & \shortstack{Batch\\/device} & \shortstack{Global\\batch} & \shortstack{Loss\\tokens\\/step (k)} & \shortstack{Max.\\steps (k)} & \shortstack{Max.\\$D/N$} & \rebuttal{\shortstack{GPU-\\hours}} \\
\midrule
PBG 30M & RTX PRO 6000 & 1 & 2 & 2 & 128 & 256 & 182.3 & 289 & 1,760 & \rebuttal{37} \\
PBG 49M & RTX PRO 6000 & 1 & 2 & 2 & 112 & 224 & 159.6 & 128 & 416 & \rebuttal{17} \\
PBG 85M & RTX PRO 6000 & 1 & 2 & 2 & 80 & 160 & 114.2 & 575 & 768 & \rebuttal{149} \\
PBG 160M & H100 NVL & 1 & 2 & 2 & 64 & 128 & 91.5 & 612 & 352 & \rebuttal{102} \\
PBG 320M & H100 NVL & 1 & 2 & 2 & 32 & 64 & 45.8 & 619 & 86 & \rebuttal{183} \\
PBG 640M & MI250X & 4 & 16 & 32 & 16 & 512 & 365.2 & 84 & 48 & \rebuttal{1,138} \\
PBG-Joint 160M & H100 NVL & 1 & 2 & 2 & 64 & 128 & 91.2 & 346 & 198 & \rebuttal{117} \\
\bottomrule
\end{tabular*}
\endgroup